\documentclass[10pt,twocolumn,letterpaper]{article}

\usepackage[pagenumbers]{cvpr} 

\usepackage{graphicx}
\usepackage{amsmath}
\usepackage{amssymb}
\usepackage{times}
\usepackage{mathtools}
\usepackage{mathrsfs}  
\usepackage{amssymb}
\usepackage{booktabs}
\usepackage{multirow,multicol}
\usepackage{array}
\usepackage{tabularx}
\usepackage{balance}
\usepackage{placeins}
\usepackage{subcaption}
\usepackage{colortbl}
\usepackage{xcolor}
\usepackage{paralist}
\setdefaultleftmargin{1.5em}{}{}{}{}{}

\usepackage{fontawesome5}
\usepackage{pifont}
\usepackage{enumitem}
\usepackage{tikz}
\usepackage{pgfplots}
\usepackage{styles}
\usepackage{caption}
\usepackage[outline]{contour}
\usepackage{microtype}
\usetikzlibrary{tikzmark,patterns,shapes.misc,shapes.geometric,positioning,fit,calc}
\usepgfplotslibrary{polar}

\usepackage[hang,flushmargin]{footmisc}
\usepackage[accsupp]{axessibility}  

\definecolor{cvprblue}{rgb}{0.21,0.49,0.74}
\usepackage[pagebackref,breaklinks,colorlinks,allcolors=cvprblue]{hyperref}

\usepackage[capitalize]{cleveref}
\crefname{section}{Sec.}{Secs.}
\Crefname{section}{Section}{Sections}
\Crefname{table}{Table}{Tables}
\crefname{table}{Tab.}{Tabs.}

\setdefaultleftmargin{1.5em}{2em}{}{}{}{}

\def\paperID{33} 
\def\confName{CVPR}
\def\confYear{2025}

\begin{document}

\title{AnySat: One Earth Observation Model\\ for Many Resolutions, Scales, and Modalities}

\author{
    Guillaume Astruc \textsuperscript{1,3,4}
    \and
    Nicolas Gonthier \textsuperscript{1,2} 
    \and
    Clément Mallet \textsuperscript{1}
    \and
    Loic Landrieu\textsuperscript{1,4}
    \and 
    {\textsuperscript{1} LASTIG, Univ Gustave Eiffel, IGN, ENSG, France}
    \and 
     {\textsuperscript{2} IGN, France}
     \and 
     {\textsuperscript{3} CNES, France}
     \and 
    {\textsuperscript{4} LIGM, Ecole Nationale des Ponts et Chaussées, IP Paris, Univ Gustave Eiffel, CNRS, France}
}
\maketitle

\begin{abstract}
Geospatial models must adapt to the diversity of Earth observation data in terms of resolutions, scales, and modalities. However, existing approaches expect fixed input configurations, which limits their practical applicability. We propose AnySat, a multimodal model based on joint embedding predictive architecture (JEPA) and scale-adaptive spatial encoders, allowing us to train a single model on highly heterogeneous data in a self-supervised manner. To demonstrate the advantages of this unified approach, we compile GeoPlex, a collection of $5$ multimodal datasets with varying characteristics and $11$ distinct sensors. We then train a single powerful model on these diverse datasets simultaneously. 
Once fine-tuned or probed, we reach state-of-the-art results on the test sets of GeoPlex and for $6$ external datasets across various environment monitoring tasks: land cover mapping, tree species identification, crop type classification, change detection, climate type classification, and segmentation of flood, burn scar, and deforestation. Our code and models are available at \GITHUB.
\end{abstract}

\section{Introduction}

From a remote sensing perspective, the natural images of computer vision are remarkably uniform: they are captured by nearly identical sensors (standard cameras) with the same RGB channels and are often taken from similar perspectives. This consistency allows the creation of large composite image datasets from various sources \cite{oquab2023dinov2,schuhmann2022laion,gadre2024datacomp}, which are key for image foundation models to learn powerful, general-purpose features \cite{awais2023foundational}.

In contrast, Earth observation (EO) data displays significant variability in modalities, scales, and spatial, temporal, and spectral resolutions. Existing EO foundation models are generally trained on a single dataset with a specific format \cite{lu2024ai,guo2024skysense,bastani2023satlaspretrain,wang2022ssl4eo}, 
and cannot be applied to datasets with different input types without retraining from scratch---defeating the purpose of foundation models. EO foundation models should be able to seamlessly integrate new datasets for training and prediction, regardless of their resolution, scale, and modalities. As recent efforts provide more flexibility in terms of modalities \cite{astruc2024omnisat,jakubik2310foundation}, scale \cite{reed2023scale}, or spectral resolutions \cite{xiong2024dofa}, none fully leverage the diversity of EO sensors.

\begin{figure}
    \centering
   \input{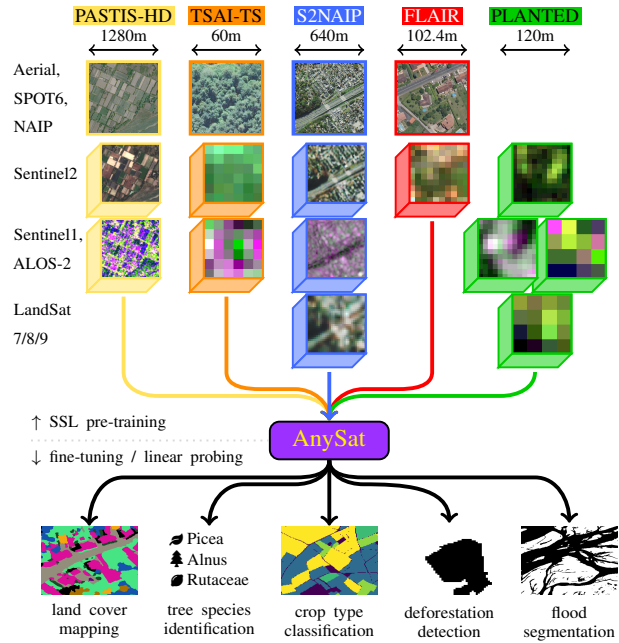}
   \vspace{-12mm}
    \caption{{\bf Multi-Dataset Training.}
    For the first time, {a single model} can be pretrained \textbf{simultaneously} on a collection of Earth Observation datasets with heterogeneous resolutions, scales, and modalities. The resulting model can be fine-tuned to achieve state-of-the-art results for a wide variety of data types and tasks.}
    \label{fig:teaser}
\end{figure}

{We introduce \textbf{AnySat}, a novel EO model using the spatial alignment of multiple modalities as a source of self-supervision. Indeed, while multiple observations of the same area from distinct sensors capture different information, they share the same underlying semantics. Therefore, we can expect the learned representations to be consistent across modalities. Moreover, we should be able to reconstruct missing modalities from available ones, encouraging the use of cross-modal masked auto-encoding techniques \cite{he2022masked,astruc2024omnisat}. However, EO data are subject to complex disruptors such as weather conditions, acquisition angles, and variations in time of day or year. To overcome this issue, we design a new multimodal Joint Embedding Predictive Architecture (JEPA) \cite{assran2023self} to learn representations that are consistent \emph{in feature space}.}

{A key advantage of our JEPA model is that it eliminates the need for modality-specific decoders, allowing us to handle a wide variety of sensors seamlessly. Combined with our scale-adaptive patch encoder architecture, this approach enables us to train a single model on highly heterogeneous collections of multimodal EO datasets. Notably, over 75\% of the learnable parameters in our model are shared across all modalities and resolutions, and thus fully benefit from large and varied training data for self-supervision.}

To evaluate our approach, we compile \textbf{GeoPlex}, a collection of $5$ multimodal datasets including $11$ distinct modalities, with aerial images and satellite time series, radar and optical sensors. GeoPlex spans various spatial resolutions (from $0.2$ to $250$ m per pixel), revisit times (from single images to weekly time series), channel counts ($3$ to $11$), and spatial extent (samples ranging from $0.4$ to $160$K hectares). 
To showcase the versatility of AnySat, we also consider {$6$} external evaluation datasets with diverse characteristics. 
After fine-tuning, AnySat achieves state-of-the-art performance on {$9$} downstream tasks, including classification, segmentation, and change segmentation across domains such as land cover mapping, crop type classification, tree species identification, and deforestation detection.
Our contributions are as follows: 
\begin{compactitem} 
\item We present {AnySat}, a versatile architecture capable of learning from multiple EO sources with heterogeneous resolutions, scales, and modalities. 
\item We introduce the first application of JEPA for multimodal EO data, enabling large-scale and efficient self-supervised learning.
\item We demonstrate that, when pretrained on a curated collection of EO datasets, AnySat can be fine-tuned or linearly-probed to achieve state-of-the-art performance across a diverse array of tasks and datasets. 
\end{compactitem}
Thanks to its flexible design, our pretrained model can be applied to scales ranging from a single forest plot to tiles covering hundreds of square kilometers, and adapt to diverse sensor setups---from unimodal data to any combination of the 11 sensors featured in GeoPlex. In addition, we demonstrate that AnySat successfully generalizes to new sensor configurations not present in its training set.

\section{Related Work}

In this section, we review the dynamic field of self-supervised learning in geospatial models, highlighting recent efforts to enhance their adaptability to diverse inputs. Finally, we present the feature-predictive paradigm, which is instrumental to improve the versatility of EO models.

\paragraph{Self-Supervised Geospatial Models.}
The abundance of raw EO data makes it particularly suitable for self-supervised learning approaches \cite{tseng2022croco, ayush2021geography, manas2021seasonal, bourcier2024learning}. Generative models 
leverage the unique properties of EO data with adapted strategies such as spectral  \cite{cong2022satmae}, temporal \cite{dumeur2024self,dumeur2024paving}, and spatio-temporal \cite{yuan2022sits,jakubik2310foundation}, and hybrid \cite{tseng2023lightweight} masking.
Other approaches predict rotated \cite{li2024masked} or rescaled \cite{noman2024rethinking,tang2024cross, reed2023scale} versions of the input data, or predict missing modalities from available ones \cite{fuller2023croma,astruc2024omnisat}.
However, these models are often trained on specific combinations of modalities and are limited to those modalities during inference, which hinders their applicability as foundation models expected to adapt to diverse scenarios.

\paragraph{Versatile EO Models.}
Several approaches have been proposed to improve the generalizability of EO models. Some models address variability in {spatial resolutions}  by training on images of different resolutions and generalizing to coarser scales \cite{reed2023scale}, while others manage { spectral variability} by training on sensors with different spectral bands \cite{xiong2024dofa}. { Temporal adaptability} is achieved in models capable of handling both single-date images and image time series \cite{astruc2024omnisat, guo2024skysense, bastani2023satlaspretrain}. Attempts have also been made to generalize across modalities by training on data from different sensors \cite{jakubik2310foundation, hsu2024geospatial} or {and even text or audio~\cite{sastry2025taxa}}. Despite these efforts, many models are still trained with a single scale and expect the input to have a certain shape, typically $224\times 224$ pixels. They resize other inputs to fit the model architecture, leading to inefficiencies for smaller inputs \cite[Tab~5]{tseng2023lightweight}. A key obstacle preventing the creation of truly versatile generative self-supervised models is the requirement for multiple encoders, decoders, and augmentations to handle different configurations. In this paper, we explore feature-predictive architectures as a promising solution to this challenge.

\paragraph{Feature-Predictive Architectures.}
Self-supervised learning methods have achieved significant success in image analysis \cite{chen2020simple, he2020momentum, oquab2023dinov2}. These approaches learn without labels using pretext tasks, which can be discriminative \cite{gidaris2018unsupervised, noroozi2016unsupervised}, contrastive \cite{chen2020simple, he2020momentum, grill2020bootstrap, caron2021emerging}, or generative, where the model predicts a degraded version of its input \cite{vincent2008extracting, he2022masked}. Recent works have proposed performing reconstruction in feature space rather than input space (\textit{e.g.}, pixel space) \cite{baevski2022data2vec, yi2023masked}.
Among feature-predictive architectures, the Joint Embedding Predictive Architecture (JEPA) has shown particular promise \cite{assran2023self} by learning to predict the features of masked parts of an input image.
Feature space reconstruction based model can also be combined with contrastive objectives for improved stability and representation quality \cite{baevski2022data2vec}.
\begin{figure*}[t]
    \centering
    \input{figures/architecture_patch}
    \vspace{-2mm}
    \caption{{\bf Scale-Adaptive Patch Encoding.}
    We consider a patch $x^m_p$ of resolution $\Delta_m=P/R_m$ pixels.
    We first split $x^m_p$ into sub-patches of size $\delta_m$ pixels, which are mapped by a modality-specific projector $\ProjEncod_m$ to a $E$-dimensional embedding. Then, a shared spatial transformer module $\TransEncod$ combines all sub-patches into a vector of size $E$. As the sub-patch size $\delta_m$ is fixed, the patch sizes $\Delta_m$ only influences the number of input tokens to $\TransEncod$, allowing us to use the same network for different resolutions. 
    }
    \label{fig:patchencod}
\end{figure*}

Because it bypasses the need for complex data augmentations or decoder networks, JEPA is particularly well-suited for massively multimodal applications such as Earth observation. SAR-JEPA \cite{li2024predicting} introduces the first implementation of JEPA concepts for EO, focusing exclusively on SAR data. In this paper, we combine JEPA with a versatile spatial encoder architecture, allowing a single model to handle diverse data scales, resolutions, and modalities. 
\section{Method}

We first describe our proposed architecture (\cref{sec:archi}) and self-supervised training procedure (\cref{sec:training}). Then, we detail the fine-tuning and probing methods used for downstream tasks (\cref{sec:downstream}). Our work focuses primarily on multi-dataset self-supervised training. However, for clarity, we initially describe the method for a single multimodal dataset, later generalizing it to multiple datasets.

\subsection{Architecture}

\label{sec:archi}
Tiles with multimodal observations are first partitioned into spatially aligned patches. Unlike classical Vision Transformers~\cite{dosovitskiy2020image}, our model supports patches of varying sizes, accommodating the significant scale variations common in Earth Observation (EO) datasets. Each patch is embedded via a scale-adaptive patch encoder, after which a combiner network integrates representations from multiple modalities into a unified spatial embedding.

Formally, we consider a tile $x$ of size $S \times S$ meters, observed through multiple modalities $\bM$. Each modality $m \in \bM$ has its own resolution $R_m$ (meters per pixel), temporal observations $T_m$ (with $T_m = 1$ for single-date modalities), and number of channels $C_m$ (e.g., spectral or polarization channels). The tile $x$ observed in modality $m$ is denoted $x^m$ and is represented as a tensor of shape $(S/R_m) \times (S/R_m) \times T_m \times C_m$.

\paragraph{Spatially Consistent Patching.} Tiles are partitioned into a set $\bP$ of non-overlapping patches, each of size $P \times P$ meters. An input token ${x}_p^m$ represents the observation of patch $p \in \bP$ in modality $m$. All modalities share the same spatial patch layout, ensuring spatial consistency across modalities. The total number of tokens is thus $|\bM| \cdot (S / P)^2$. Although the patch size is constant across modalities, each token may have distinct tensor dimensions due to differing resolutions, temporal extents, and channel numbers.

\paragraph{Patch Encoding.} 
We design a scale-adaptive patch encoder $\phi^{\text{patch}}$ to map each input token ${x}_p^m$ into a fixed-size vector $f_p^m \in \mathbb{R}^E$, regardless of modality resolution $R_m$ or patch size $P$. The encoding scheme, illustrated in \cref{fig:patchencod}, involves three stages:
\begin{compactitem}
\item[(i)] We first subdivide each token into fixed-size sub-patches of $\delta_m \times \delta_m$ pixels, flattening their spatial dimensions to vectors of size $\delta_m^2 T_m C_m$.
\item[(ii)] Each flattened sub-patch is mapped to dimension $E$ via a modality-specific MLP \text{$\ProjEncod_m$}. For multi-temporal modalities ($T_m>1$), a Lightweight Temporal Attention Encoder (LTAE)\cite{garnot2020lightweight} collapses the temporal dimension.
\item[(iii)] We add positional encodings based on ground sampling distance \cite{reed2023scale} to the sub-patch embeddings. A shared transformer network \text{$\TransEncod$} with $B$ blocks aggregates the sub-patch embeddings into a single representation $f_p^m$ per modality using a \texttt{CLS}-like token.
\end{compactitem}
Using sub-patches of fixed sizes $\delta_m$ allows $\PatchEncod$ to process patches of different patch sizes $P$ \emph{without rescaling} the input data. Indeed, changes in $P$ only influence the number of input tokens processed by $\TransEncod$, which has no incidence on the embedding size.

\begin{figure*}
    \centering
    \def\xinput{-0.75}
\def\xpatch{3}
\def\xencoder{5}
\def\xtoken{7.5}
\def\xcombiner{9}
\def\xtokenrec{11}
\def\xpredictor{12}
\def\xtokenrecrec{14}
\def\ystudent{2.5}
\def\yteacher{-2.5}
\def\ymoda{+2}
\def\ymodb{0}
\def\ymodc{-2}

\def\modsize{1}
\def\patchsize{0.7}
\def\xshift{0.5}
\def\posencsize{0.7}

\def\imgsize{1}
\def\sitssize{0.85}
\def\sitssizeb{0.85}
\def\appwidth{1.5}
\def\appheight{1.1}
\def\pastisx{0}
\def\tsaix{1.5}
\def\snaipx{3}
\def\flairx{4.5}
\def\plantedx{6.0}
\def\sitsshift{0.1}
\def\namey{+1.2}
\def\vhry{0}
\def\stwoy{-1.1}
\def\soney{-2.2}
\def\landsaty{-3.3}
\def\legendx{-1.1}
\def\anysaty{-5}
\def\arrowy{+0.7}

\begin{tikzpicture}


     \node[draw=none, fill=none, thick, inner sep=2pt, scale=0.55] (geoplex) at (\xinput,0) {
    \begin{tikzpicture}
    \fill [fill=gray!10, rounded corners=1cm] (-0.6,0.6) rectangle (7.1,-4.1);


 \node  [draw=none, inner sep=0pt, draw=PASTISCOLOR, line width=1mm] (pastisVHR) at (\pastisx,\vhry) {\includegraphics[width= \imgsize cm]{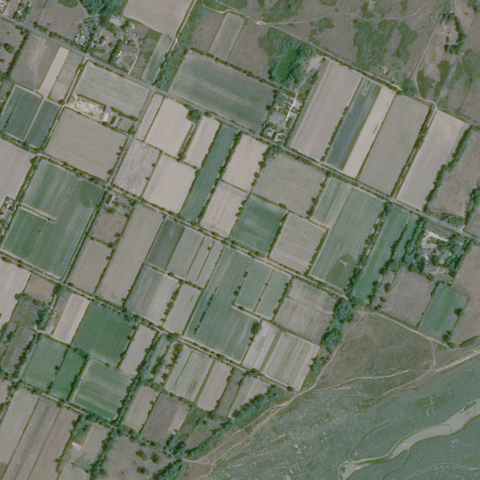}};

\node  [draw=none, inner sep=0pt] (pastisS2) at (\pastisx+\sitsshift,\stwoy) {\includegraphics[width= \sitssize cm]{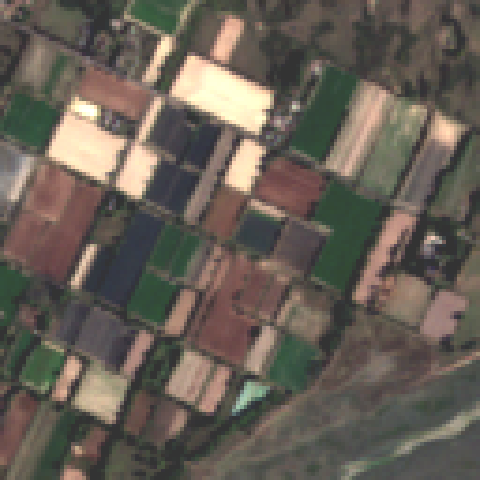}};
 \cube{\pastisx+\sitsshift}{\stwoy}{\sitssize}{PASTISCOLOR}

 \node  [draw=none, inner sep=0pt] (pastisS1) at (\pastisx+\sitsshift,\soney) {\includegraphics[width= \sitssize cm]{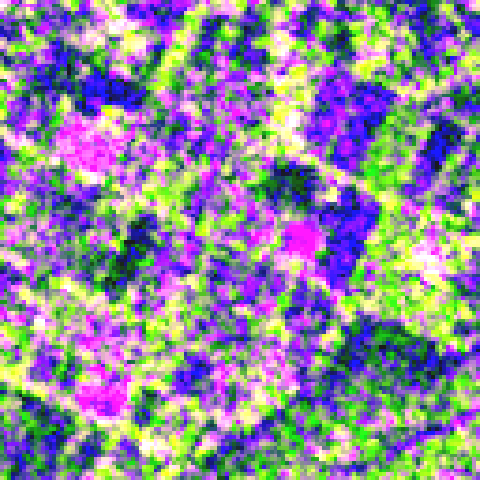}};
 \cube{\pastisx+\sitsshift}{\soney}{\sitssize}{PASTISCOLOR}


 \node  [draw=none, inner sep=0pt, draw=S2NAIPCOLOR, line width=1mm] (snaipVHR) at (\snaipx,\vhry) {\includegraphics[width= \imgsize cm]{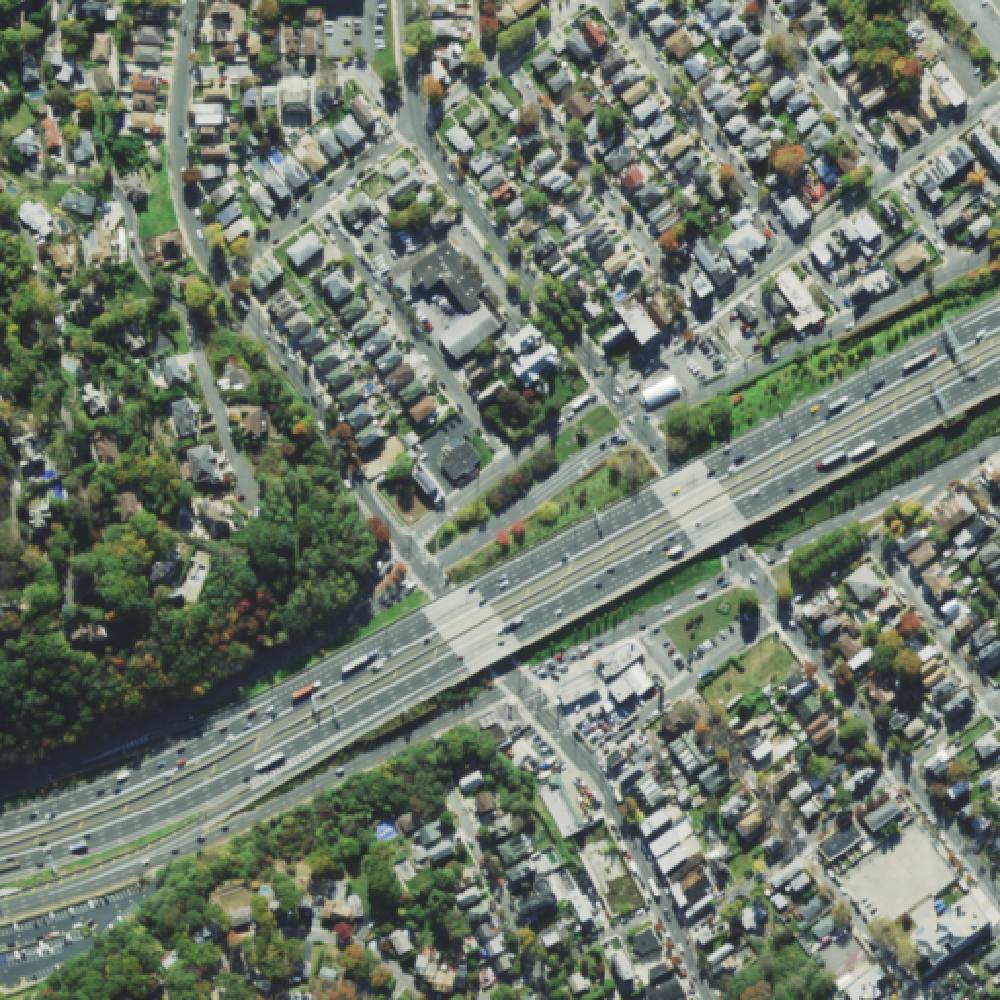}};

   \node  [draw=none, inner sep=0pt] (snaipS2) at (\snaipx+\sitsshift,\stwoy) {\includegraphics[width= \sitssize cm]{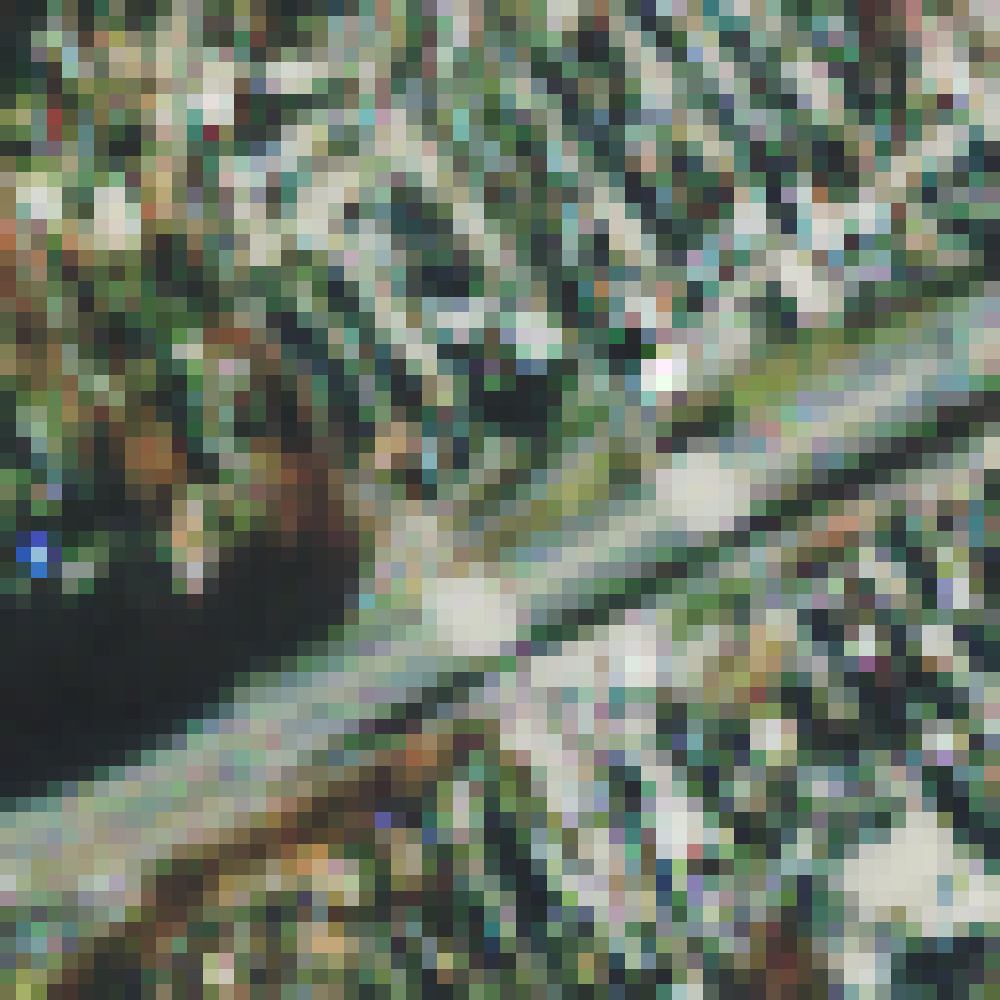}};
 \cube{\snaipx+\sitsshift}{\stwoy}{\sitssize}{S2NAIPCOLOR}
 
 \node  [draw=none, inner sep=0pt] (snaipS1) at (\snaipx+\sitsshift,\soney) {\includegraphics[width= \sitssize cm]{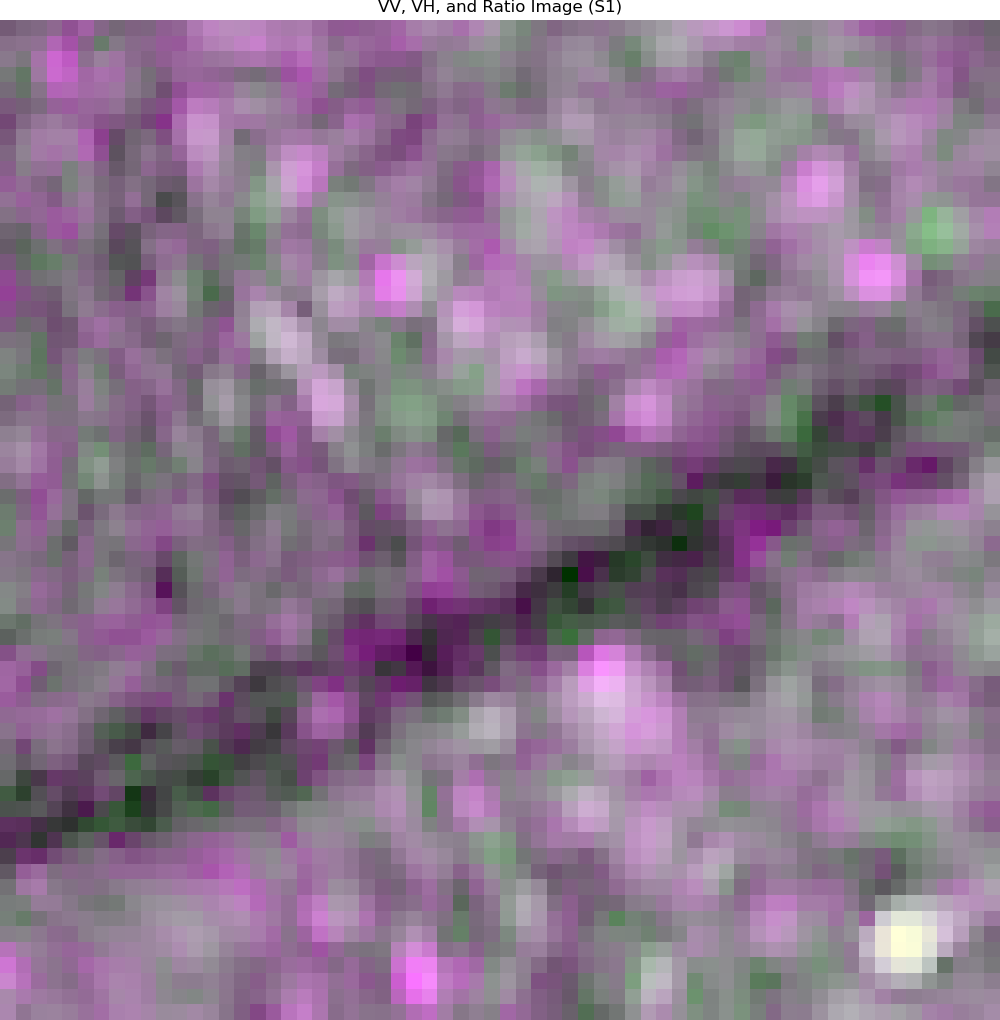}};
 \cube{\snaipx+\sitsshift}{\soney}{\sitssize}{S2NAIPCOLOR}

   \node  [draw=none, inner sep=0pt] (snaipS2) at (\snaipx+\sitsshift,\landsaty) {\includegraphics[width= \sitssize cm]{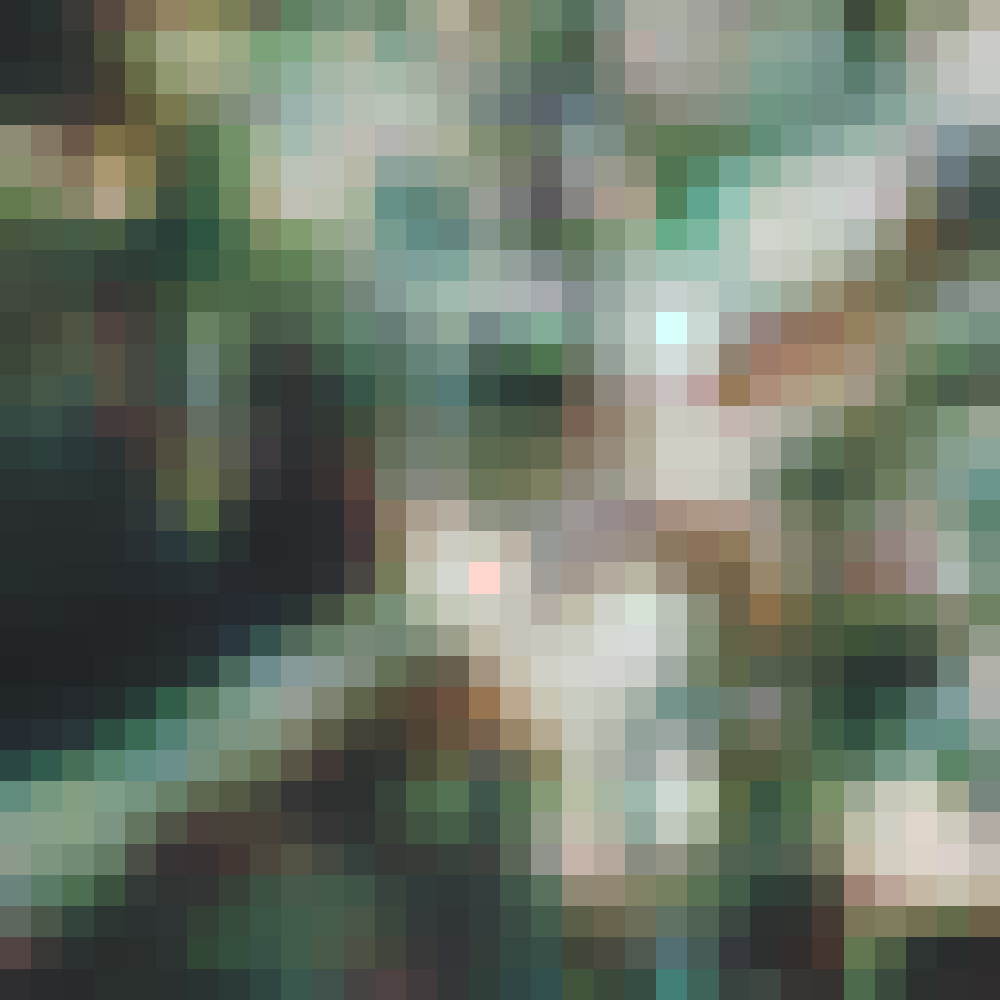}};
 \cube{\snaipx+\sitsshift}{\landsaty}{\sitssize}{S2NAIPCOLOR}


 \node  [draw=none, inner sep=0pt, draw=TSAITCOLOR, line width=1mm] (snaipVHR) at (\tsaix,\vhry) {\includegraphics[width= \imgsize cm]{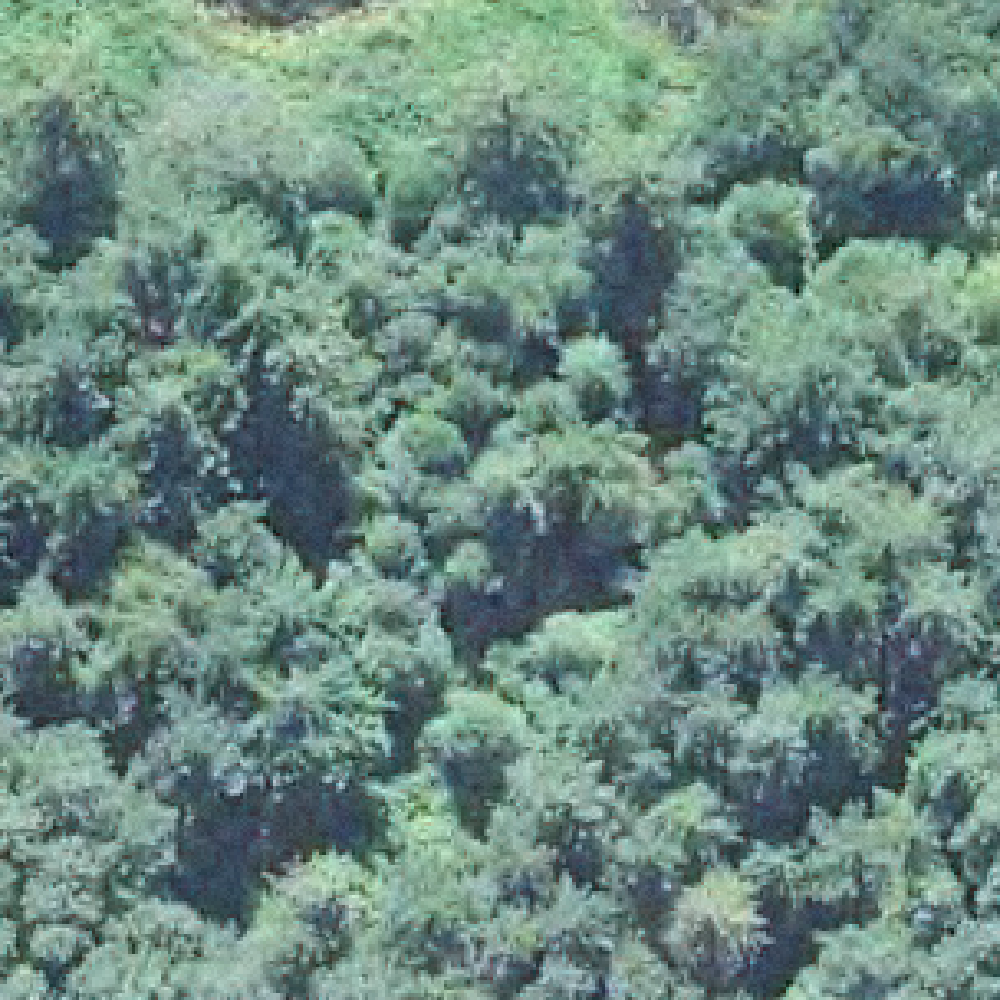}};

   \node  [draw=none, inner sep=0pt] (snaipS2) at (\tsaix+\sitsshift,\stwoy) {\includegraphics[width= \sitssize cm]{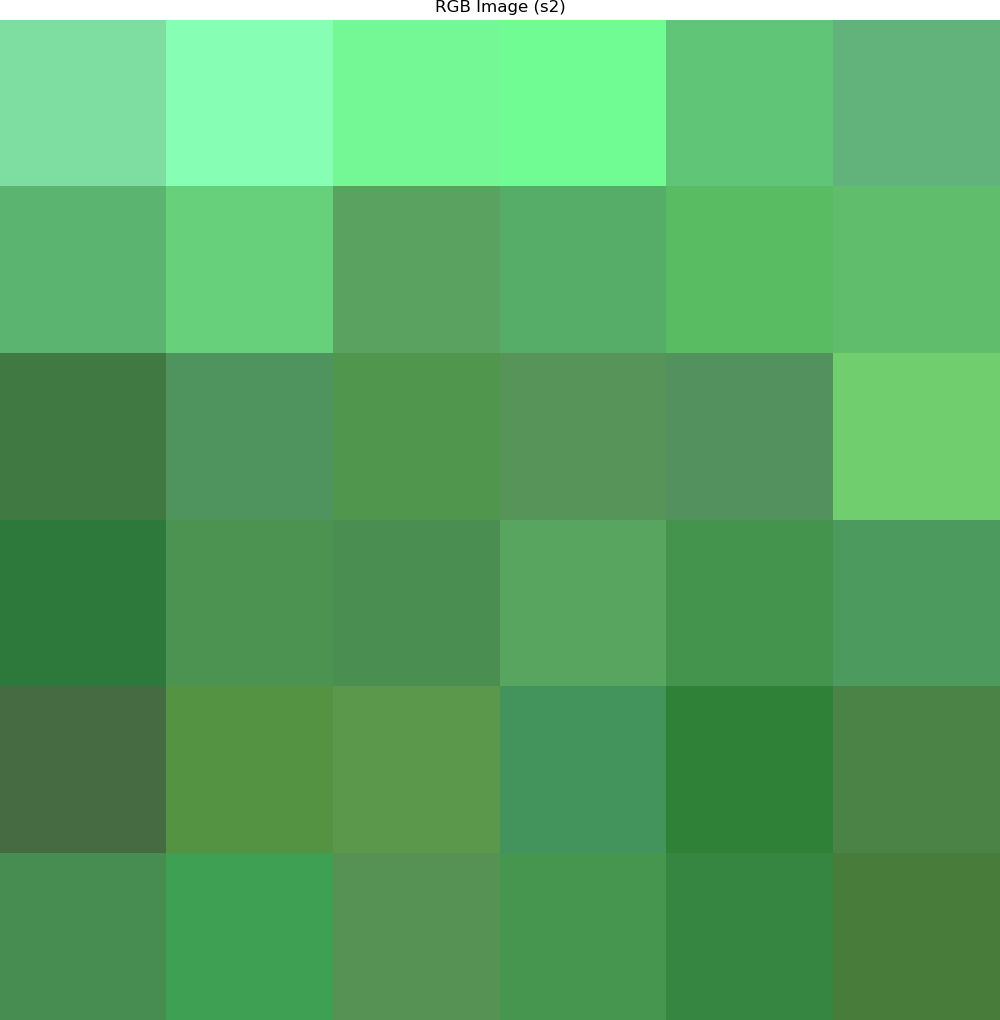}};
 \cube{\tsaix+\sitsshift}{\stwoy}{\sitssize}{TSAITCOLOR}
 
 \node  [draw=none, inner sep=0pt] (snaipS1) at (\tsaix+\sitsshift,\soney) {\includegraphics[width= \sitssize cm]{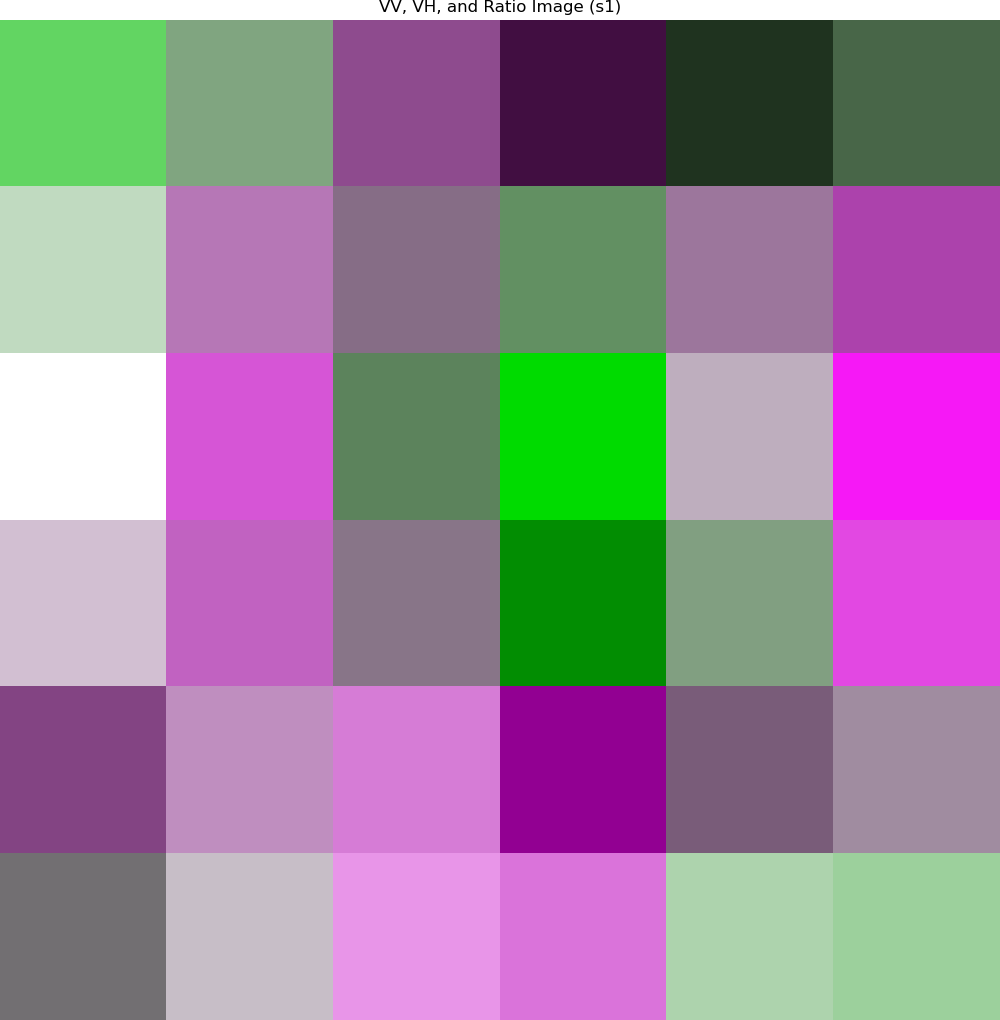}};
 \cube{\tsaix+\sitsshift}{\soney}{\sitssize}{TSAITCOLOR}
 

 \node  [draw=none, inner sep=0pt, draw=FLAIRCOLOR, line width=1mm] (snaipVHR) at (\flairx,\vhry) {\includegraphics[width= \imgsize cm]{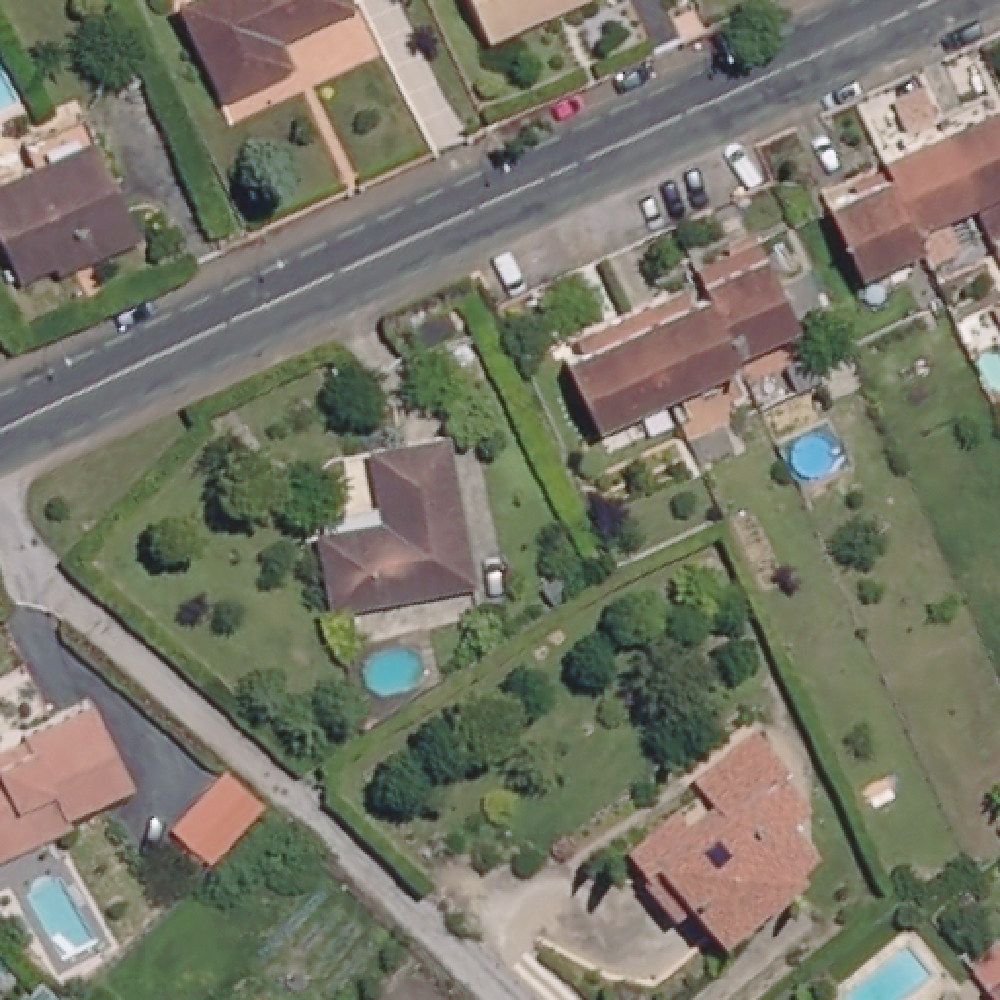}};

  \node  [draw=none, inner sep=0pt] (snaipS2) at (\flairx+\sitsshift,\stwoy) {\includegraphics[width= \sitssize cm]{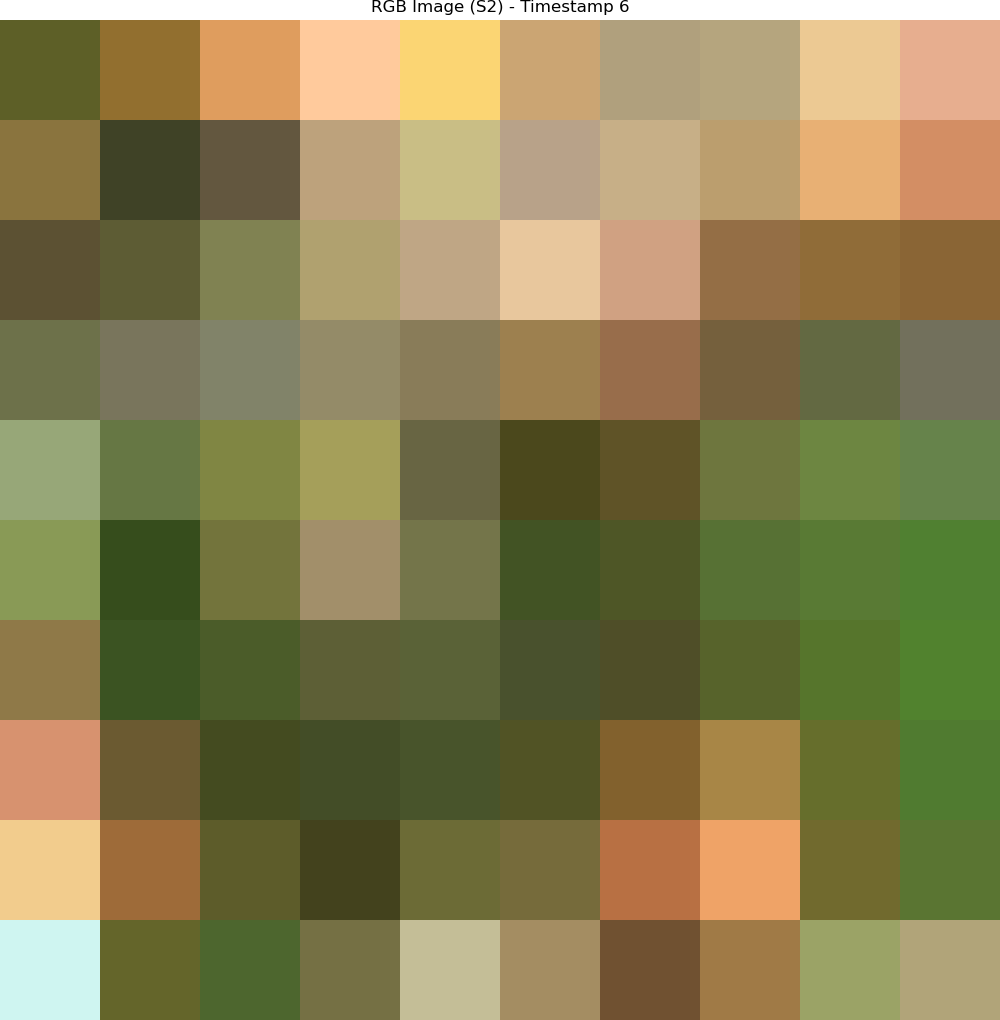}};
 \cube{\flairx+\sitsshift}{\stwoy}{\sitssize}{FLAIRCOLOR}

   \node  [draw=none, inner sep=0pt] (plantedS2) at (\plantedx+\sitsshift,\stwoy) {\includegraphics[width= \sitssize cm]{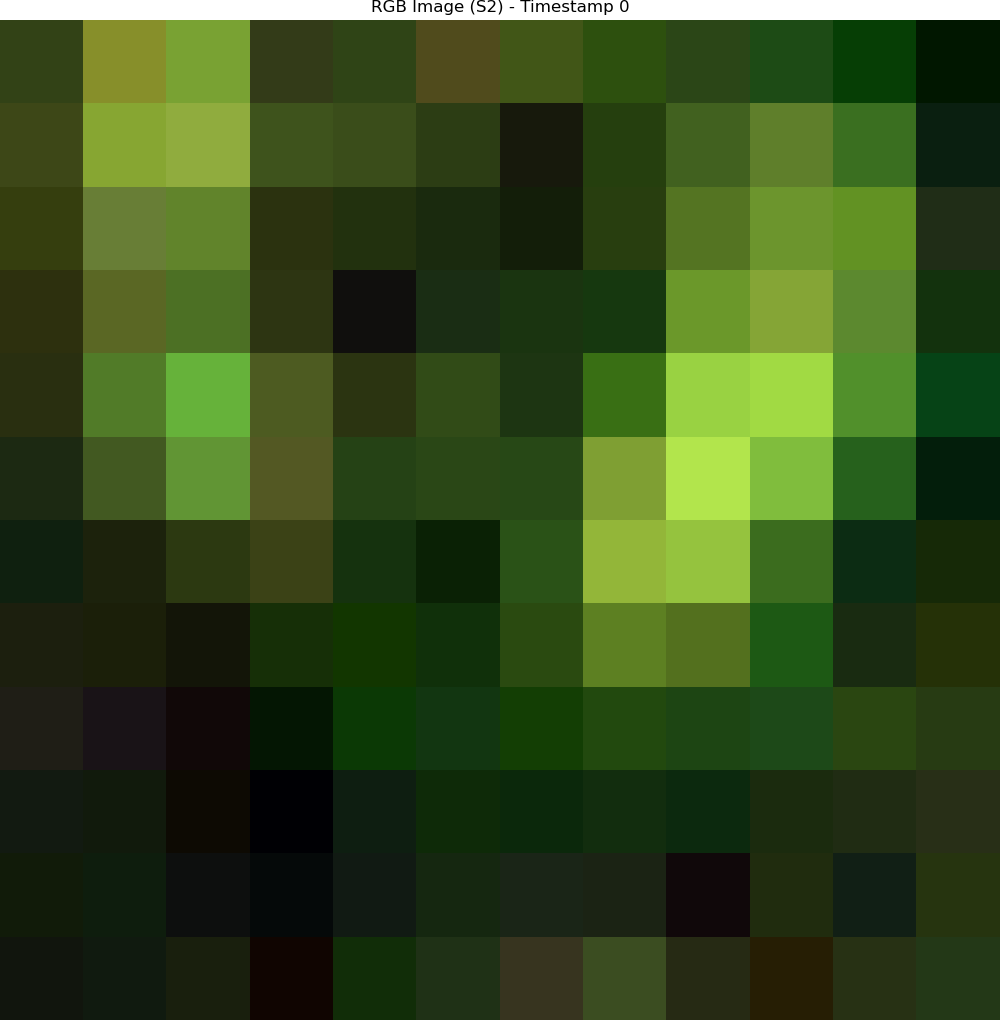}};
 \cube{\plantedx+\sitsshift}{\stwoy}{\sitssize}{PLANTEDCOLOR}

   \node  [draw=none, inner sep=0pt] (plantedS1) at (\plantedx+0.5+\sitsshift,\soney) {\includegraphics[width= \sitssizeb cm]{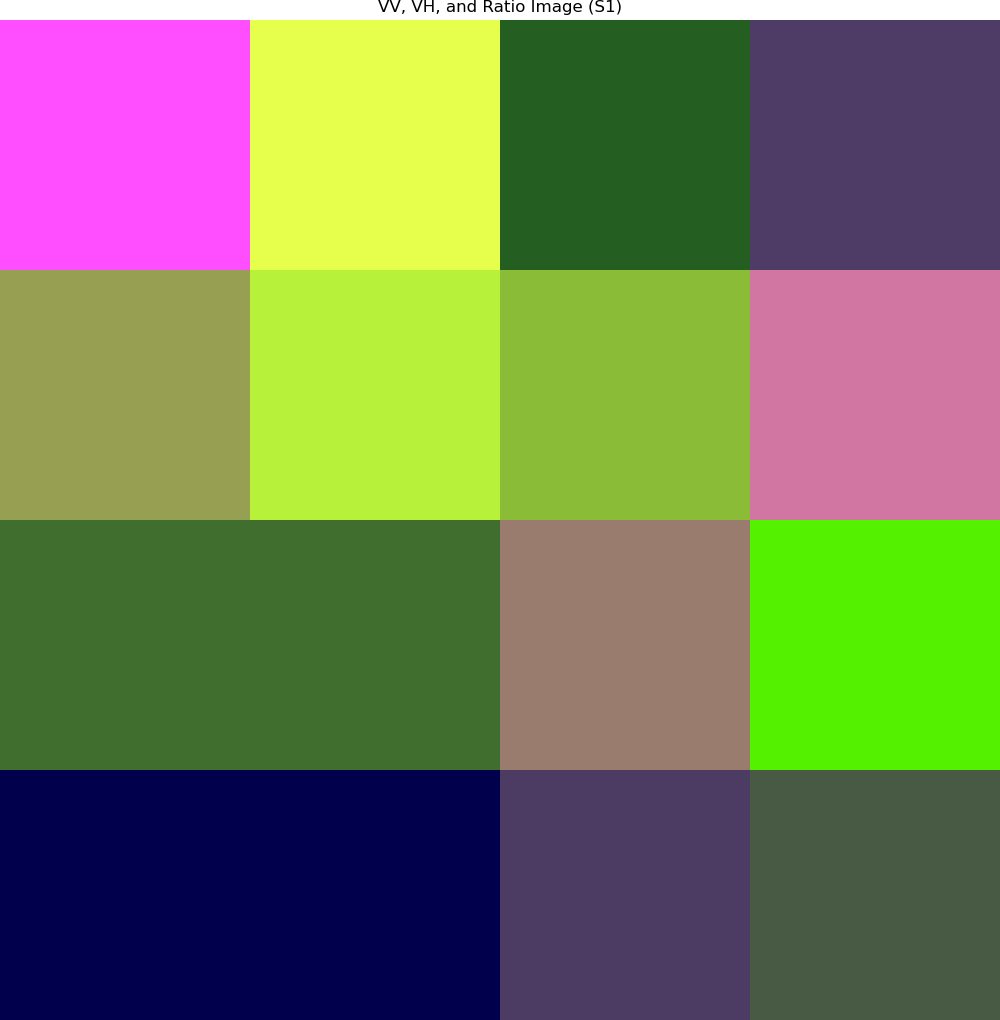}};
 \cube{\plantedx+0.5+\sitsshift}{\soney}{\sitssizeb}{PLANTEDCOLOR}
 
 \node  [draw=none, inner sep=0pt] (plantedS1) at (\plantedx-0.5+\sitsshift,\soney) {\includegraphics[width= \sitssizeb cm]{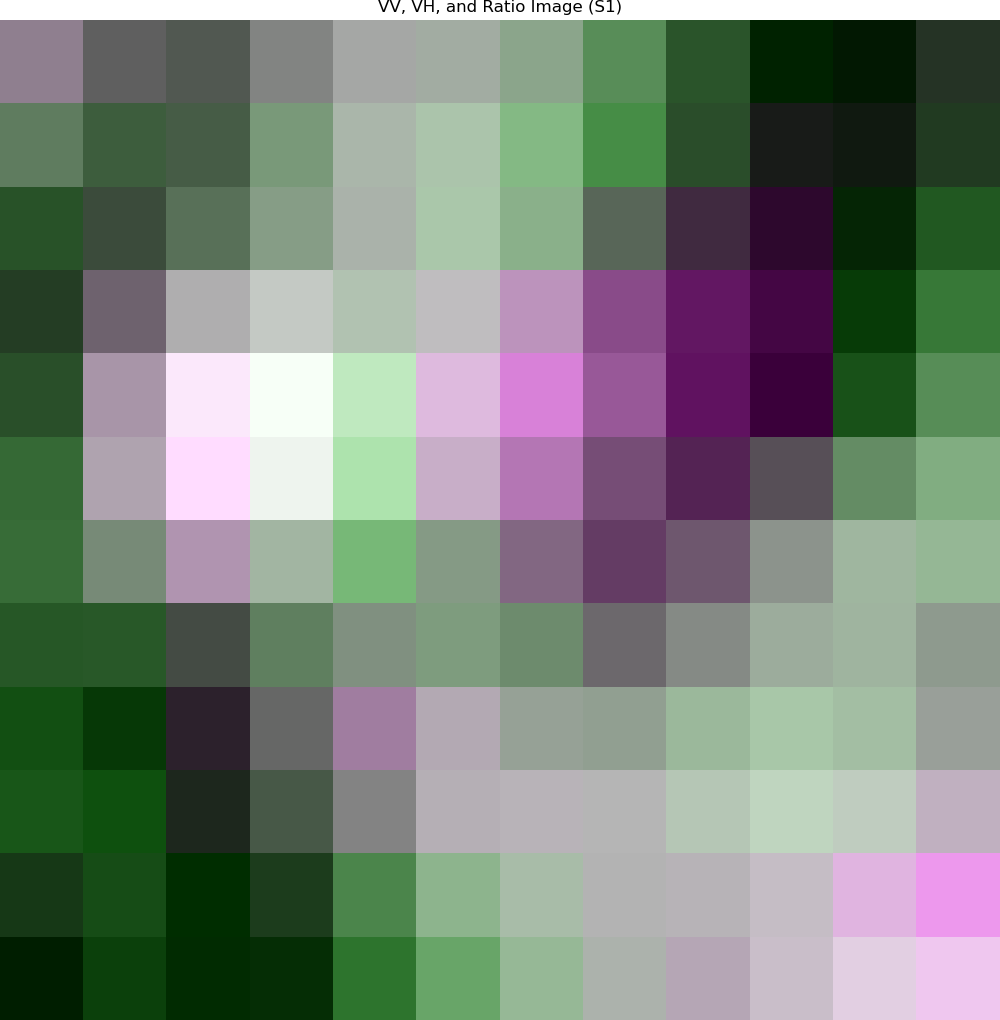}};
 \cube{\plantedx-0.5+\sitsshift}{\soney}{\sitssizeb}{PLANTEDCOLOR}

 \node  [draw=none, inner sep=0pt] (plantedS2) at (\plantedx+\sitsshift,\landsaty) {\includegraphics[width= \sitssize cm]{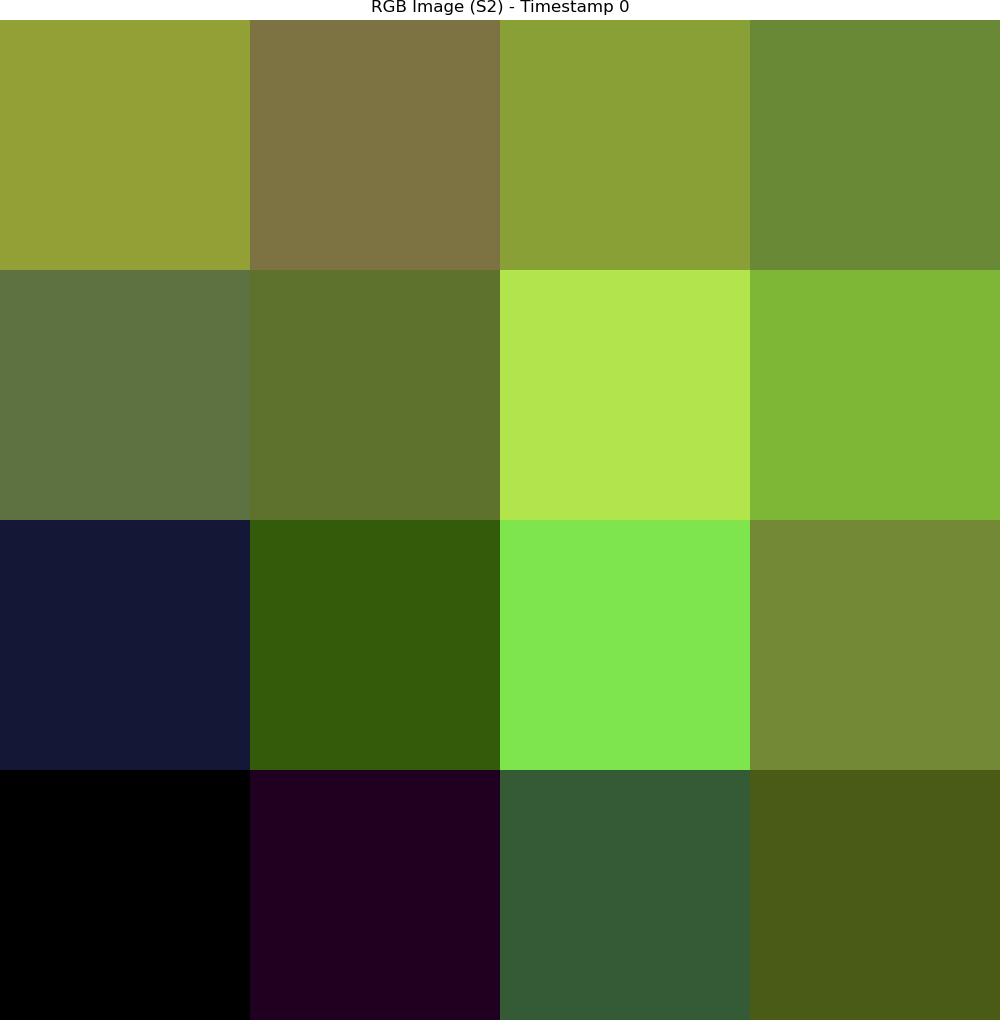}};
 \cube{\plantedx+\sitsshift}{\landsaty}{\sitssize}{PLANTEDCOLOR}

    \end{tikzpicture}
    };

    \node[draw=none, above =0mm of geoplex.north] {GEOPLEX};

    \draw[rounded corners=10pt, very thick, draw=STUDENTCOLOR!40, fill=STUDENTCOLOR!10] 
    ($ (\xencoder, \ystudent) + (-3mm, 15mm) $) rectangle 
    ($ (\xtokenrecrec, \ystudent) + (5mm, -15mm) $);

    \draw[rounded corners=10pt, very thick, draw=TEACHERCOLOR!40, fill=TEACHERCOLOR!10] 
    ($ (\xencoder, \yteacher) + (-3mm, 15mm) $) rectangle 
    ($ (\xtokenrecrec, \yteacher) + (5mm, -15mm) $);

    \node[text=STUDENTCOLOR] at ($ (\xpredictor,0.7) + (5mm,0mm) $) {\bf STUDENT};
    \node[text=TEACHERCOLOR] at ($ (\xpredictor,-0.7) + (5mm,0mm) $) {\bf TEACHER};

    \node (modapatch1) [minimum width=\patchsize cm,minimum height=\patchsize cm, draw=none, inner sep=0pt, draw=MODONECOLOR, line width=1mm] at (\xpatch-0.5,\ymoda+0.5) {\includegraphics[width=\patchsize cm]{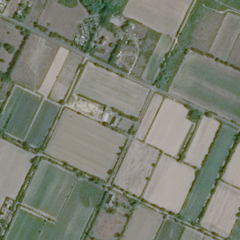}};
    \node (modapatch2) [minimum width=\patchsize cm,minimum height=\patchsize cm, draw=none,inner sep=0pt, draw=MODONECOLOR, line width=1mm] at (\xpatch+0.5,\ymoda+0.5) {\includegraphics[width=\patchsize cm]{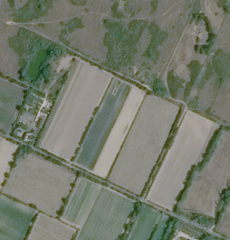}};
    \node (modapatch3) [minimum width=\patchsize cm,minimum height=\patchsize cm, draw=none, inner sep=0pt, draw=MODONECOLOR, line width=1mm] at (\xpatch-0.5,\ymoda-0.5) {\includegraphics[width=\patchsize cm]{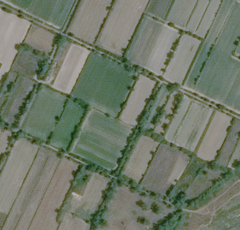}};
    \node (modapatch4) [minimum width=\patchsize cm,minimum height=\patchsize cm, draw=none, inner sep=0pt, draw=MODONECOLOR, line width=1mm] at (\xpatch+0.5,\ymoda-0.5) {\includegraphics[width=\patchsize cm]{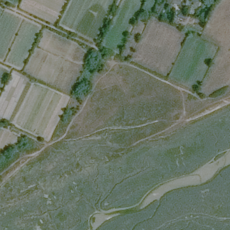}};

    \node[inner sep=0pt, opacity=0.0, draw=none, line width=0mm] (modb) at (\xpatch,\ymodb) {\includegraphics[width=\modsize cm, height=\modsize cm]{example-image}};
    \node (modbpatch1) [minimum width=\patchsize cm,minimum height=\patchsize cm, draw=none, inner sep=0pt] at (\xpatch-0.5,\ymodb+0.5) {\includegraphics[width=\patchsize cm]{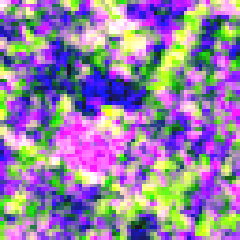}};
    \node (modbpatch2) [minimum width=\patchsize cm,minimum height=\patchsize cm, draw=none,inner sep=0pt] at (\xpatch+0.5,\ymodb+0.5) {\includegraphics[width=\patchsize cm]{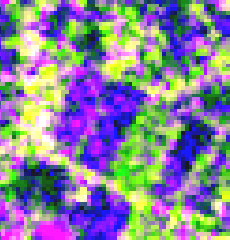}};
    \node (modbpatch3) [minimum width=\patchsize cm,minimum height=\patchsize cm, draw=none, inner sep=0pt] at (\xpatch-0.5,\ymodb-0.5) {\includegraphics[width=\patchsize cm]{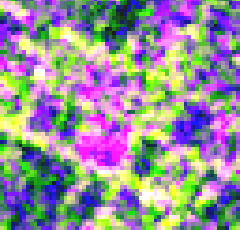}};
    \node (modbpatch4) [minimum width=\patchsize cm,minimum height=\patchsize cm, draw=none, inner sep=0pt] at (\xpatch+0.5,\ymodb-0.5) {\includegraphics[width=\patchsize cm]{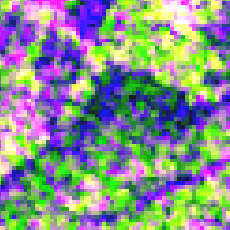}};
    \cube{\xpatch-0.5}{\ymodb+0.5}{\patchsize}{MODTWOCOLOR}
    \cube{\xpatch+0.5}{\ymodb-0.5}{\patchsize}{MODTWOCOLOR}
    \cube{\xpatch-0.5}{\ymodb-0.5}{\patchsize}{MODTWOCOLOR}
    \cube{\xpatch+0.5}{\ymodb+0.5}{\patchsize}{MODTWOCOLOR}

    \node (modcpatch1) [minimum width=\patchsize cm,minimum height=\patchsize cm, draw=none, inner sep=0pt] at (\xpatch-0.5,\ymodc+0.5) {\includegraphics[width=\patchsize cm]{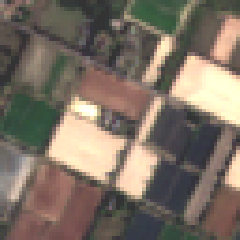}};
    \node (modcpatch2) [minimum width=\patchsize cm,minimum height=\patchsize cm, draw=none,inner sep=0pt] at (\xpatch+0.5,\ymodc+0.5) {\includegraphics[width=\patchsize cm]{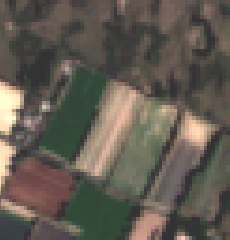}};
    \node (modcpatch3) [minimum width=\patchsize cm,minimum height=\patchsize cm, draw=none, inner sep=0pt] at (\xpatch-0.5,\ymodc-0.5) {\includegraphics[width=\patchsize cm]{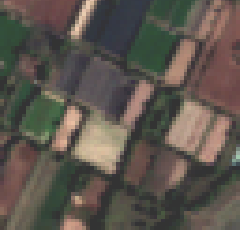}};
    \node (modcpatch4) [minimum width=\patchsize cm,minimum height=\patchsize cm, draw=none, inner sep=0pt] at (\xpatch+0.5,\ymodc-0.5) {\includegraphics[width=\patchsize cm]{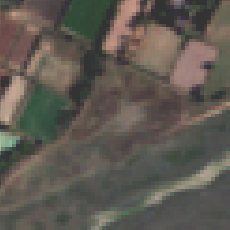}};
    \cube{\xpatch-0.5}{\ymodc+0.5}{\patchsize}{MODTHREECOLOR}
    \cube{\xpatch+0.5}{\ymodc-0.5}{\patchsize}{MODTHREECOLOR}
    \cube{\xpatch-0.5}{\ymodc-0.5}{\patchsize}{MODTHREECOLOR}
    \cube{\xpatch+0.5}{\ymodc+0.5}{\patchsize}{MODTHREECOLOR}

    \node[text width= 2cm] at (\xpatch,\ymodc -1.3) {\scriptsize Patches of size $P$};

    \node[isosceles triangle, isosceles triangle apex angle=90, minimum width=1.2cm,minimum height=0.5cm, draw, very thick,rotate=-0, fill=STUDENTCOLOR!50, anchor=west] (patchencoders) at (\xencoder, \ystudent) {$\PatchEncod_\Student$};
    
    \node[isosceles triangle, isosceles triangle apex angle=90, minimum width=1.2cm,minimum height=0.5cm, draw, very thick,rotate=-0, fill=TEACHERCOLOR!50, anchor=west] (patchencodert) at (\xencoder, \yteacher) {$\PatchEncod_\Teacher$};

    \node (token11s) [minimum width=0.4cm,minimum height=0.4cm, very thick, draw=black, fill=MODONECOLOR] at (\xtoken-0.6,\ystudent-0.9) {};
    \node (token21s) [minimum width=0.4cm,minimum height=0.4cm, very thick, draw=black, fill=MODONECOLOR] at (\xtoken-0.6,\ystudent-0.3) {};
    \node (token31s) [minimum width=0.4cm,minimum height=0.4cm, very thick, draw=black, fill=MODONECOLOR] at (\xtoken-0.6,\ystudent+0.3) {};
    \node (token41s) [minimum width=0.4cm,minimum height=0.4cm, very thick, draw=black, fill=MODONECOLOR] at (\xtoken-0.6,\ystudent+0.9) {};
    \node (token12s) [minimum width=0.4cm,minimum height=0.4cm, very thick, draw=black, fill=MODTWOCOLOR] at (\xtoken-0.0,\ystudent-0.9) {};
    \node (token22s) [minimum width=0.4cm,minimum height=0.4cm, very thick, draw=black, fill=MODTWOCOLOR] at (\xtoken-0.0,\ystudent-0.3) {};
    \node (token32s) [minimum width=0.4cm,minimum height=0.4cm, very thick, draw=black, fill=MODTWOCOLOR] at (\xtoken-0.0,\ystudent+0.3) {};
    \node (token42s) [minimum width=0.4cm,minimum height=0.4cm, very thick, draw=black, fill=MODTWOCOLOR] at (\xtoken-0.0,\ystudent+0.9) {};
    \node (token13s) [minimum width=0.4cm,minimum height=0.4cm, very thick, draw=black, fill=MODTHREECOLOR] at (\xtoken+0.6,\ystudent-0.9) {};
    \node (token23s) [minimum width=0.4cm,minimum height=0.4cm, very thick, draw=black, fill=MODTHREECOLOR] at (\xtoken+0.6,\ystudent-0.3) {};
    \node (token33s) [minimum width=0.4cm,minimum height=0.4cm, very thick, draw=black, fill=MODTHREECOLOR] at (\xtoken+0.6,\ystudent+0.3) {};
    \node (token43s) [minimum width=0.4cm,minimum height=0.4cm, very thick, draw=black, fill=MODTHREECOLOR] at (\xtoken+0.6,\ystudent+0.9) {};

    \draw[very thick] (token21s.north west) -- (token21s.south east);
    \draw[very thick] (token21s.north east) -- (token21s.south west);

     \draw[very thick] (token23s.north west) -- (token23s.south east);
    \draw[very thick] (token23s.north east) -- (token23s.south west);

    \draw[very thick] (token32s.north west) -- (token32s.south east);
    \draw[very thick] (token32s.north east) -- (token32s.south west);

    \draw[very thick] (token11s.west) -- (token13s.east);
    \draw[very thick] (token41s.west) -- (token43s.east);
    
    \node (token11t) [minimum width=0.4cm,minimum height=0.4cm, very thick, draw=black, fill=MODONECOLOR] at (\xtoken-0.6,\yteacher-0.9) {};
    \node (token21t) [minimum width=0.4cm,minimum height=0.4cm, very thick, draw=black, fill=MODONECOLOR] at (\xtoken-0.6,\yteacher-0.3) {};
    \node (token31t) [minimum width=0.4cm,minimum height=0.4cm, very thick, draw=black, fill=MODONECOLOR] at (\xtoken-0.6,\yteacher+0.3) {};
    \node (token41t) [minimum width=0.4cm,minimum height=0.4cm, very thick, draw=black, fill=MODONECOLOR] at (\xtoken-0.6,\yteacher+0.9) {};
    \node (token12t) [minimum width=0.4cm,minimum height=0.4cm, very thick, draw=black, fill=MODTWOCOLOR] at (\xtoken-0.0,\yteacher-0.9) {};
    \node (token22t) [minimum width=0.4cm,minimum height=0.4cm, very thick, draw=black, fill=MODTWOCOLOR] at (\xtoken-0.0,\yteacher-0.3) {};
    \node (token32t) [minimum width=0.4cm,minimum height=0.4cm, very thick, draw=black, fill=MODTWOCOLOR] at (\xtoken-0.0,\yteacher+0.3) {};
    \node (token42t) [minimum width=0.4cm,minimum height=0.4cm, very thick, draw=black, fill=MODTWOCOLOR] at (\xtoken-0.0,\yteacher+0.9) {};
    \node (token13t) [minimum width=0.4cm,minimum height=0.4cm, very thick, draw=black, fill=MODTHREECOLOR] at (\xtoken+0.6,\yteacher-0.9) {};
    \node (token23t) [minimum width=0.4cm,minimum height=0.4cm, very thick, draw=black, fill=MODTHREECOLOR] at (\xtoken+0.6,\yteacher-0.3) {};
    \node (token33t) [minimum width=0.4cm,minimum height=0.4cm, very thick, draw=black, fill=MODTHREECOLOR] at (\xtoken+0.6,\yteacher+0.3) {};
    \node (token43t) [minimum width=0.4cm,minimum height=0.4cm, very thick, draw=black, fill=MODTHREECOLOR] at (\xtoken+0.6,\yteacher+0.9) {};

    \node[isosceles triangle, isosceles triangle apex angle=90, minimum width=1.2cm,minimum height=0.5cm, draw, very thick,rotate=-0, fill=STUDENTCOLOR!50, anchor=west] (combiners) at (\xcombiner, \ystudent) {$\Combiner_\Student$};
    
    \node[isosceles triangle, isosceles triangle apex angle=90, minimum width=1.2cm,minimum height=0.5cm, draw, very thick,rotate=-0, fill=TEACHERCOLOR!50, anchor=west] (combinert) at (\xcombiner, \yteacher) {$\Combiner_\Teacher$};

    \node (token15sr) [minimum width=0.4cm,minimum height=0.4cm, very thick, draw=black, fill=MASKCOLOR, dots=15 per 1cm] at (\xtokenrec-0.0,\ystudent-0.9) {};
    \node (token25sr) [minimum width=0.4cm,minimum height=0.4cm, very thick, draw=black, fill=MODALLCOLOR, text=yellow] at (\xtokenrec-0.0,\ystudent-0.3) {$\star$};
    \node (token35sr) [minimum width=0.4cm,minimum height=0.4cm, very thick, draw=black, fill=MODALLCOLOR, text=yellow] at (\xtokenrec-0.0,\ystudent+0.3) {$\star$};
     \node (token85sr) [minimum width=0.4cm,minimum height=0.4cm, very thick, draw=black, fill=MASKCOLOR, dots=15 per 1cm,] at (\xtokenrec-0.0,\ystudent+0.9) {};

    \node (token15tr) [minimum width=0.4cm,minimum height=0.4cm, very thick, draw=black, fill=MODALLCOLOR, text=yellow] at (\xtokenrec-0.0,\yteacher-0.9) {$\star$};
    \node (token25tr) [minimum width=0.4cm,minimum height=0.4cm, very thick, draw=black, fill=MODALLCOLOR, text=yellow] at (\xtokenrec-0.0,\yteacher-0.3) {$\star$};
    \node (token35tr) [minimum width=0.4cm,minimum height=0.4cm, very thick, draw=black, fill=MODALLCOLOR, text=yellow] at (\xtokenrec-0.0,\yteacher+0.3) {$\star$};
    \node (token45tr) [minimum width=0.4cm,minimum height=0.4cm, very thick, draw=black, fill=MODALLCOLOR, text=yellow] at (\xtokenrec-0.0,\yteacher+0.9) {$\star$};

    \node[isosceles triangle, isosceles triangle apex angle=90, minimum width=1.2cm,minimum height=0.5cm, draw, very thick,rotate=-0, fill=STUDENTCOLOR!50, anchor=west] (predictor) at (\xpredictor, \ystudent) {$\phi^\text{pred}_\Student$};

    \node (token15srr) [minimum width=0.4cm,minimum height=0.4cm, very thick, draw=black, fill=MODALLCOLOR, text=yellow] at (\xtokenrecrec-0.0,\ystudent-0.9) {$\star$};
    \node (token25srr) [minimum width=0.4cm,minimum height=0.4cm, very thick, draw=black, fill=MODALLCOLOR, text=yellow] at (\xtokenrecrec-0.0,\ystudent-0.3) {$\star$};
    \node (token35srr) [minimum width=0.4cm,minimum height=0.4cm, very thick, draw=black, fill=MODALLCOLOR, text=yellow] at (\xtokenrecrec-0.0,\ystudent+0.3) {$\star$};
    \node (token45srr) [minimum width=0.4cm,minimum height=0.4cm, very thick, draw=black, fill=MODALLCOLOR, text=yellow] at (\xtokenrecrec-0.0,\ystudent+0.9) {$\star$};
    \node [draw=none] (lossjepa) at (\xtokenrecrec,0) {$\mathcal{L}_\text{JEPA}$};
    \node [draw=none] (losscon) at (\xtoken+1.5,4.25) {$\mathcal{L}_\text{con}$};

    \node [draw=none] (legend1) at (\xinput, \ystudent+0.4) {
    \begin{tabular}{r@{\;}l}
    \raisebox{0.5ex}{\begin{tikzpicture}[baseline=(current bounding box.center)]
       \node (token) [minimum width=0.2cm,minimum height=0.2cm,draw=black, thick] at (0.2,0) {}; 
     \draw[very thick] (0,0.0) -- (0.4,0.0);
     \end{tikzpicture}}
     & \raisebox{-0.25\height}{\footnotesize dropped patches} \\
    \begin{tikzpicture}[baseline=(current bounding box.center)]
        \node (token) [minimum width=0.2cm,minimum height=0.2cm,draw=black, thick] at (0,0) {};
        \draw[very thick] (token.north west) -- (token.south east);
        \draw[very thick] (token.north east) -- (token.south west);
    \end{tikzpicture}\; &  \raisebox{-0.25\height}{\footnotesize mask token $f^\text{mask}$} \\
    \begin{tikzpicture}[baseline=(current bounding box.center)]
        \node (empty) [minimum width=0.2cm,minimum height=0.2cm,draw=black, fill=MODALLCOLOR, text=yellow, thick] at (0,0) {};
        \node (empty) [draw=none, fill=none, text=yellow, text width=1mm] at (-0.025,0) {\small $\star$};
    \end{tikzpicture}\; & \raisebox{-0.25\height}{\footnotesize multimodal token} \\
    \begin{tikzpicture}[baseline=(current bounding box.center)]
        \node (empty) [minimum width=0.2cm,minimum height=0.2cm,draw=black, fill=MASKCOLOR, dots=20 per 1cm, thick] at (0,0) {};
    \end{tikzpicture}\; & \raisebox{-0.25\height}{\footnotesize drop token $f^\text{drop}$} 
    \end{tabular}
    };

    \node [draw=none] (legend2) at (\xinput,\yteacher-0.25) {
    \begin{tabular}{c}
    \begin{tikzpicture}
    \node[draw, circle, minimum size=\posencsize cm, very thick] (posenc2) at (0,0) {};
    \draw[domain=0:360,samples=100,smooth,variable=\x, thick] 
    plot({0+\posencsize/2+\x/360*\posencsize - \posencsize},{sin(\x)*(0.3*\posencsize)});
    \draw[domain=0:360,samples=100,smooth,variable=\x, thick] 
    plot({0+\posencsize/2+\x/360*\posencsize - \posencsize},{sin(2*\x)*(0.3*\posencsize)});
    \end{tikzpicture}
    \\
    \footnotesize  positional encoding
    \\
      \tikz \node [very thick, draw=black] at (0,0) {EMA};
      \\
     \footnotesize  exponential moving average 
    \end{tabular}
    };

    \draw [very thick,-] (geoplex) -- ($(modb.west) + (-6mm,0)$);
    \draw [very thick, ->] ($(modb.east) + (+4mm,0)$) -- ++ (\xshift,0) |- (patchencoders);
    \draw [very thick, ->] ($(modb.east) + (+4mm,0)$) -- ++ (\xshift,0) |- (patchencodert);
    \draw [very thick, ->] (token13s.east|-patchencoders) ++ (0.15,0) -- (combiners);
    \draw [very thick, ->] (token13t.east|-patchencodert) ++ (0.15,0) -- (combinert);
    \draw [very thick, ->] (token15sr.east|-patchencoders) ++ (0.15,0) -- (predictor);

    \draw [very thick, dashed, ->] (patchencoders) -- node[pos=0.5, yshift=-\pgflinewidth, fill=white, draw=black, rectangle, inner sep=3pt, solid] {EMA} ($(patchencodert.north) + (0,2mm)$);
   \draw [very thick,dashed, ->] (combiners) -- node[pos=0.5, yshift=-\pgflinewidth, fill=white, draw=black, rectangle, inner sep=3pt, solid] {EMA} ($(combinert.north) + (0,2mm)$);
    \draw [very thick, ->] (token15srr.south) -- (lossjepa);
    \draw [very thick, ->] (token15tr.east|-combinert) -| (lossjepa);
    \draw [very thick, ->]  (token42s.north) ++ (0,0.15) |- (losscon);

    \node[draw, circle, minimum size=\posencsize cm, very thick] (posenc1) at (\xtoken+.0,0) {};
    \draw[domain=0:360,samples=100,smooth,variable=\x, thick] 
    plot({\xtoken+.0+\posencsize/2+\x/360*\posencsize - \posencsize},{sin(\x)*(0.3*\posencsize)});
    \draw[domain=0:360,samples=100,smooth,variable=\x, thick] 
    plot({\xtoken+.0+\posencsize/2+\x/360*\posencsize - \posencsize},{sin(2*\x)*(0.3*\posencsize)});

    \node[draw, circle, minimum size=\posencsize cm, very thick] (posenc2) at (\xtokenrec,0) {};
    \draw[domain=0:360,samples=100,smooth,variable=\x, thick] 
    plot({\xtokenrec+\posencsize/2+\x/360*\posencsize - \posencsize},{sin(\x)*(0.3*\posencsize)});
    \draw[domain=0:360,samples=100,smooth,variable=\x, thick] 
    plot({\xtokenrec+\posencsize/2+\x/360*\posencsize - \posencsize},{sin(2*\x)*(0.3*\posencsize)});

    \draw[very thick, ->] (posenc1) -- ++ (0,1.2);
    \draw[very thick, ->] (posenc1) -- ++ (0,-1.2);
    \draw[very thick, ->] (posenc2) -- ++ (0,1.2);

    \node[draw=none, above =0mm of modapatch2.north]  {$x^m_p$};
    \node[draw=none, left =0mm of token41s.west]  {$f^m_{p,\Student}$};
    \node[draw=none, left =0mm of token85sr.west]  {$f^\star_{p,\Student}$};
    \node[draw=none, left =0mm of token45srr.west]  {$f^\star_{p,\text{pred}}$};
    \node[draw=none, left =0mm of token41t.west]  {$f^m_{p,\Teacher}$};
    \node[draw=none, left =0mm of token45tr.west]  {$f^\star_{p,\Teacher}$};
\end{tikzpicture}
    \vspace{-2mm}
    \caption{{\bf Architecture of AnySat.} 
 We begin each iteration by randomly selecting a dataset among GeoPlex and sampling a tile. Each available modality is divided into spatially aligned patches of size $P$. The student network's patch encoder $\PatchEncod_\Student$ embeds each patch and we apply a contrastive loss to encourage spatial consistency across modalities. We then apply dropping and masking : some patches have all modalities removed (dropping), while others have only random modalities removed (masking). The remaining patches are merged in the modality combiner $\Combiner_\Student$ to form multimodal representations $f^\star_\Student$ for the non-dropped patches. The predictor $\Predictor_\Student$ then reconstructs the embeddings of the dropped patches. Finally, the student network's output is compared to the teacher's, whose weights are an Exponential Moving Average (EMA) of the student's weights and which processes the complete set of patches without masking or dropping. 
}
\label{fig:training}
\end{figure*}

\paragraph{Modality-Combiner Network.} The combiner network $\phi^{\text{comb}}$ merges embeddings $f_p^m$ from all available modalities into a multimodal representation $f_p^\star$ for each patch $p \in \bP$. We use the cross-attention-based architecture proposed by OmniSat \cite[3.1]{astruc2024omnisat}: 
(i)  We first add to each $f^m_p$ an absolute positional encoding $\text{pos}(p)$---the same one used for sub-patches.;
(ii) The tokens go through a sequence of $B$ self-attention blocks;
(iii) We associate each patch with a token with a shared learned value and add positional encoding; and (iv) We compute the cross-attention between these tokens and the embeddings of the last self-attention block. This results in one embedding per patch $f^\star_p$.
\subsection{Training}
\label{sec:training}
We adapt the Joint Embedding Predictive Architecture (JEPA) framework~\cite{assran2023self} to multimodal Earth Observation, enabling self-supervised pretraining on datasets of varying modalities without labels. A student network operates on heavily masked inputs, aiming to predict embeddings generated by an unmasked teacher network whose parameters follow an Exponential Moving Average (EMA) of the student's weights~\cite{he2020momentum}. Training leverages two losses: a contrastive loss to enforce modality consistency and a JEPA loss for masked embedding prediction. 

The student network consists of a patch encoder $\PatchEncod_\Student$, a modality combiner $\Combiner_\Student$, and a predictor network $\Predictor_\Student$ with 3 self attention blocks. The teacher network includes a patch encoder $\PatchEncod_\Teacher$ and a modality combiner $\Combiner_\Teacher$ and no predictor.
The student network  first embeds all input tokens $x_p^m$ into vectors of size $E$ using the patch encoder: 
\begin{align} f_{p,\Student}^m = \PatchEncod_\Student(x_p^m)~. 
\end{align}

\paragraph{Contrastive Loss.} 
For a fixed patch $p$, the observations $x_p^m$ for $m \in \mathcal{M}$ capture different aspects of the same spatial region but share the same underlying semantics: the content of $p$. Therefore, we expect the representations $f_p^{m,\Student}$ to be consistent across modalities.
We enforce this intuition with a contrastive loss inspired by OmniSat~\cite{astruc2024omnisat}. Specifically, we use a modified InfoNCE loss~\cite{oord2018representation}, where each token $(p,m)$ is positively paired with those from the same spatial patch but different modalities: 
\begin{align}
     \mathcal{L}_\text{con} =
    &
    \!\!\!\!\!\!\!\!
    \sum_{\substack{~\\~\\(p,m) \in \bP \times \bM}}
    \!\!\!\!\!\!\!\!
    \frac{-\log}{\vert \bP \vert\vert \bM \vert} 
    \left(
    \frac
    {\displaystyle{\sum_{n \neq m}}
    \exp
    \left( 
        {
        \langle
        f_{p,\Student}^m,f_{p,\Student}^n
        \rangle}/
        {\tau}
    \right)}
    {\displaystyle{\sum_{\substack{n \neq m \\ q \neq p}}}
    \exp
    \left( 
    {\langle
    f_{p,\Student}^m,f_{q,\Student}^n
    \rangle}/
    {\tau}
    \right)}
    \right)~,
\end{align}
where $\tau$ is a temperature parameter, and $\langle \cdot, \cdot \rangle$ denotes the cosine similarity between embeddings.

\paragraph{Joint Embedding Predictive Architecture.} 
We adapt the JEPA self-supervised learning framework~\cite{assran2023self} to the context of multimodal Earth Observation. Avoiding reconstruction in pixel space is particularly beneficial for EO data, which can be heavily influenced by factors such as weather, time of day, or acquisition angle. Reconstructing in latent space allows us to learn more consistent and semantically meaningful features. The training process proceeds as follows:
\begin{compactitem}
    \item \textbf{Patch Dropping.} We apply JEPA's masking strategy by randomly selecting five rectangular regions on the tile. Let $\bDrop \subset \bP$ be the set of patches intersected by these rectangles, and $\bar{\bDrop} = \bP \setminus \bDrop$ the remaining patches. We drop all the student's tokens $f_{p,\Student}^m$ for patches $p \in \bDrop$. 
    \item \textbf{Modality \& Temporal Masking:} 
    We randomly mask a subset $\bMask \subset \bar{\bDrop} \times \bM$ of the remaining tokens, ensuring that at least one modality per patch remains unmasked. Masked token embeddings are replaced with a fixed value $f^{\text{mask}} \in \mathbb{R}^E$, which is learned as a parameter of the network. We also randomly mask $50$\% of the timestamps of all time series. 
    \item \textbf{Combiner:} We input all tokens (masked or not) to the student's combiner $\Combiner_\Student$, producing multimodal embeddings $f^{\star}_{p,\Student}$ for all $p \in \bar{\bDrop}$:
\begin{align}
        f^{\star}_{p,\Student} = \Combiner
        (
        \{f_{p,\Student}^m
        \}_{(p,m) \not\in \bMask} 
        \cup
        \{f^{\text{mask}}
        \}_{(p,m) \in \bMask}
        )~.
    \end{align}  
    \item \textbf{Predictor:} 
    We replace each dropped patch $p \in \bDrop$ with a fixed value $f^{\text{drop}} \in \mathbb{R}^E$. We add positional encodings to all tokens (including the dropped ones) and input them to the predictor $\Predictor_\Student$, yielding embeddings $f^{\star}_{p,\text{pred}}$ for all patches $p \in \bP$:
    \begin{align}
        f^{\star}_{p,\text{pred}} = \Predictor_\Student
        (
        \{ f^{\star}_{p,\Student}
        \}_{p \in \bar{\bDrop}} 
        \cup
        \{ f^{\text{drop}}
        \}_{p \in \bDrop}
        )~.
    \end{align}
    \item \textbf{Teacher Encoding:} The teacher network receives all input tokens $x_p^m$, embeds them using $\PatchEncod_\Teacher$, and combines them with $\Combiner_\Teacher$ without any dropping, masking, or temporal dropout. The teacher outputs patch embeddings $f^{\star}_{p,\Teacher}$ for all $p \in\bP$.

    \item \textbf{Loss Function:} The training objective is the $L_2$ distance between the  student predictions and the teacher's multimodal embeddings for the dropped patches:
    \begin{align}
    \mathcal{L}_\text{JEPA} = \frac1{\vert \bK \vert}\sum_{p \in \bDrop} \left\| f^{\star}_{p,\text{pred}} - f^{\star}_{p,\Teacher} \right\|_2^2~.
    \end{align}
\end{compactitem}

After training, we use the teacher network for downstream tasks and discard the student. Note that all modules are shared across all modalities except for the projection layers $\ProjEncod_m$ in the patch encoder $\PatchEncod$.

\paragraph{Training with Multiple Datasets.} The flexibility of AnySat enables us to train a single model simultaneously on several datasets of various sizes and scales with the same weights and without rescaling. We consider a set $\bD$ of such datasets. Each dataset $d \in \bD$ is characterized by the subset $M_d \subset \bM$ of its available modalities and $S_d$ the size of its tiles. We also consider a batch size $B_d$ and a set $P_d$ of acceptable patch sizes, which depend on the nature of the data, the available resolution, and the tile size.
We use the following procedure:
\begin{compactitem}
    \item[1.] Randomly select a dataset $d$ in $\bD$.
    \item[2.] Randomly select a patch size $P$ in $P_d$.
    \item[3.] Randomly sample $B_d$ tiles in $d$.
    \item[4.] Process the tiles and backpropagate the loss.
\end{compactitem}

\begin{figure}[t]
    \centering
    \input{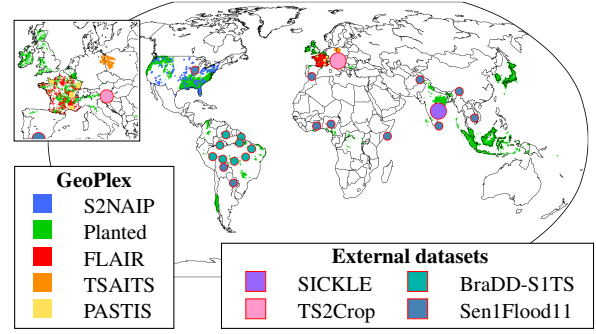}
    \vspace{-1mm}
    \caption{{\bf Datasets Considered.}
    GeoPlex is composed of $5$ diverse dataset spanning the entire world,  with a higher concentration in Europe and the US where open-data are more abundant. We also consider external evaluation datasets with a more diverse spread.}
    \label{fig:geoplex}
\end{figure}

\begin{figure*}[t]
    \centering
\resizebox{\textwidth}{!}{%
        \input{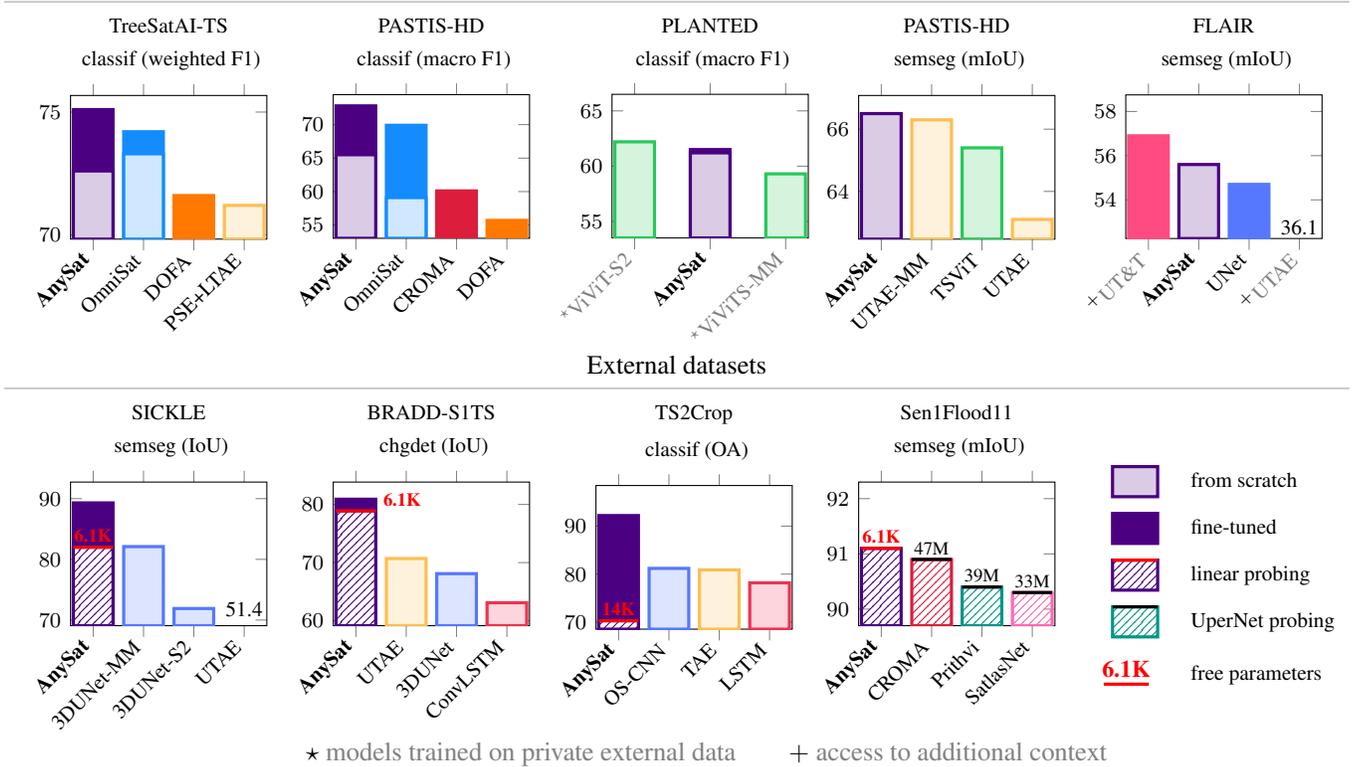}
    }
\vspace{-1mm}
\caption{{\bf Quantitative Evaluation.} We evaluate AnySat across 9 open-access datasets and for four tasks: multilabel classification (classif), semantic segmentation (semseg), pixel-wise change detection (chgdet), and pixel-wise regression (regression). For clarity, we only visualize the four best performance per dataset, see Appendix for full results. We report the number of trainable parameters for probing evaluations.}
\label{fig:barplots}
\end{figure*}

\subsection{Downstream Tasks}
\label{sec:downstream}
After pretraining, AnySat can be fine-tuned or probed for various downstream tasks, including classification and semantic segmentation.

\paragraph{Classification.} For tile-level classification, we insert a \texttt{[CLS]} token into the combiner network's cross-attention module. This token generates a tile-level embedding, subsequently mapped to label logits through a linear classifier.

\paragraph{Semantic Segmentation.}
For semantic segmentation, we predict labels at pixel-level resolution by first selecting a modality whose resolution is close to the annotation resolution. A dense feature map at the sub-patch scale ($\delta_m$) is formed by concatenating sub-patch embeddings (outputs of $\ProjEncod_m$) with corresponding multimodal patch embeddings (outputs of $\Combiner$). An MLP then maps these concatenated embeddings to logits of dimension $\delta_m \times \delta_m \times N$, where $N$ is the number of semantic classes. Unfolding these logits yields pixel-level predictions. Using sub-patches results in higher-resolution predictions compared to methods that rely only on patch-level representations.

\paragraph{Probing.} AnySat supports linear probing, where a simple linear classifier can be attached directly to the class token for classification or to the dense feature maps for segmentation. This approach avoids complex segmentation heads typically required in earlier methods~\cite{marsocci2024cross}, leveraging the dense features produced by our architecture.

\paragraph{New Sensor Configurations.} 
AnySat can adapt to sensors with configurations differing from those in the training datasets well as new sensors. During self-supervised pretraining, we learn a sensor-specific scalar value representing missing data, which is subsequently used wherever modality channels are absent during fine-tuning or probing. For sensors not featured in the training sensors, we randomly initialize a new projector and fine-tune it along the other free parameters. This effectively extends AnySat to previously unseen sensors but cannot be used in probing for sensors too different from the training mix.

\section{Experiments}

\subsection{Datasets and Evaluation}
\label{sec:datasets}

We present the datasets used for training and evaluation, as well as our evaluation protocol.

\paragraph{GeoPlex.} 
As argued by Roscher \etal \cite{roscher_better_2024}, EO models benefit from high-quality, diverse, and curated data rather than extensive but uniform acquisitions. We follow this principle by compiling a collection of five multimodal datasets, each featuring different combinations of modalities, scales, and resolutions. GeoPlex comprises the training sets of the following datasets:
\begin{compactitem}

\item \textbf{TreeSatAI-TS} \cite{ahlswede2022treesatai, astruc2024omnisat}: A forest-centric dataset in Germany with Sentinel-1 \& 2 time series and Very High Resolution (VHR) images at 0.2 m resolution.

\item \textbf{FLAIR} \cite{garioud2023flair}: A French land cover dataset with Sentinel-2 time series and VHR images with elevation data (0.2~m). To form multimodal patches, we crop the Sentinel-2 time series to match the extent of the VHR images (discarding $93.5$\% of pixels).

\item \textbf{PLANTED} \cite{pazos2024planted}: A global forest dataset comprising time series from multiple sensors, including Sentinel-1/2, Landsat-7, ALOS-2, and MODIS. Only $1.3$ of the $2.3$M images used in the paper are publicly available.

\item \textbf{S2NAIP-URBAN} \cite{bastani2023satlaspretrain}: An urban dataset in the continental US with VHR images (1.25m) and time series from Sentinel-1/2 and Landsat-8/9.

\item \textbf{PASTIS-HD} \cite{garnot2021panoptic, astruc2024omnisat}: A French crop mapping dataset with VHR images (1.5m) and Sentinel-1 \& 2 time series. As PASTIS is evaluated in 5-fold cross-validation, there are no dedicated train and test sets. We include the entire dataset (without labels) in GeoPlex.

\end{compactitem}
~\\\noindent
As illustrated in \cref{fig:geoplex}, GeoPlex spans 249K km\textsuperscript{2} across five continents and 171 billion pixels. The sampled tiles range in size from $0.36$ to $164$ hectares. GeoPlex includes $11$ distinct modalities with resolutions ranging from $0.2$ m to $250$ m, with both VHR images and time series data:
\begin{compactitem} 
\item \textbf{Very High Resolution  Images:} 
\begin{compactitem} 
\item \textbf{Aerial:} RGB+NIR (near-infrared) at 0.2 m
\item \textbf{Aerial+NMS:} RGB+NIR+Elevation at 0.2 m  
\item \textbf{NAIP:} RGB+NIR at 1.25 m 
\item \textbf{SPOT6:} RGB+NIR at 1.5 m.
\end{compactitem} 
\item \textbf{Time Series Data:} \begin{compactitem}
\item \textbf{Sentinel-1:} 3 channels (VV/VH polarization + ratio) at 10 m,  Ascending \& Descending Orbits 
\item \textbf{Sentinel-2:} 10 channels at 10 m  
\item \textbf{ALOS-2:} 3 channels (polarization) at 30 m
\item \textbf{Landsat-7:} 6 channels at 30 m
\item \textbf{Landsat-8/9:} 11 channels at 30 m
\item \textbf{MODIS:} 7 channels at 250 m. 
\end{compactitem} 
\end{compactitem}
~\\\noindent
We select the possible patch size per dataset, while we set the sub-patch size per modality $1$ pixel for very high-resolution images and $10$  pixels for time series data.
See the Appendix for the complete characteristics of all datasets.

\paragraph{External Datasets.} 
To showcase AnySat's flexibility, we also consider $6$ datasets not included in GeoPlex. AnySat can be directly fine-tuned or linearly probed on new datasets, even if their modality combination is not featured in GeoPlex. We consider the following datasets:
\begin{compactitem}

    \item {\bf SICKLE} \cite{sani2024sickle}: A multimodal crop mapping dataset in India featuring Sentinel-1, Sentinel-2, and Landsat-8 time series. As the test set has not been released, we use the validation set.

    \item \textbf{BraDD-S1TS} \cite{karaman2023deforestation}: A change detection dataset comprising Sentinel-1 time series of the Amazon rainforest, aiming to segment deforested areas.

    \item \textbf{TimeSen2Crop} \cite{weikmann2021timesen2crop}: A crop mapping dataset in Slovenia consisting of \emph{single-pixel} Sentinel-2 time series, a modality not present in GeoPlex.

    \item \textbf{Sen11Flood1} \cite{bonafilia2020sen1floods11}: A global flood mapping dataset with pixel annotations and single-date Sentinel-1 and 2 observations, a configuration not present in GeoPlex. Each tile covers $2600$ hectares.

    \item \textbf{So2Sat} \cite{zhu2019so2sat}: A local climate zone classification dataset containing co-registered Sentinel-1 and Sentinel-2 imagery across multiple cities worldwide, with single-date observations---a configuration not present in GeoPlex.

    {\item \textbf{HLS Burn Scar} \cite{HLS_Foundation_2023}: A dataset for burn scar detection using Harmonized Landsat-Sentinel (HLS) imagery, featuring time series data to identify post-fire affected areas and large tiles of 24K hectares.}

\end{compactitem}

\paragraph{Evaluation.} We evaluate our model on the annotated datasets of GeoPlex (excluding S2NAIP-URBAN) and the {$6$} external datasets across three tasks: 
(i) {\bf Classification:} 
TSAIT-TS, PASTIS-HD, PLANTED, TimseSenCrop, {So2Sat};
(ii)  {\bf Semantic Segmentation:} PASTIS-HD, FLAIR, SICKLE, Sen1Flood11, {HLS Burn Scars};
and (iii) 
 {\bf Binary pixel-wise change detection:} BraDD-S1TS.

We use three evaluation settings to evaluate the models:  
\begin{compactitem}
    \item {\bf From Scratch.}  The model is trained directly on the labeled training set in a supervised manner.
    \item {\bf Fine-tuning.} The model is pretrained in a self-supervised manner, then fine-tuned on the training set.
    \item {\bf Linear Probing.} The model is initially pretrained in a self-supervised manner, and a linear layer is fitted with the training set.
\end{compactitem}

\paragraph{Competing Methods.} {We compare AnySat against state-of-the-art Earth Observation models.
Most foundation models pre-trained on external data cannot be directly applied to target datasets with different input configurations. For example, the ScaleMAE and SatMAE models are trained on the Functional Map of the World \cite{christie2018functional} and limited to RGB bands, while CROMA is trained on single-date Sentinel-2 data. Since these specific modalities are not present in any of our evaluation datasets, we cannot directly evaluate these pretrained models. Instead, we modify the input layers of these models to match the target number of spectral bands.

\begin{table*}[t]
\centering

\resizebox{1\linewidth}{!}{
\centering
\begin{tabular}{l cc cc cc cc c cc}
\toprule
\multirow{3}{*}{Models} 
& \multicolumn{2}{c}{\multirow{2}{*}{free param.}} 
& \multicolumn{2}{c}{\colorbox{green!50}{HLS Burns (mIoU)}} 
& \multicolumn{2}{c}{\colorbox{green!50}{Spacenet7 (mIoU)}} 
& \multicolumn{2}{c}{\colorbox{cyan!50}{Sen1Floods11 (mIoU)}} 
& \multicolumn{1}{c}{\colorbox{cyan!50}{So2Sat (OA)}} 
& \multicolumn{2}{c}{PASTIS-HD Fold1 (mIoU)}
\\
&&& \multicolumn{2}{c}{\colorbox{green!50}{HLS}} 
 & \multicolumn{2}{c}{\colorbox{green!50}{Planet}} 
 & \multicolumn{2}{c}{\colorbox{cyan!50}{S1\mono+S2\mono}} 
 & \multicolumn{1}{c}{\colorbox{cyan!50}{S1\mono+S2\mono}}
 & \multicolumn{2}{c}{S1+S2+VHR}
\\
& FT & Prob. 
& FT & Prob.
& FT & Prob.
& FT & Prob.
& Prob.
& FT & Prob.
\\
\cmidrule(lr){2-3}\cmidrule(lr){4-5}\cmidrule(lr){6-7}\cmidrule(lr){8-9}\cmidrule(lr){10-10}\cmidrule(lr){11-12}
\textbf{AnySat (ours)} 
& \underline{125M} & \bf 6K-3M 
& \bf 90.6 & \bf 87.7$^\bullet$    
& \bf 58.8 & 58.1$^\bullet$      
& -    & \bf 91.1$^\star$    
& \underline{59.1$^\star$}          
& \bf 65.9 & \bf 42.7$^\dagger$    
\\

Prihtvi2 \cite{szwarcman2024prithvi}
& 630M & -
& \underline{90.5} & -  
& - & -               
& 89.9 & -           
& -                  
& - & -              
\\

Prihtvi \cite{jakubik2310foundation}
& 130M & 39M
& 86.9 & \underline{83.6}    
& - & 56.5       
& \underline{90.4} & 88.3    
& -              
& - & 33.9       
\\

CROMA \cite{fuller2023croma}
& 350M & 47M
& - & 82.4             
& - & 59.3             
& \bf 90.9 & \underline{90.9}
& 49.2$^\star$         
& - & 32.3             
\\

SatMAE \cite{cong2022satmae}
& 304M & 31M
& - & -             
& - & -             
& - & -             
& 46.9$^\star$         
& - & -             
\\

ScaleMAE \cite{reed2023scale}
& 350M & 47M
& - & 76.7             
& 54.0 & \bf 63.0          
& - & 74.1             
& -                    
& - & 24.6             
\\

DOFA \cite{xiong2024dofa}
& 151M & 39M
& - & 80.6             
& - & 61.8             
& - & 89.4             
& \bf 59.3$^\star$         
& - & 30.0             

\\
RemoteCLIP \cite{liu2024remoteclip}
& 168M & 39M
& 65.1 & 76.6             
& - & 57.8             
& - & 74.3             
& -         
& - & 18.2             
\\
DINO-S12 \cite{oquab2023dinov2}
& \bf 54M & 31M
& - & 81.7             
& \underline{56.6} & 56.5             
& - & 88.6             
& -         
& \underline{44.0} & \underline{36.2}             
\\
SatlasNet \cite{irvin2023usat}
& 111M & 33M
& - & 80.0             
& - & \underline{61.9}             
& - & 90.3             
& -         
& 29.1 & 17.5             
\\\bottomrule
\end{tabular}
}
\caption{{\bf Comparison to Foundation Models.} We evaluate AnySat fine-tuned or linearly probed on datasets taken from the GeoBench \cite{lacoste2023geo} and PANGEA benchmarks \cite{marsocci2024pangaea}. \colorbox{green!50}{Unseen sensor}, \colorbox{cyan!50}{unseen sensor configuration},  \mono: single date, {\bf FT}: fine-tuning, {\bf Prob.}: linear or UperNet probing. $\star/\dagger/\bullet$: Probing requires 6K/33K/3M free parameters. The best performance is in {\bf bold} whereas the second one is  \underline{underlined}.}
\label{tab:rebuttal}
\end{table*}
\subsection{Results and Analysis}
We evaluate our model on different datasets from and outside of GeoPlex with fine-tuning and linear probing.

\paragraph{Performance on GeoPlex' Test Sets.}
We evaluate AnySat on the test sets of the GeoPlex datasets, as shown in \cref{fig:barplots}, with detailed results provided in the Appendix. Despite using a single pretrained model, AnySat sets new state-of-the-art results for TreeSatAI-TS ($+0.9$ weighted F1 score) and PASTIS-HD ($+2.8$ mIoU in classification and $+0.2$ in segmentation). AnySat also achieves near state-of-the-art performance on PLANTED \cite{pazos2024planted}, even though the ViViT models \cite{arnab2021vivit} were trained on a withheld dataset with nearly 80\% more data of the same type. Similarly, our model performs close to the state-of-the-art on FLAIR, despite having access to only 6.5\% of the extent of the Sentinel-2 tiles used by  UT\&T \cite{garioud2023flair}.

Pretraining on GeoPlex consistently improves performance, indicating that training on a collection of datasets with varied modalities leads to richer and more robust representations. The improvement is more pronounced for smaller datasets like TreeSatAI-TS and in classification tasks rather than segmentation. We attribute this to the amount of supervision available in larger datasets and dense annotations, which make pretraining less beneficial.

\paragraph{Performance on External Datasets.} 
\cref{fig:barplots} shows that AnySat significantly outperforms the state-of-the-art for 6 external datasets, improving SICKLE by +3.6 mIoU, BraDD-S1TS by +10.2 mIoU, and TimeSen2Crop by +11.0 OA. These gains highlight AnySat's strong spatial generalization as GeoPlex primarily covers the northern hemisphere, while the external datasets have global coverage.
We also evaluate AnySat on datasets from GeoBench \cite{lacoste2023geo} and PANGAEA-Bench \cite{marsocci2024pangaea}, see \cref{tab:rebuttal}, AnySat performs on par or better than all foundation models across all benchmarks. Note that these methods use resizing (on PASTIS-HD and  TreeSatAI) and sliding window  aggregation (on SN7) instead of our proposed adaptive patch encoding.

Moreover, AnySat can be effectively linearly probed for semantic segmentation. It surpasses all specialized approaches on BraDD-S1TS when linearly probed, and likewise exceeds the performance of foundation models with fine-tuned UperNet segmentation heads on Sen1Flood11. Notably, a linearly probed AnySat outperforms a fine-tuned Prithvi2 \cite{szwarcman2024prithvi} on Sen1Floods11 with $10^5$ fewer free parameters. These findings underscore the expressive power of AnySat’s self-supervised features and confirm that it can be adapted to new tasks and datasets at minimal training cost and still deliver competitive performance.

\paragraph{Performance on New Sensor Configurations.}
We demonstrate AnySat's robustness in handling sensor configurations not present in GeoPlex.
For instance, SICKLE's LandSat8 requires three additional bands beyond those used in S2NAIP's LandSat8, while TimeSen2Crop provides only 9 of the 10 bands employed by our Sentinel-2 projector network.
Applying the padding strategy described in \cref{sec:downstream}, AnySat achieves state-of-the-art results on both datasets.
We also evaluate AnySat on single-date Sentinel images (So2Sat, Sen1Flood11) and single-pixel time series (TimeSen2Crop), which were never part of GeoPlex, and again observe state-of-the-art performance.
Finally, we test AnySat on the HLSBurnScar dataset \cite{HLS_Foundation_2023}. As GeoPlex does not contain HLS data (but contains Sentinel and LandSat), we train a new projector for this new modality. AnySat outperforms all competing methods, including Prithvi \cite{jakubik2310foundation}, which was trained on 252M km\textsuperscript{2} of HLS imagery. In comparison, GeoPlex comprises only 249K km\textsuperscript{2} without any HLS data, further illustrating the strong generalization capability of AnySat.


\paragraph{Ablation Study.} We evaluate the impact of several key design choices and report the results in \cref{tab:ablation}. All results are presented for the Fold~5 of PASTIS-HD and for the classification and semantic segmentation tasks. We do not pretrain on the entire GeoPlex but use Fold~1 to 4 of PASTIS-HD in a self-supervised fashion.
\begin{compactitem} 
\item \textbf{Random Token Dropping.} We replaced JEPA's block masking strategy with purely random token dropping for the student network. This modification decreased classification performance but slightly improved segmentation results. In order to use a single model configuration for all tasks, we maintained a unified approach. Interestingly, block masking does not appear to be as critical for EO data than for natural images (see Table~6 in \cite{assran2023self}).

\item \textbf{No Contrastive Loss.} We remove the contrastive loss and retain only the reconstruction loss $\mathcal{L}_\text{JEPA}$. This substantially reduces the classification performance ($-4.3$ F1) but only a moderate decrease in segmentation performance ($-0.2$ mIoU). These findings suggest that the contrastive loss can help the feature-predictive approach learn more discriminative features, particularly benefiting classification tasks.

\item \textbf{Naive Semantic Segmentation.} We predict pixel-wise logits directly from the patch embeddings without utilizing subpatch features. This results in a decrease in segmentation performance by $2.4$ mIoU, highlighting the importance of subpatches in providing fine-grained spatial information.

\end{compactitem}

\begin{table}[t]
    \centering
    \caption{{\bf Ablation.} We evaluate the impact for several critical design choices of our model on the Fold~1 of PASTIS-HD.}
      \label{tab:ablation} 
      \vspace{-0mm}
      \begin{tabular}{lcc}
\multirow{2}{*}{Experiment} & classification  & segmentation  \\
& macro F1 & mIoU \\
\midrule
best configuration & \bf 72.0 & 63.6 \\\greyrule 
random token dropping & 71.3 & \bf 64.1 \\
no contrastive & 67.7 & 63.4 \\
naive semseg & - & 61.2 \\
\end{tabular}

\end{table}

\paragraph{Inference and Training Times.} 
Our model was pretrained on GeoPlex using 1760 GPU-hours on an NVIDIA H100 GPU. Fine-tuning takes between 10 and 40 hours, depending on the dataset size. Linear probing takes approximately 2 hours on BraDD-S1TS.

In terms of inference speed, AnySat processes one mono-date tile from TreeSatAI \cite{ahlswede2022treesatai} in 3ms on average, which is faster than ScaleMAE \cite{reed2023scale} ($10$ms) and comparable to DOFA \cite{xiong2024dofa} (3ms) and OmniSat \cite{astruc2024omnisat} (2ms).

\section{Conclusion}

We have presented AnySat, a versatile architecture designed to address the diversity of EO data in terms of resolutions, scales, and modalities. By leveraging a joint embedding predictive architecture and scale-adaptive spatial encoders, AnySat can be trained in a self-supervised manner on highly heterogeneous datasets. Pretrained on GeoPlex, a comprehensive collection of multimodal datasets with varying characteristics, our model achieved state-of-the-art performance across multiple datasets, tasks, and modalities.

A key advantage of AnySat is its ability to be applied and fine-tuned on a wide array of combinations of data types and scales with a single model. Moreover, new datasets can be easily incorporated into GeoPlex for self-supervised pretraining. Our goal is to generalize this approach to develop a versatile foundation model for environmental monitoring on a global scale.

\section*{Acknowledgement}
This work was granted access to the HPC resources of IDRIS under the allocations AD011014719 and AD011014286R1 made by GENCI.
We thank Jordi Inglada, Antoine Labatie, Dimitris Samaras, Yohann Perron, Vincent Lepetit for inspiring discussions and valuable feedback.

\FloatBarrier
\balance{
\bibliographystyle{ieeenat_fullname}
\bibliography{mybib}
}

\pagebreak

\maketitlesupplementary

\renewcommand\thefigure{\Alph{figure}}
\renewcommand\thesection{\Alph{section}}
\renewcommand\thetable{\Alph{table}}
\renewcommand\theequation{\Alph{equation}}
\setcounter{equation}{0}
\setcounter{section}{0}
\setcounter{figure}{0}
\setcounter{table}{0}

In this appendix, we provide detailed results in \cref{sec:results}, an extended ablation study in \cref{sec:ablation}, and provide implementation details in \cref{sec:details}.  Finally, we provide more details on the datasets and experiments of the main paper in \cref{sec:datasets}

\section{Detailed Results}
\label{sec:results}

We provide qualitative illustrations of our predictions and detailed quantitative results for the test sets of GeoPlex.

\paragraph{Qualitative Results.} 
We present qualitative illustrations in  \cref{fig:quali} for four segmentation tasks: PASTIS, FLAIR, SICKLE, and BraDD-S1TS. AnySat predicts precise segmentations that closely follow the extents of buildings, trees, and parcels. Notably, the predictions do not display grid artifacts despite our segmentation head being a simple linear layer applied to each subpatch. This suggests that using subpatches of small sizes (\eg, $4 \times 4$ pixels for PASTIS and $10 \times 10$ pixels for FLAIR), combined with larger context through patch embeddings, is an effective strategy for producing smooth and consistent segmentation maps.

\begin{figure}
    \centering
    \resizebox{\linewidth}{!}{
    \def\ntasks{12} 
\def\tasknames{{"TSatAI-TS", "PASTIS-HD", "PASTIS-HD", "PASTIS-HD", "PLANTED",  "FLAIR", "\underline{SICKLE}", "\underline{BRADD-SITS}", "\underline{TimeSen2Crop}", "\underline{Sen1Floods11}", "\underline{So2Sat}", "\underline{HLS Burn Scar}"}}
\def\specnames{{"classif (wF1)", "classif (mF1)",  "semseg (mIoU)", "classif (wF1)", "semseg (mIoU)", "semseg LP (mIoU)", "semseg (IoU)", "chgdet (IoU)", "classif (OA)", "semseg LP (mIoU)", "classif LP (OA)", "semseg LP (mIoU)"}}
\def\rangemin{{72,65,60,30,55,50,70,60,70, 90,50,80}}
\def\rangemax{{76,75,70,50,65,60,90,90,100,92,65,90}}

\def\nmethods{2} 
\def\methodsname{{"SOTA", "AnySat", "Method C", "Method D", "Method E"}}
\def\methodshade{{1,0,0,0,0,1}}
\def\AlphList{{"A","B","C","D","E"}} 
\def\performanceA{{74.2, 69.9, 66.3, 36.2, 62.2, 56.9, 82.1, 70.7, 81.2, 90.9,59.3,83.6}}
\def\performanceB{{75.1, 72.8, 66.5, 42.7, 61.5, 55.6, 89.3, 80.9, 92.2, 91.1, 59.1,87.7}}

\definecolor{colorA}{RGB}{255, 190, 80}
\definecolor{colorB}{RGB}{75, 0, 130} 
\definecolor{colorC}{RGB}{85, 120, 255}
\definecolor{colorD}{RGB}{240, 50, 80} 
\definecolor{colorE}{RGB}{0, 150, 136} 

\tikzset{titlenode/.style={font=\footnotesize}}

\newcommand{\DO}[1]{\edef\temp{\noexpand #1}\temp} 

\pgfmathsetmacro{\stepangle}{360/\ntasks} 
\pgfmathsetmacro{\ntasksminusone}{\ntasks-1}
\pgfmathsetmacro{\nmethodsminusone}{\nmethods-1}

\begin{tikzpicture}[every text node part/.style={align=center}]
\path[use as bounding box] (0.09 * \linewidth,-0.05 * \linewidth) rectangle (.91*\linewidth,.93*\linewidth);

\begin{polaraxis}[
    ymin=0,
    ymax=100,
    xtick={0, \stepangle, ..., 360}, 
    xticklabels={},
    ytick={25,50,75},
    yticklabels={,,},
    grid=both,
    line width=1pt,
    width=1\linewidth,
    height=1\linewidth,
    clip mode=individual,
    grid=both,            
    axis line style={draw=none},
    grid style={line width=0.5pt, draw=gray}, 
    major grid style={line width=0.5pt, draw=gray}, 
    extra y ticks={100}, 
    extra  y tick labels={},
    extra y tick style={
        grid=major,
        major grid style={black, line width=2pt}, 
            },
    clip=false,
    legend style={
        at={(-0.2,1.13)},
        anchor=north west, 
        font=\footnotesize,
    },
    legend cell align={left},
]

\foreach \taskindex in {0,...,\ntasksminusone} {
    \pgfmathsetmacro{\angle}{\taskindex * \stepangle}
    \pgfmathsetmacro{\titleangleA}{\angle-90}
    \pgfmathsetmacro{\titleangleB}{ifthenelse(\titleangleA>90,\titleangleA-180,\titleangleA)}

    \pgfmathsetmacro{\postask}{ifthenelse(\titleangleA>90,110,120)}
    \pgfmathsetmacro{\posspec}{ifthenelse(\titleangleA>90,120,110)}
    \pgfmathsetmacro{\taskname}{\tasknames[\taskindex]}
    \DO{\node[titlenode] at (\angle:\postask){\rotatebox{\titleangleB}{\taskname}};}

    \pgfmathsetmacro{\specname}{\specnames[\taskindex]}
    \DO{\node[titlenode] at (\angle:\posspec){\rotatebox{\titleangleB}{\specname}};}

    \pgfmathsetmacro{\fiftytik}{round(0.5*(\rangemax[\taskindex]+\rangemin[\taskindex]))}
    \pgfmathsetmacro{\hundredtick}{round(\rangemax[\taskindex])}

    \DO{\node[titlenode] at (\angle:45) {\rotatebox{\titleangleB}{\pgfmathprintnumber[fixed, precision=0]{\fiftytik}}};}
    \DO{\node[titlenode] at (\angle:95) {\rotatebox{\titleangleB}{\pgfmathprintnumber[fixed, precision=0]{\hundredtick}}};}  
}

\foreach \methodindex in {0,...,\nmethodsminusone} {
 No Modality or Temporal Masking.
    \pgfmathsetmacro{\methodletter}{\AlphList[\methodindex]}
    \expandafter\def\expandafter\performance\expandafter{\csname performance\methodletter\endcsname}
    \edef\methodcolor{color\methodletter}
    \pgfmathsetmacro{\methodshading}{\methodshade[\methodindex]}
    \pgfmathsetmacro{\methodname}{\methodsname[\methodindex]}

    \DO{\addplot[draw=\methodcolor] coordinates {(0,0)};}
    \DO{\addlegendentry{\methodname}}
    \DO{\node[] (lastnode) at (0,0) {};}
    
    \foreach \taskindex in {0,...,\ntasksminusone} {
        \pgfmathsetmacro{\angle}{\taskindex * \stepangle}
        \pgfmathsetmacro{\perf}{\performance[\taskindex]}
        \pgfmathsetmacro{\perfscaled}{100*(\perf-\rangemin[\taskindex])/(\rangemax[\taskindex]-\rangemin[\taskindex])}

        \DO{\node[] (nextnode) at (\angle:\perfscaled) {};}

        \ifnum\taskindex=0
            \DO{\node[] (firstnode) at (\angle:\perfscaled) {};}
        \else
            \ifnum\methodshading=1
                \DO{\fill[fill=\methodcolor, opacity=0.2]  (lastnode.center) -- (nextnode.center) -- (0,0) -- cycle;}
            \fi
            \DO{\draw[draw=\methodcolor]  (lastnode.center) -- (nextnode.center);}
           
        \fi
    
        \ifnum\taskindex=\ntasksminusone
            \ifnum\methodshading=1
                \DO{\fill[fill=\methodcolor, opacity=0.2]  (nextnode.center) -- (firstnode.center) -- (0,0) -- cycle;}
            \fi
            \DO{\draw[draw=\methodcolor]  (nextnode.center) -- (firstnode.center);}
        \fi
        
        \DO{\node[] (lastnode) at (\angle:\perfscaled) {};}
    }   
}

\end{polaraxis}

\end{tikzpicture}
    }
    \vspace{1mm}
    \caption{{\bf Overall Performance.} We underline external datasets. LP stands for Linear Probing.}
    \label{fig:enter-label}
\end{figure}
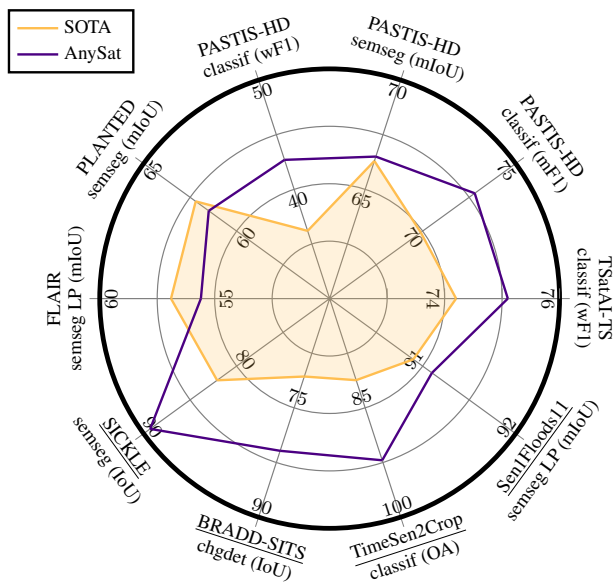


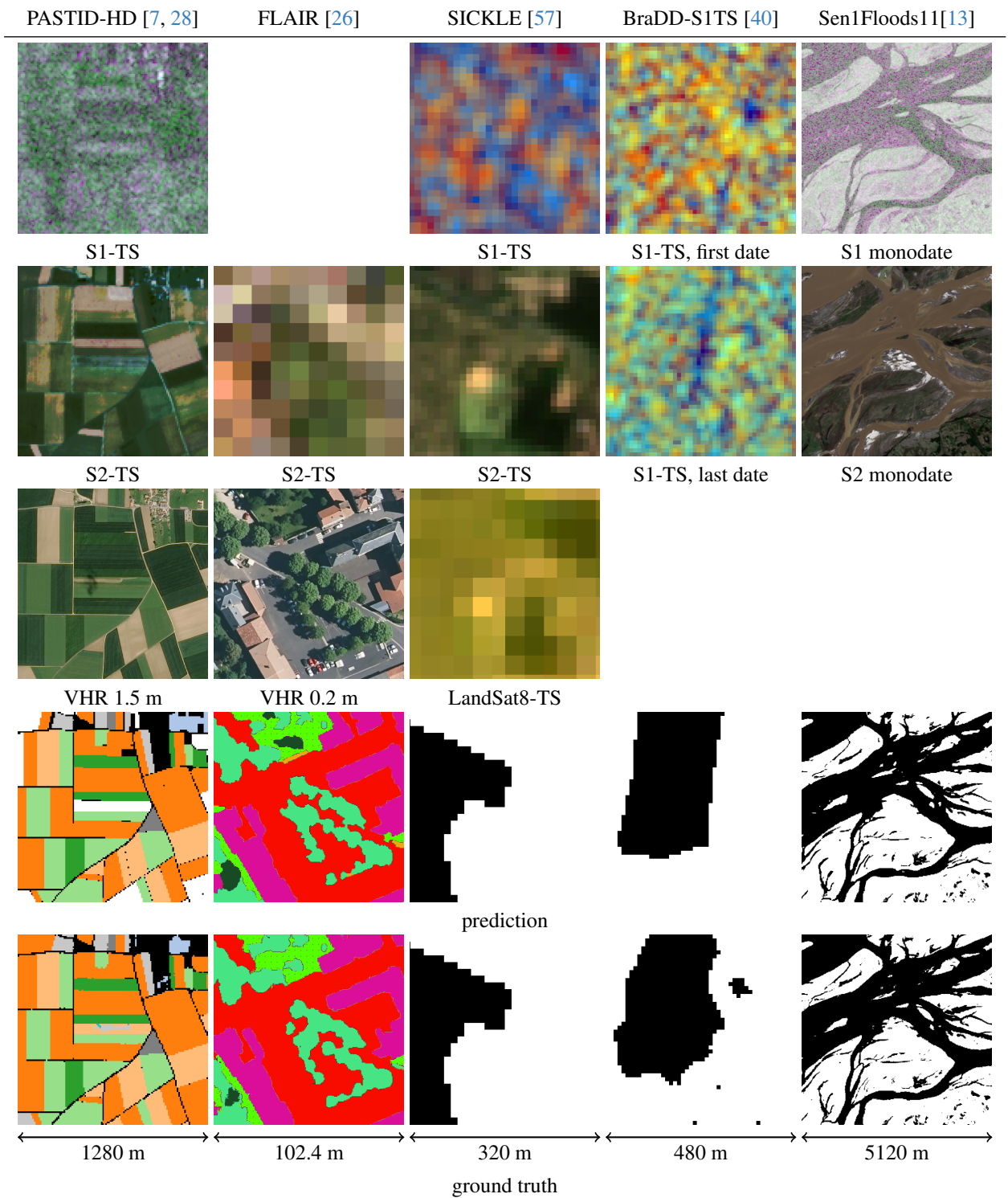
\begin{figure*}
    \centering
\begin{tabular}{c@{\,}c@{\,}c@{\,}c@{\,}c}
PASTID-HD \cite{astruc2024omnisat,garnot2021panoptic}
&
FLAIR \cite{garioud2023flair}
&
SICKLE \cite{sani2024sickle}
&
BraDD-S1TS \cite{karaman2023deforestation}
&
Sen1Floods11\cite{bonafilia2020sen1floods11}
\\\midrule
\includegraphics[width=.18\textwidth, height=.18\textwidth]{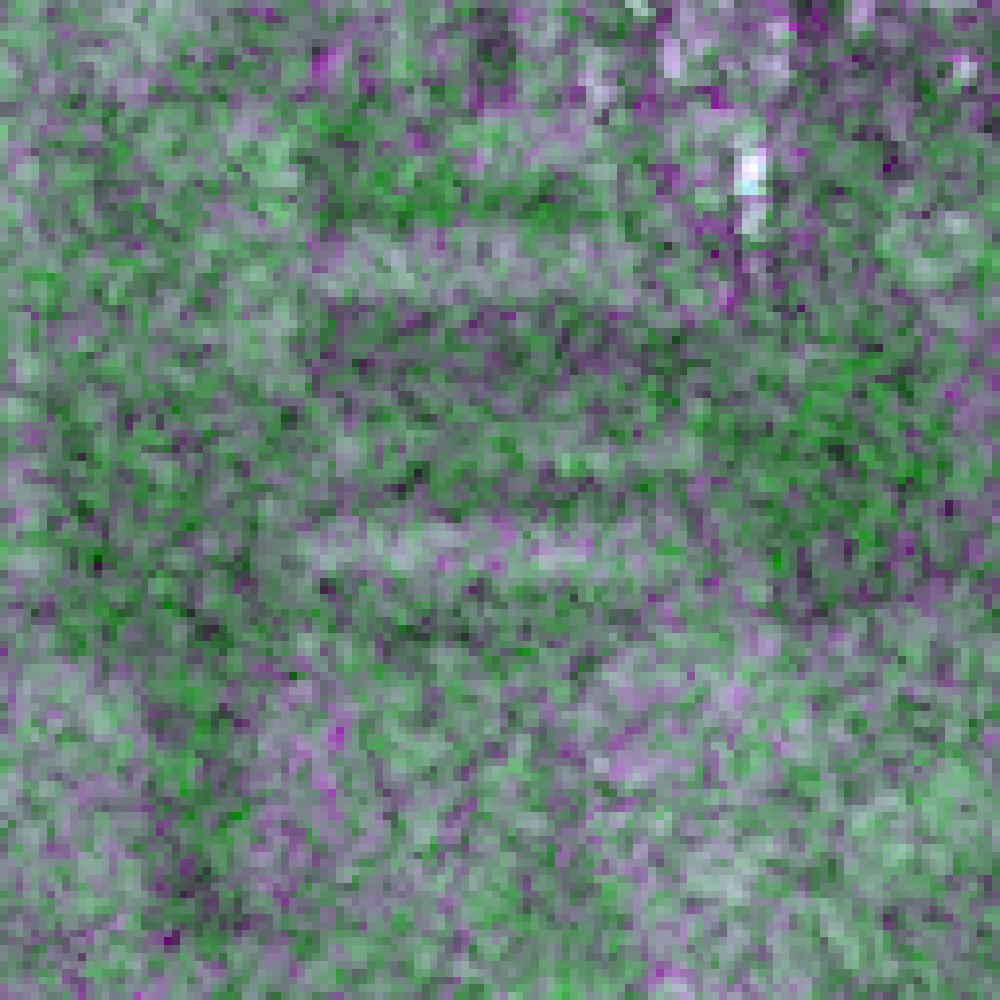}
&
&
\includegraphics[width=.18\textwidth, height=.18\textwidth,trim={1cm 1cm 1cm 1cm},clip]{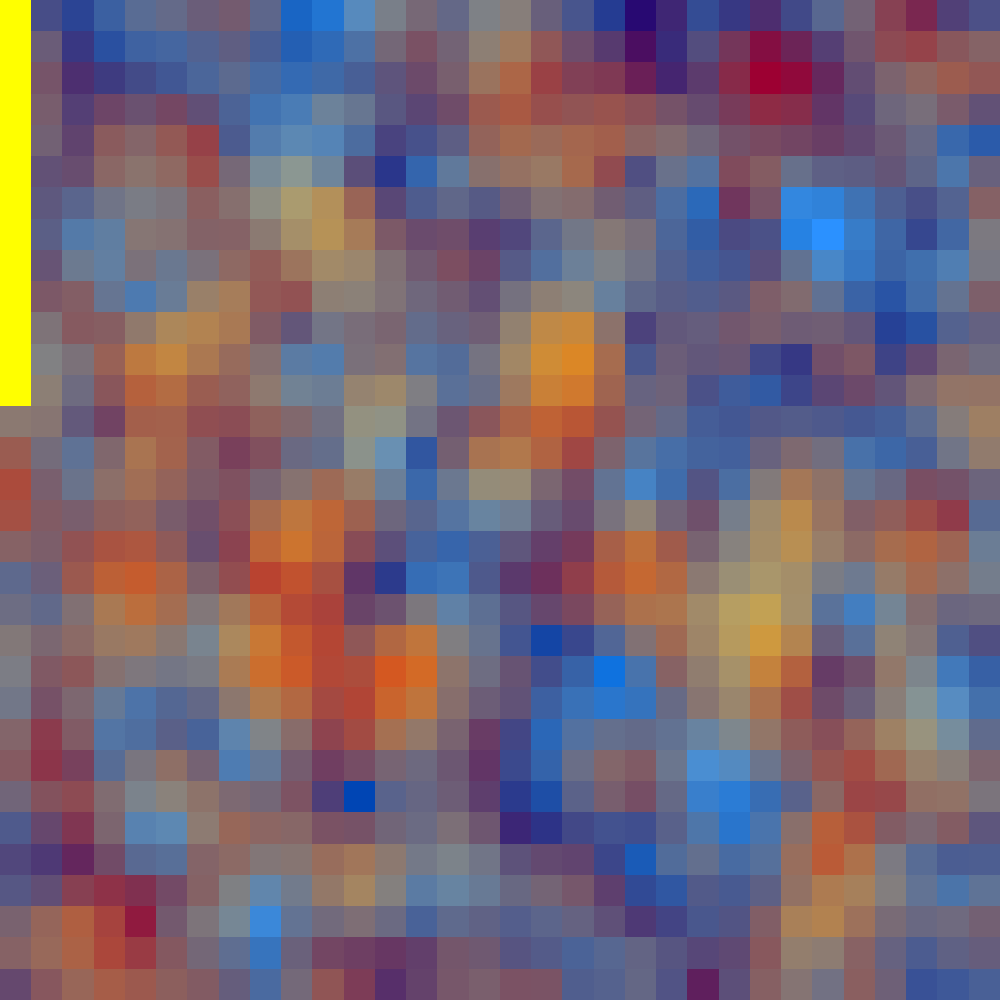}
&
\includegraphics[width=.18\textwidth, height=.18\textwidth]{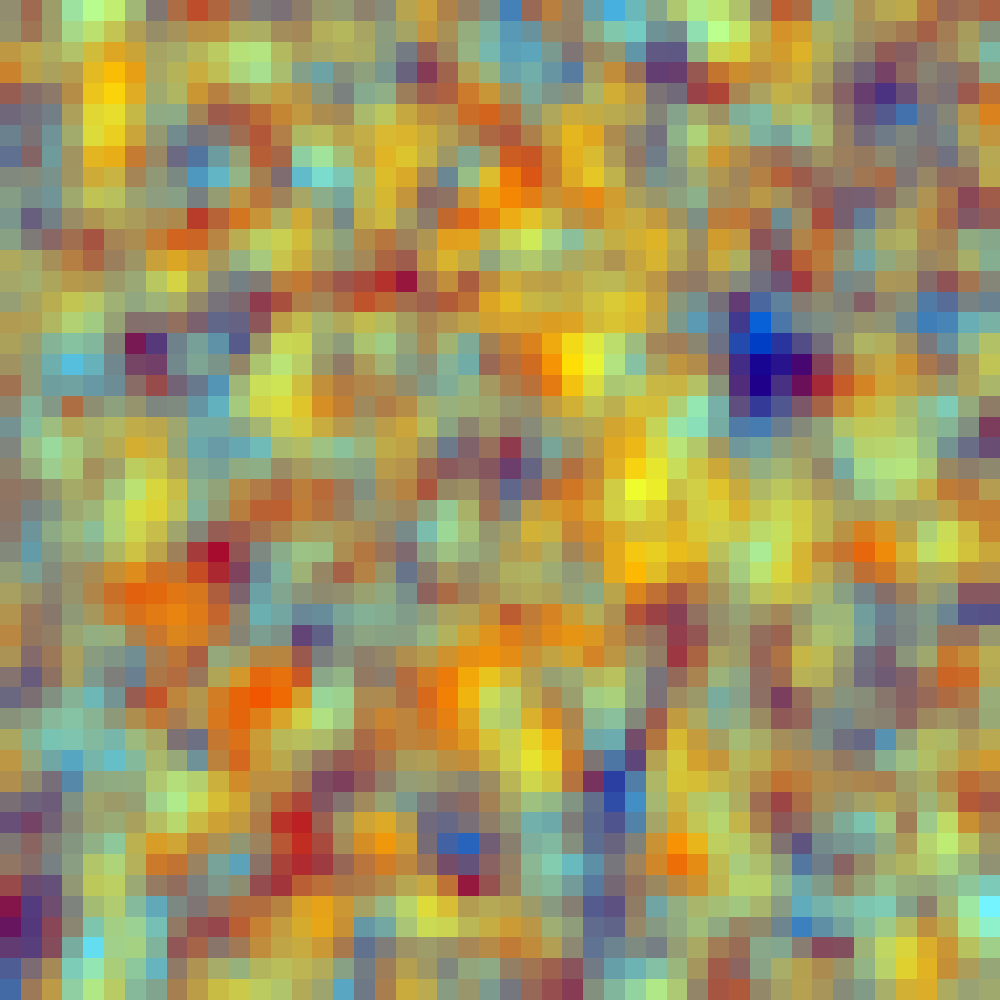}
&
\includegraphics[width=.18\textwidth, height=.18\textwidth]{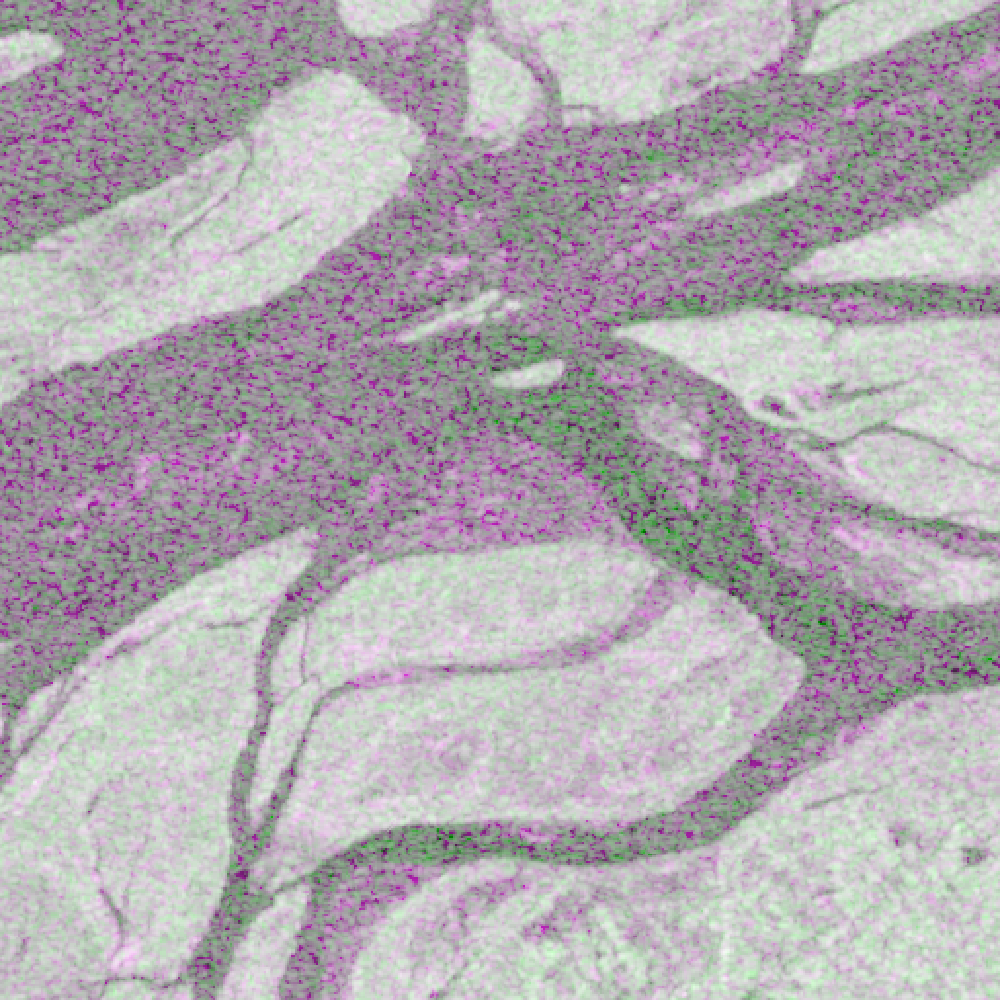}
\\
S1-TS & & S1-TS & S1-TS, first date & S1 monodate
\\
\includegraphics[width=.18\textwidth, height=.18\textwidth]{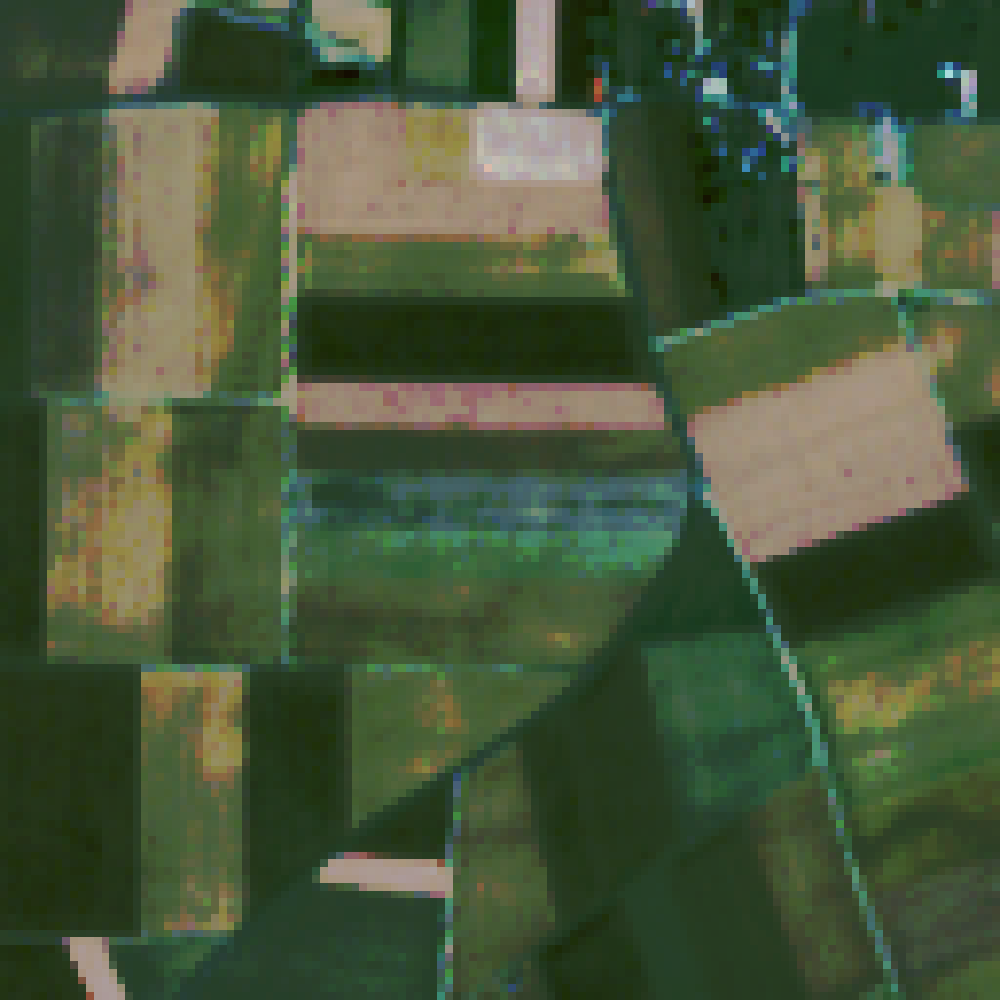}
&
\includegraphics[width=.18\textwidth, height=.18\textwidth]{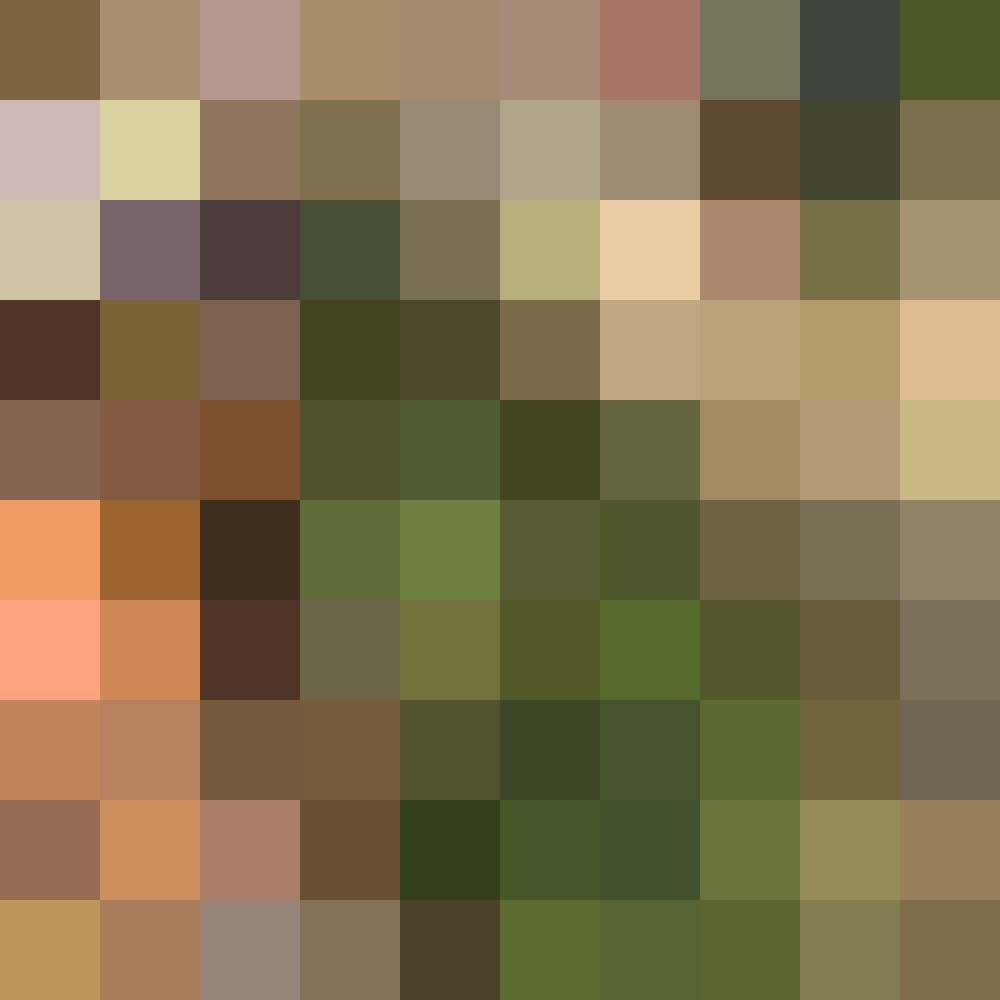}
&
\includegraphics[width=.18\textwidth, height=.18\textwidth,trim={1cm 1cm 1cm 1cm},clip]{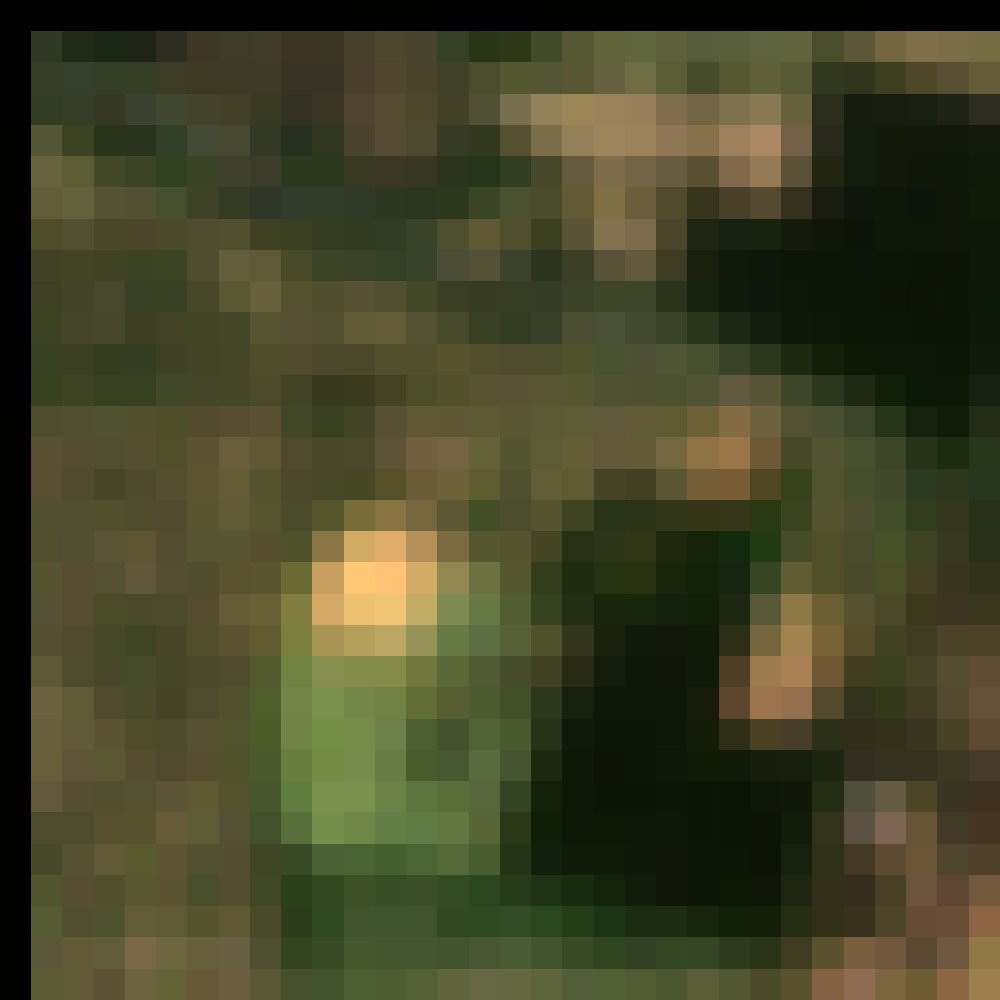}
&
\includegraphics[width=.18\textwidth, height=.18\textwidth]{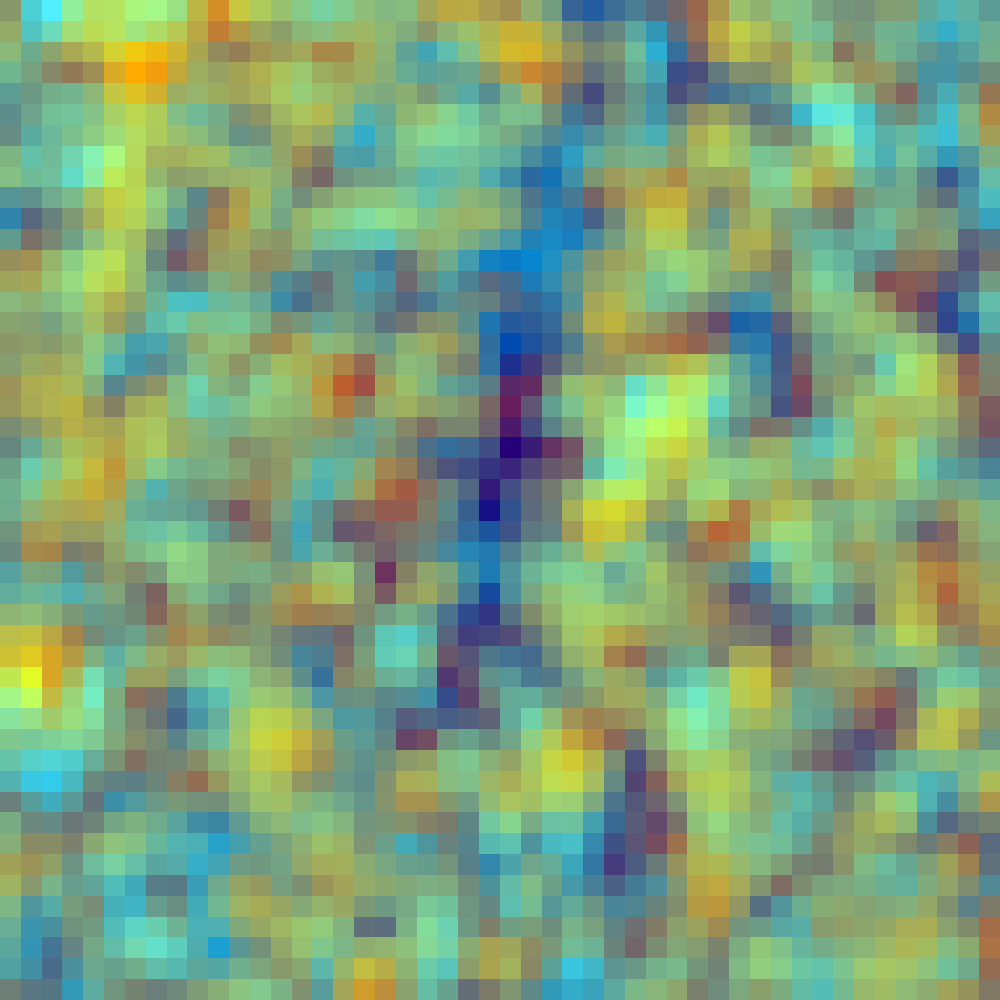}
&
\includegraphics[width=.18\textwidth, height=.18\textwidth]{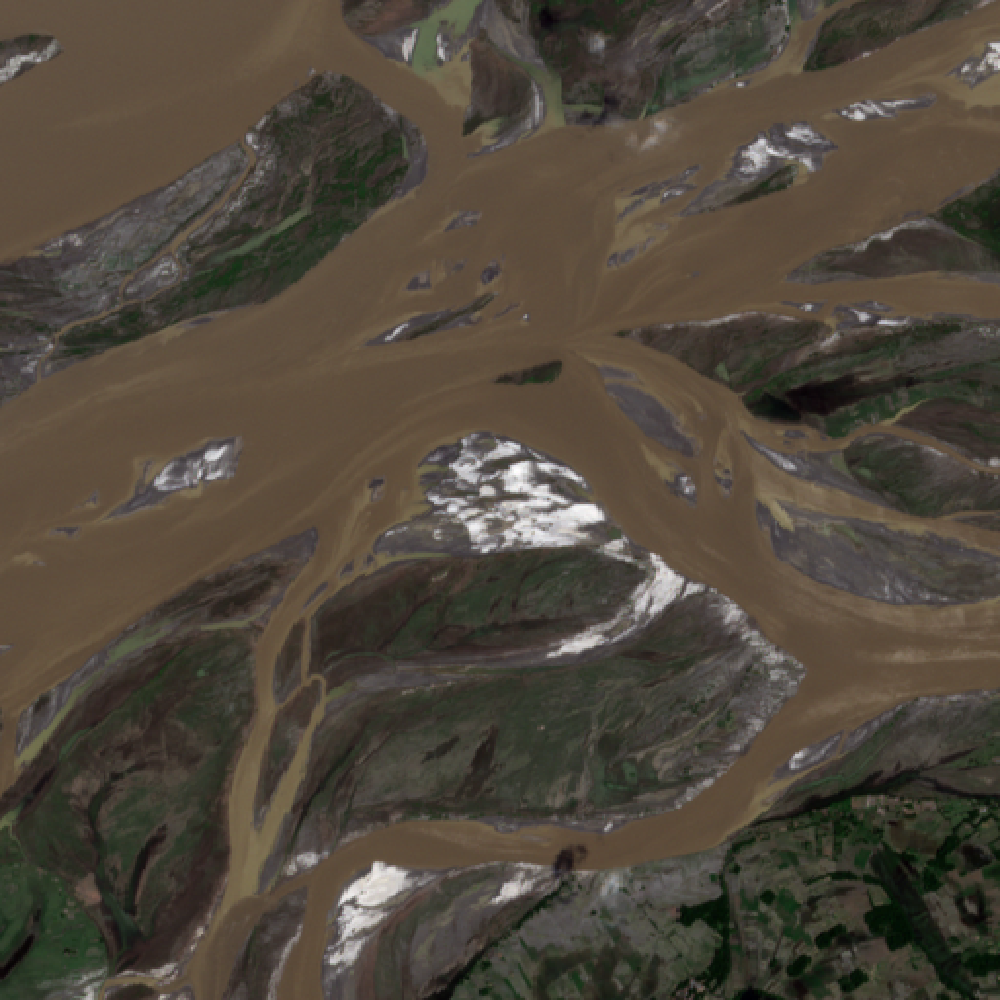}
\\
S2-TS & S2-TS & S2-TS & S1-TS, last date & S2 monodate
\\
\includegraphics[width=.18\textwidth, height=.18\textwidth]{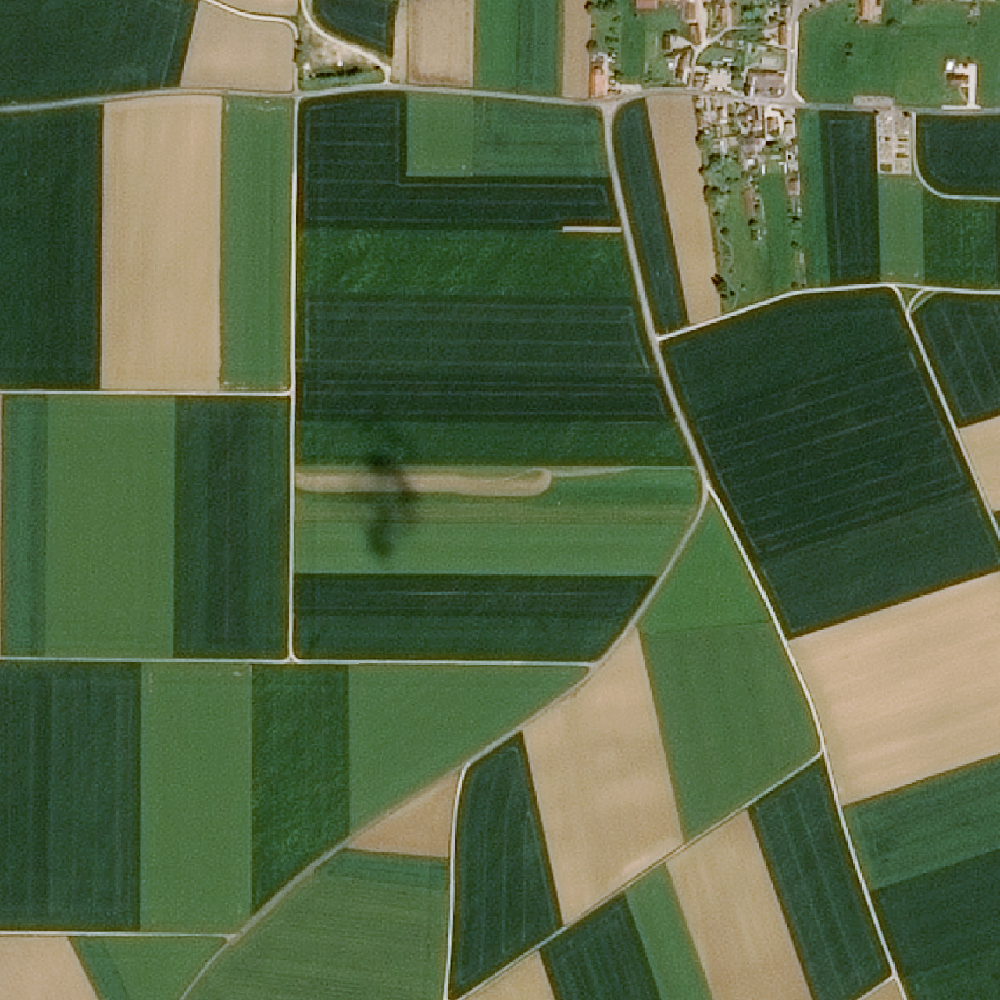}
&
\includegraphics[width=.18\textwidth, height=.18\textwidth]{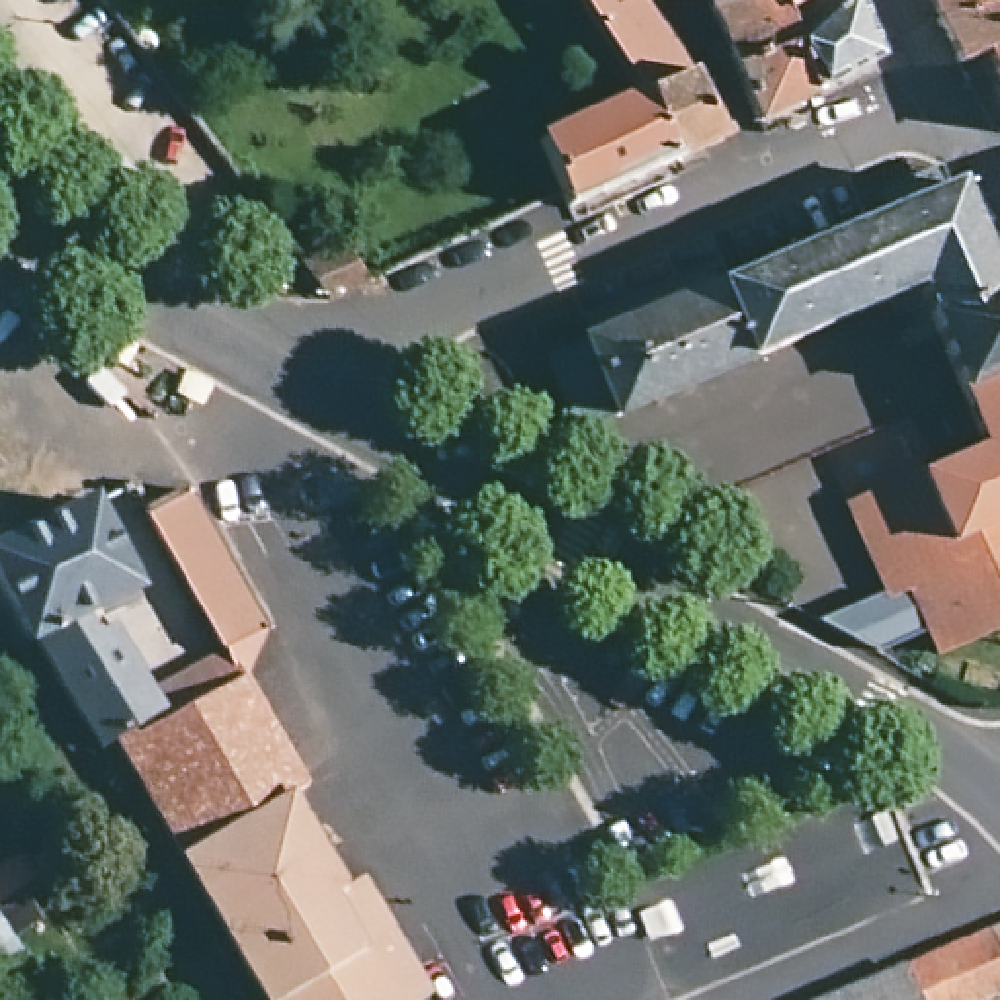}
&
\includegraphics[width=.18\textwidth, height=.18\textwidth,trim={1cm 1cm 1cm 1cm},clip]{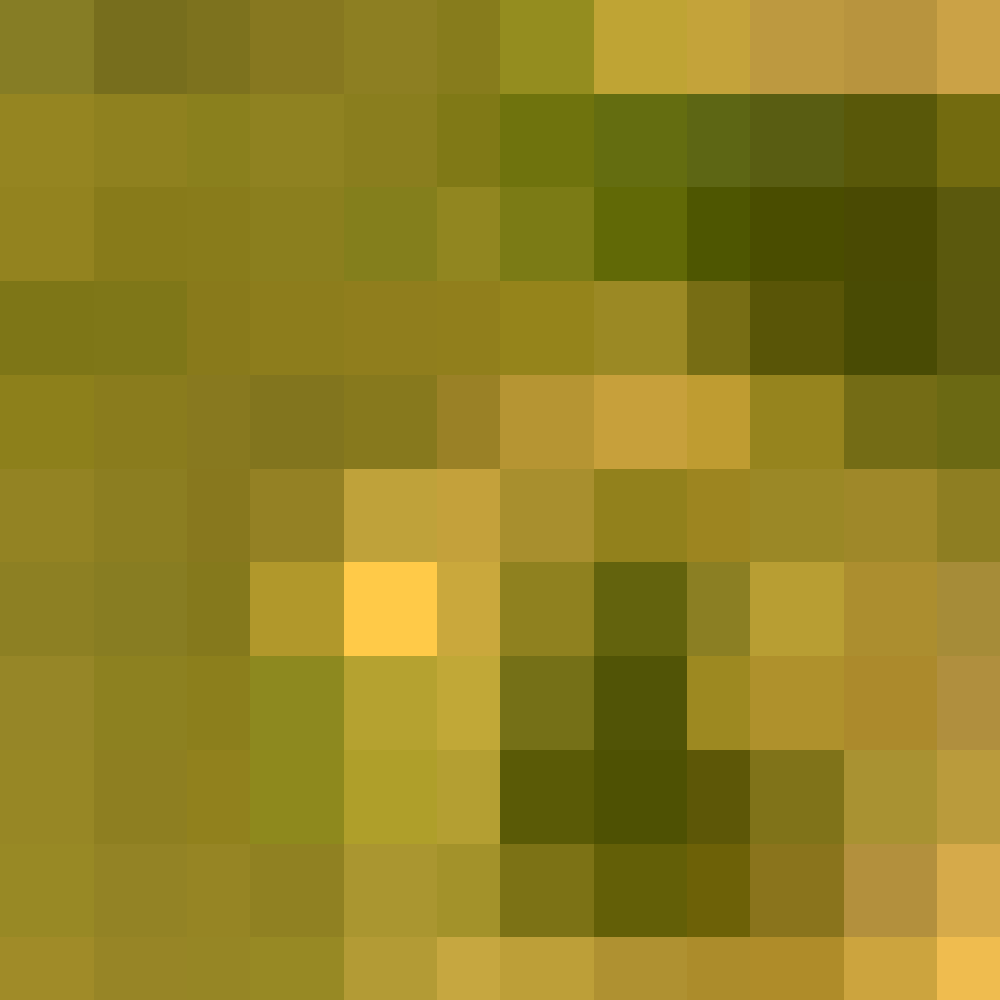}
\\
VHR 1.5 m& VHR 0.2 m & LandSat8-TS
\\
\includegraphics[width=.18\textwidth, height=.18\textwidth]{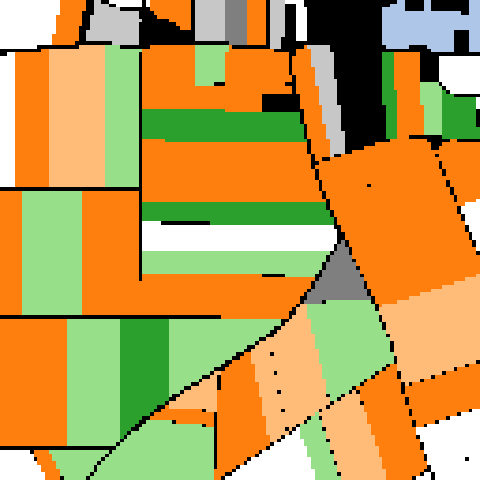}
&
\includegraphics[width=.18\textwidth, height=.18\textwidth]{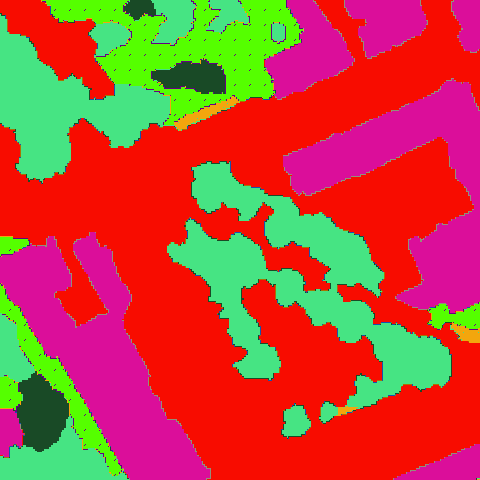}
&
\includegraphics[width=.18\textwidth, height=.18\textwidth,trim={1cm 1cm 1cm 1cm},clip]{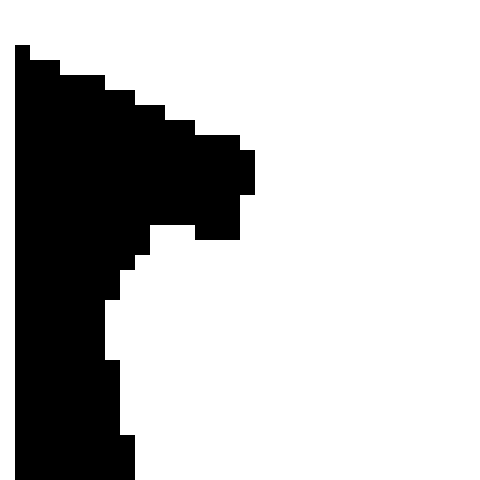}
&
\includegraphics[width=.18\textwidth, height=.18\textwidth]{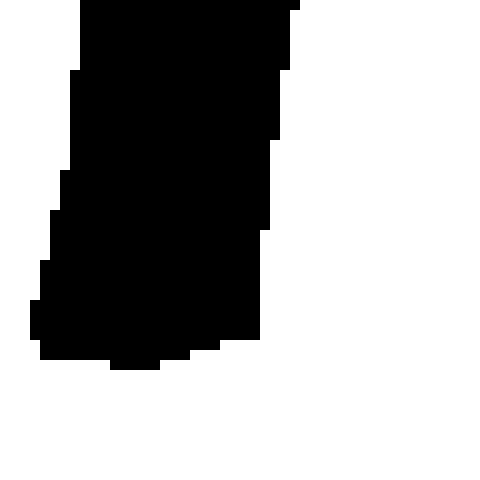}
&
\includegraphics[width=.18\textwidth, height=.18\textwidth]{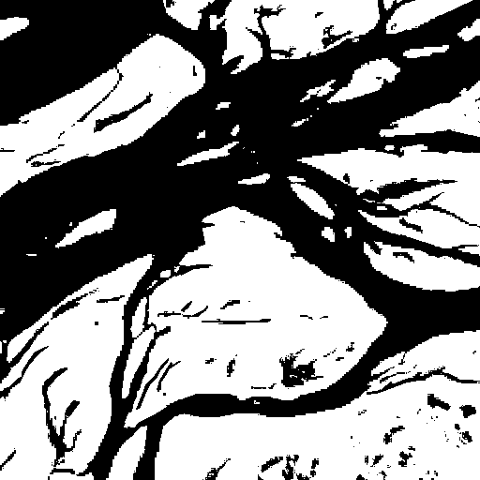}
\\
\multicolumn{5}{c}{ground truth}
\\
\begin{tikzpicture}
    \node [inner sep=0pt, draw=none] (img) {\includegraphics[width=.18\textwidth, height=.18\textwidth]{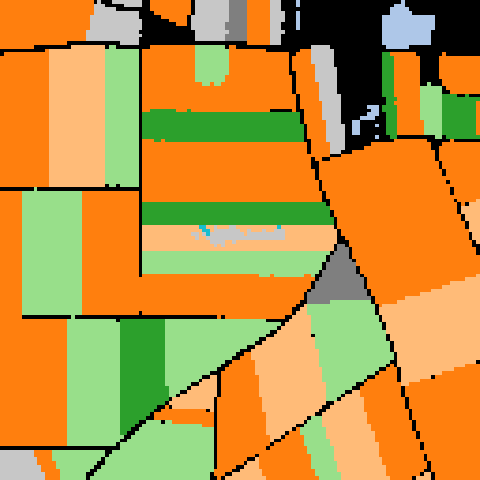}};
    \draw [thick, <->] ([yshift=-2mm] img.south west) -- ([yshift=-2mm] img.south east) node[midway, below] {1280 m};
\end{tikzpicture}
&
\begin{tikzpicture}
    \node [inner sep=0pt, draw=none] (img) {\includegraphics[width=.18\textwidth, height=.18\textwidth]{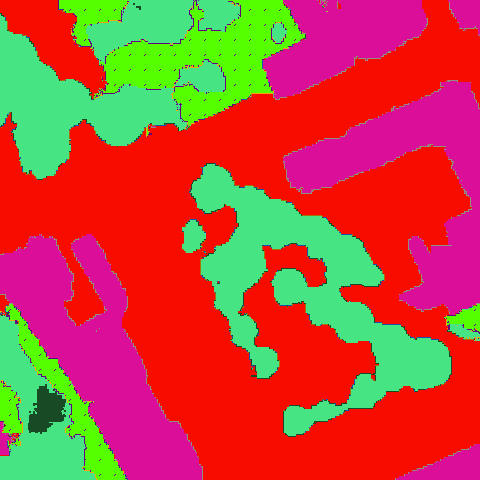}};
    \draw [thick, <->] ([yshift=-2mm] img.south west) -- ([yshift=-2mm] img.south east) node[midway, below] {102.4 m};
\end{tikzpicture}
&
\begin{tikzpicture}
    \node [inner sep=0pt, draw=none] (img) {\includegraphics[width=.18\textwidth, height=.18\textwidth,trim={1cm 1cm 1cm 1cm},clip]{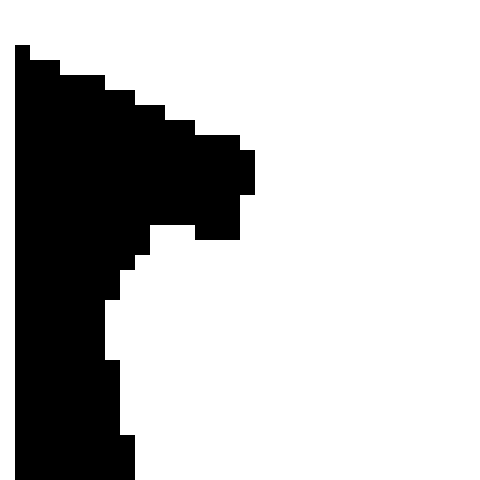}};
    \draw [thick, <->] ([yshift=-2mm] img.south west) -- ([yshift=-2mm] img.south east) node[midway, below] {320  m};
\end{tikzpicture}
&
\begin{tikzpicture}
    \node [inner sep=0pt, draw=none] (img) {\includegraphics[width=.18\textwidth, height=.18\textwidth]{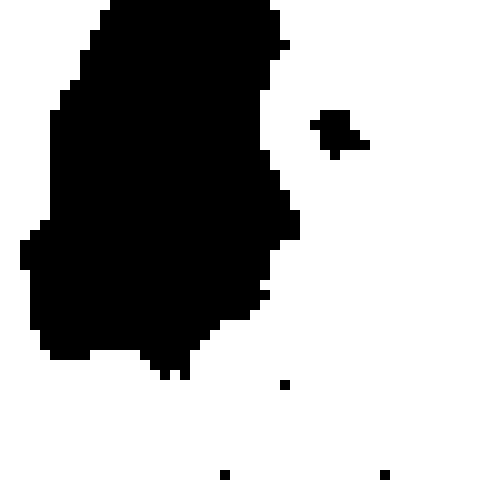}};
    \draw [thick, <->] ([yshift=-2mm] img.south west) -- ([yshift=-2mm] img.south east) node[midway, below] {480 m};
\end{tikzpicture}
&
\begin{tikzpicture}
    \node [inner sep=0pt, draw=none] (img) {\includegraphics[width=.18\textwidth, height=.18\textwidth]{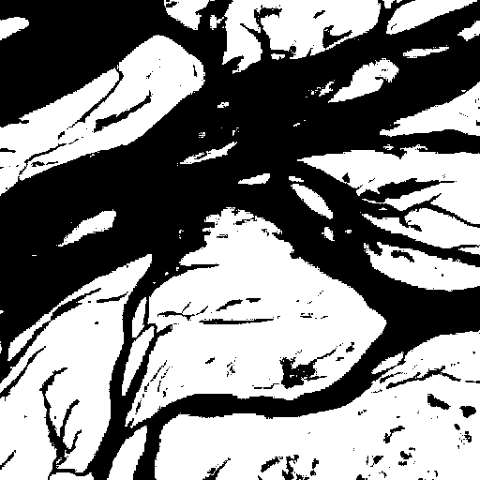}};
    \draw [thick, <->] ([yshift=-2mm] img.south west) -- ([yshift=-2mm] img.south east) node[midway, below] {5120 m};
\end{tikzpicture}
\\
\multicolumn{5}{c}{prediction}
\end{tabular}
    \caption{{\bf Illustration of Results.}
    We represent the inputs, predictions, and ground truth for tiles from four datasets.
    The colormaps are taken directly from the papers.
    TS: time series, a single date has been chosen.  S1/2  stands for Sentinel-1/2. For PASTIS-HD, white parcels are not annotated (void label).~\\~\\}
    \label{fig:quali}
\end{figure*}

\paragraph{Quantitative Results.}  
We provide in \cref{tab:detailed} and \cref{tab:external} the detailed performance of AnySat, with and without pretraining, and an extensive comparison with recent EO models. Pretraining on GeoPlex improves performance for smaller datasets (\eg., TreeSatAI-TS, PASTIS in classification), but this effect is more limited for segmentation datasets (FLAIR, PASTIS in segmentation) or larger ones like PLANTED. We hypothesize that this is due to the quantity of available supervision; for instance, FLAIR has over 20 billion individual labels. In the case of FLAIR, the pretrained model is 0.5 points behind training from scratch, which we attribute to stochastic noise, as our performance on the validation set is on par with training from scratch: $54.7$ for pretrained \vs $54.8$ from scratch.

\section{Additional Ablation}
\label{sec:ablation}

We propose an additional experiment to evaluate the impact of one of our design choices.

\paragraph{No Modality or Temporal Masking.}
In this experiment, we remove the modality and temporal masking for the student encoder during pretraining. This modification results in a slight increase in segmentation performance by $+0.4$ mIoU but a decrease in classification performance by $-0.6$ F1 score. These ambiguous results are similar to the effects we observed with naive patch dropping. An advantage of including modality and temporal masking is that it reduces the memory requirements during training by up to 30\%. Since our goal is to train a single model on several datasets aimed to be fine-tuned for multiple tasks, we keep a unique configuration and adopt this masking strategy.

\begin{table*}[h]
    \centering
        \caption{{\bf Model Performance on the Test Sets of GeoPlex.} For time series, we denote by \mono\; when a single date has been selected, and \stacked\; when seasonal medians
have been concatenated in the channel dimension.\;\;\; AL stands for ALOS-2 and MO for MODIS. LP stands for linear probing
}        \label{tab:detailed}
\vspace{-3mm}
\noindent
\begin{tabular}{cc}
\vtop{\hsize=0.5\linewidth \centering
\small{
\begin{tabular}{p{2.5cm}@{}p{1.6cm}@{}
>{\centering\arraybackslash}p{0.9cm}@{} 
>{\centering\arraybackslash}p{0.9cm}@{} 
>{\centering\arraybackslash}p{0.9cm}@{} >{\raggedleft\arraybackslash}p{0.95cm}}
\toprule
Model & Pre-training & \multicolumn{3}{c}{Modalities} \\
\midrule
\multicolumn{2}{l}{TSAI-TS - multilabel classif.} &  VHR & S1 & S2 & wF1 \\\midrule
\bf AnySat (ours)& GeoPlex &  \yes & \yes & \yes & \bf 75.1\\
\bf AnySat (ours) & None &  \yes & \yes & \yes & 72.7 \\ 
\greyrule
OmniSat \cite{astruc2024omnisat} & TSAI-TS &  \yes & \yes & \yes & 74.2 \\
DOFA \cite{xiong2024dofa} & DOFA &  \yes & \mono & \mono  & 71.6\\
PSE+LTAE \cite{garnot2020lightweight} & None &   & \yes & \yes & 71.2 \\
PSE + ResNet \cite{astruc2024omnisat} & None & \yes & \mono & \mono & 68.1\\
ScaleMAE \cite{reed2023scale} & TSAI & \yes &   & \mono & 62.5 \\
SatMAE \cite{cong2022satmae} & TSAI & \yes &  & \mono &  61.5 \\
CROMA \cite{fuller2023croma} & TSAI & \yes &   & \mono &  61.0 \\
UT\&T \cite{garioud2023flair} & ImageNet &  \yes & \yes & \yes & 56.7 \\
MOSAIKS\cite{rolf2021generalizable} & TSAI & & & \mono  & 56.0 \\
PRESTO \cite{tseng2023lightweight} & PRESTO & & & \mono  & 46.3 
\end{tabular}
\begin{tabular}{p{2.5cm}@{}p{1.6cm}@{}
>{\centering\arraybackslash}p{0.54cm}@{} 
>{\centering\arraybackslash}p{0.54cm}@{}
>{\centering\arraybackslash}p{0.54cm}@{} 
>{\centering\arraybackslash}p{0.54cm}@{} 
>{\centering\arraybackslash}p{0.54cm}@{} 
>{\raggedleft\arraybackslash}p{0.95cm}}
\toprule
Model & Pre-training & \multicolumn{5}{c}{Modalities} \\
\midrule
\multicolumn{2}{l}{PLANTED - classif.} & S1 & S2 & LS & AL & MO & \!maF1
\\\midrule
\bf AnySat (ours) & GeoPlex & \yes & \yes & \yes & \yes &\yes  & 61.5 \\
\bf AnySat (ours) & None &  \yes & \yes & \yes & \yes & \yes & 61.2 \\
\greyrule
ViViT \cite{pazos2024planted,arnab2021vivit} & None & \yes & \yes &  &   &  & \bf 62.2 \\
ViViT \cite{pazos2024planted,arnab2021vivit} & None & \yes & \yes & \yes & \yes &\yes & 59.3\\
\end{tabular}
\begin{tabular}{p{2.5cm}@{}p{1.6cm}@{}
>{\centering\arraybackslash}p{1.35cm}@{}
>{\centering\arraybackslash}p{1.35cm}@{}
>{\raggedleft\arraybackslash}p{0.95cm}}
\midrule
\multicolumn{2}{l}{FLAIR - semantic seg} & VHR & S2 & mIoU
\\\midrule
\bf AnySat (ours) & GeoPlex & \yes & \yes & 55.1 \\
\bf AnySat (ours) & None &  \yes & \yes & 55.6 \\
\greyrule
UT\&T \cite{garioud2023flair} & ImageNet & \yes & \yes &  \textbf{56.9} \\
UNet \cite{he2016deep} & ImageNet & \yes &  & 54.7 \\
UTAE \cite{garnot2021panoptic} & None &  & \yes & 36.1 \\
\bottomrule
\end{tabular}
}} &
\vtop{\vspace{-3.18cm}\hsize=0.5\linewidth \centering
\small{
\begin{tabular}{p{2.5cm}@{}p{1.6cm}@{}
>{\centering\arraybackslash}p{0.9cm}@{} 
>{\centering\arraybackslash}p{0.9cm}@{} 
>{\centering\arraybackslash}p{0.9cm}@{} >{\raggedleft\arraybackslash}p{0.95cm}}
\midrule
\multicolumn{2}{l}{PASTIS-HD - multilabel classif.} &  VHR & S1 & S2 & maF1
\\\midrule
\bf AnySat (ours) & GeoPlex & \yes & \yes & \yes & \bf 72.8 \\
\bf AnySat (ours)& None &  \yes & \yes & \yes & 65.5 \\ 
\greyrule
OmniSat \cite{astruc2024omnisat} & PASTIS-HD & \yes & \yes & \yes   & 69.9 \\
CROMA \cite{fuller2023croma} &  PASTIS-HD & & \stacked & \stacked & 60.1 \\
DOFA \cite{xiong2024dofa} & DOFA & \yes & \stacked & \stacked & 55.7 \\
UT\&T \cite{garioud2023flair} & ImageNet & \yes & \yes & \yes & 53.5 \\
UTAE \cite{garnot2021panoptic} & None & & \yes & \yes & 46.9 \\
ScaleMAE \cite{reed2023scale} & PASTIS-HD & \yes &  & \stacked & 42.2
\end{tabular}
\begin{tabular}{p{2.5cm}@{}p{1.6cm}@{}
>{\centering\arraybackslash}p{0.65cm}@{} 
>{\centering\arraybackslash}p{0.65cm}@{} 
>{\centering\arraybackslash}p{0.65cm}@{} 
>{\centering\arraybackslash}p{0.75cm}@{}
>{\raggedleft\arraybackslash}p{0.95cm}}
\midrule
\multicolumn{2}{l}{PASTIS-HD - semantic seg} &  VHR & S1 & S2 & OA & mIoU
\\\midrule
\bf AnySat (ours) & GeoPlex & \yes & \yes & \yes & 85.0 & \bf 66.5 \\
\bf AnySat (ours) & None &  \yes & \yes & \yes & 84.8 & 66.3 \\ 
\greyrule
SkySense \cite{guo2024skysense} & SkySense & \yes & \yes & \yes & \bf 85.9 & -\;\;\, \\ 
UTAE-MM \cite{garnot2022multi} 
& None & & \yes & \yes & 84.2 & 66.3 \\
TSViT \cite{tarasiou2023vits} & None & & & \yes &  83.4 & 65.4 \\
UTAE \cite{garnot2021panoptic} & None & & &\yes  & -  & 63.1 \\
\end{tabular}
\begin{tabular}{p{3.0cm}@{}p{1.4cm}@{}
>{\centering\arraybackslash}p{0.8cm}@{} 
>{\centering\arraybackslash}p{0.8cm}@{} 
>{\centering\arraybackslash}p{0.8cm}@{} 
>{\raggedleft\arraybackslash}p{0.95cm}}
\midrule
\multicolumn{2}{l}{PASTIS-HD - semseg LP} &  VHR & S1 & S2 & mIoU
\\\midrule
\bf AnySat LP (ours) & GeoPlex & \yes & \yes & \yes & \bf  42.7 \\
\greyrule
S12-DINO LP \cite{stewart2024ssl4eo,oquab2023dinov2} & foundation &\yes & \yes & \yes & 36.2 \\
S12-MoCo LP \cite{stewart2024ssl4eo,he2020momentum} &  foundation&\yes & \yes & \yes & 34.5 \\
S12-D2V LP \cite{stewart2024ssl4eo,baevski2022data2vec} & foundation &\yes & \yes & \yes & 34.3 \\
SpectralGPT \cite{hong2024spectralgpt} & foundation & \yes & \yes & \yes & 35.4 \\
Prithvi \cite{jakubik2310foundation} & foundation &\yes & \yes & \yes & 33.9 \\
\midrule
\end{tabular}

}}
\end{tabular}
\end{table*}

\begin{table}[]
    \centering
        \caption{{\bf External Datasets.} We evaluate our pretrained model on 4 external datasets, in the fine-tuning or linear probing settings.
        \mono\;stands for single-date observations. We report the number of trainable parameters for probing experiments.}      \label{tab:external}\centering
\small{
\begin{tabular}{>{\raggedright\arraybackslash}p{2.8cm}@{}
>{\centering\arraybackslash}p{1.42cm}@{} 
>{\centering\arraybackslash}p{1.42cm}@{} 
>{\centering\arraybackslash}p{1.42cm}@{} >{\raggedleft\arraybackslash}p{1cm}}
\toprule
\multicolumn{1}{l}{SICKLE \cite{sani2024sickle}} &  L8 & S1 & S2 & mIoU
\\\midrule
\bf AnySat (fine-tune)  &  \yes & \yes & \yes & \bf  89.3 \\
\bf AnySat (linear 6.1K)  &  \yes & \yes & \yes & 82.0 \\ \greyrule
Unet3d \cite{sani2024sickle,m2019semantic} &  \yes & \yes & \yes & 82.1 \\
UTAE \cite{sani2024sickle,garnot2021panoptic} &  \yes & \yes & \yes & 51.4 \\ 
\midrule
\multicolumn{1}{l}{BraDD-S1TS \cite{karaman2023deforestation}} & &S1&& mIoU
\\\midrule
\bf AnySat (fine-tune)  &  &\yes & & \bf 80.9 \\ 
\bf AnySat (linear 6.1K)  &  &\yes &  & 78.9 \\ \greyrule
UTAE \cite{garnot2021panoptic} & &\yes && 70.7 \\ 
3D-UNet \cite{m2019semantic} & &\yes&&  68.1\\
Conv-LSTM \cite{shi2015convolutional} & &\yes && 63.7\\
\midrule
\multicolumn{1}{l}{TimeSen2Crop \cite{weikmann2021timesen2crop}} & &&S2& OA
\\\midrule
\bf AnySat (fine-tune)  & & & \yes & \bf 92.2 \\
\bf AnySat (linear 14K)  & & & \yes & 70.3 \\ \greyrule
OS-CNN \cite{tang2021omni,vincent2008extracting}
& &&\yes & 81.2\\
MLP+TAE  \cite{garnot2020lightweight,vincent2024satellite} &  &&\yes & 80.9\\
W.LSTM  \cite{sepp2012long,bruzzone1997classification} &  &&\yes & 78.2\\
Transformer \cite{vaswani2017attention} & && \yes & 78.1 \\
MSResNet \cite{dang2021msresnet} &&& \yes & 76.3 \\ 
\midrule
\multicolumn{1}{l}{Sen1Floods11 \cite{bonafilia2020sen1floods11}} & &S1&S2& mIoU
\\\midrule
\bf AnySat (linear 6.1K)  & & \mono & \mono  & \bf 91.1 \\ \greyrule
\multicolumn{2}{l}{CROMA \cite{fuller2023croma} (UperNet 47M)}
& \mono & \mono & 90.9\\
\multicolumn{2}{l}{CROMA \cite{fuller2023croma} (fine-tune 350M)}
& \mono & \mono & 90.9\\
\multicolumn{2}{l}{Prithvi \cite{jakubik2310foundation} (fine-tune 130M)}
& \mono & \mono & 90.4\\
\multicolumn{2}{l}{Prithvi2 \cite{szwarcman2024prithvi} (fine-tune 630M)}
& \mono & \mono & 90.4\\
\multicolumn{2}{l}{SatlasNet \cite{bastani2023satlaspretrain} (UperNet 33M)}
& \mono & \mono & 90.3\\
\multicolumn{2}{l}{Prithvi \cite{jakubik2310foundation} (UperNet 39M)}
& \mono & \mono & 88.3\\
\\\midrule
\multicolumn{1}{l}{HLS Burn Scar \cite{HLS_Foundation_2023}} & & HLS && mIoU
\\\midrule
\bf AnySat (fine-tune) & & \yes & & \bf 90.6 \\
\bf AnySat (linear 3M)  & & \yes & &  87.7 \\ 
\greyrule
\multicolumn{2}{l}{Prithvi2 \cite{szwarcman2024prithvi} (fine-tune 630M)}
&  \yes && 90.5\\
\multicolumn{2}{l}{Prithvi \cite{jakubik2310foundation} (fine-tune 130M)}
&  \yes&& 86.9\\
\multicolumn{2}{l}{Prithvi \cite{jakubik2310foundation} (UperNet 39M)}
&  \yes && 83.6\\
\multicolumn{2}{l}{CROMA \cite{fuller2023croma} (UperNet 47M)}
&  \yes && 82.4 \\
\multicolumn{2}{l}{DOFA \cite{xiong2024dofa} (UperNet 39M)}
&  \yes && 80.6\\
\midrule
\multicolumn{1}{l}{So2Sat \cite{zhu2019so2sat}} & &S1&S2& OA
\\\midrule
\bf AnySat (linear 29k)  & & \mono & \mono  & 59.1 \\ 
\greyrule
\multicolumn{2}{l}{DOFA \cite{xiong2024dofa} (linear)}
& \mono & \mono & \bf 59.3 \\
\multicolumn{2}{l}{CROMA \cite{fuller2023croma} (linear)}
& \mono & \mono & 49.2\\
\multicolumn{2}{l}{SatMAE \cite{cong2022satmae} (linear)}
& \mono & \mono & 46.9\\
\bottomrule
\end{tabular}
}
\end{table}


\section{Implementation Details}
\label{sec:details}
\paragraph{GeoPlex.} See \cref{tab:GeoPlex} for more details on the composition of GeoPlex. GeoPlex is composed of five distinct datasets—TSAI-TS, PASTIS-HD, FLAIR, PLANTED, and S2NAIP-URBAN—which collectively offer a rich combination of data types, including images, time series, and various modalities. These datasets span extensive geographical areas, ranging from 180 km² to over 211,000 km², and provide a wide array of spatial resolutions (from 0.2m to 250m), temporal resolutions (from 1 to 140 time steps), and spectral resolutions (from 3 to 10 bands). The inclusion of multiple satellite and aerial platforms, such as Sentinel-1/2, Landsat 7/8/9, SPOT6/7, and NAIP, ensures a robust and varied training set.

\begin{table*}[]
   \caption{{\bf Considered Datasets.} We present the detailed composition of GeoPlex, the collection of datasets used for self-supervised training, and our external evaluation datasets. For each dataset, we consider a set of acceptable patch sizes. \\\noindent \textbf{img}: img, \textbf{t.s.}: time series: t.s.
   S1/2: Sentinel-1/2.\; $\dagger$ upsampled from original acquisition resolution.}
    \label{tab:GeoPlex}
    \centering
\label{tab:tsai}
\small{
    \begin{tabular}{lcccccccc}
    \toprule
    \multirow{2}{*}[-0.8mm]{Dataset} & \multirow{2}{*}[-0.8mm]{Extent} & {Sample Size (S)} & \multirow{2}{*}[-0.8mm]{Modalities} & \multicolumn{3}{c}{Resolution} \\\cmidrule(l{8pt}r{8pt}){5-7} 
    &&{Patch Size (P)} &&Spatial (R) & Temporal (T) & Spectral (C) \\\midrule
    \multicolumn{7}{c}{GeoPlex}\\\greyrule
        \multirow{3}{*}{TSAI-TS \cite{astruc2024omnisat,ahlswede2022treesatai}} &
        \multirow{3}{*}{\begin{tabular}{c} 50k $\times$ (1\,img + 2\,t.s.)  \\  180  km²\;-\; 4.7 GPix \end{tabular}}
        & \multirow{3}{*}{\begin{tabular}{c}
             $S=60$m\\
             $P\in\{10,20,30\}$m \\
        \end{tabular}} & Aerial VHR  & 0.20m & 1 & 4 \\ 
        &&& S1 & 10m & 10-70 & 3 \\
        &&& S2 & 10m & 10-70 & 10 \\\greyrule
        \multirow{3}{*}{PASTIS-HD \cite{astruc2024omnisat,garnot2021panoptic}} &
        \multirow{3}{*}{\begin{tabular}{c} 2433 $\times$ (1\,img + 2\,t.s.)  \\  3986  km²\;-\; 7.5 GPix \end{tabular}} & \multirow{3}{*}{\begin{tabular}{c}
             $S=1280$m\\
             $P\in\{40,80,160\}$m 
        \end{tabular}}& SPOT6/7  & 1m$^\dagger$
        & 1 & 4 \\
        &&& S1 & 10m & 140 & 3 \\
        &&& S2 & 10m & 38-61 & 10 \\\greyrule
        \multirow{2}{*}{FLAIR \cite{garioud2023flair}} &
         \multirow{2}{*}{\begin{tabular}{c} 78k $\times$ (1\,img + 1\,t.s.)    \\  815  km²\;-\; 20 GPix \end{tabular}} & \multirow{1}{*}{$S=102.4$m} & Aerial VHR  & 0.2m & 1 & 5 \\
        &&$P\in\{10,20,50\}$m& S2 & 10m & 20-114 & 10 \\\greyrule
        \multirow{5}{*}{Planted \cite{pazos2024planted}} &
        \multirow{5}{*}{\begin{tabular}{c} 1.3M $\times$ (5\,t.s.) \\   33,120 km²\;-\; 3.0 GPix \end{tabular}} & \multirow{5}{*}{\begin{tabular}{c}
             $S=120$m\\
             $P\in\{30,60\}$m 
        \end{tabular}} & S2 & 10m & 8 
         & 10 \\
        &&& S1 & 10m & 8 
        & 3 \\
        &&& Landsat 7 & 30m & 20 
        & 3 \\
        &&& ALOS-2 & 30m & 4 
        & 3 \\
        &&& MODIS & 250m & 60 
        & 7 \\\greyrule
         &
         \multirow{4}{*}{\begin{tabular}{c} 515k $\times$ (1\,img +  3\,t.s.) \\   211,063 km²\;-\; 136 GPix \end{tabular}} & \multirow{4}{*}{\begin{tabular}{c}
             $S=640$m\\
             $P\in\{40,80,160\}$m 
        \end{tabular}} & NAIP & 1.25m & 1 & 4 \\
       \multirow{1}{*}{S2NAIP-} &&& S2 & 10m & 16-32  & 10 \\
        \multirow{1}{*}{URBAN \cite{s2naip,satlassuperres}}
        &&& S1 & 10m & 2-8 & 3 \\
        &&& Landsat 8/9 & 10m$^\dagger$ & 4 & 8 \\\midrule
       \multicolumn{7}{c}{External datasets}\\
       \greyrule
       \multirow{2}{*}{BraDD-S1TS \cite{karaman2023deforestation}} &
        \multirow{2}{*}{\begin{tabular}{c} 13k $\times$ (1\,t.s.) \\   2,995 km²\;-\; 1.2 GPix \end{tabular}} & \multirow{2}{*}{\begin{tabular}{c}
             $S=480$m\\
             $P=10$ m
        \end{tabular}} & \multirow{2}{*}{S1} & \multirow{2}{*}{10m} & \multirow{2}{*}{20-66} 
         & \multirow{2}{*}{10} \\
         \\\greyrule
        \multirow{2}{*}{Sickle \cite{sani2024sickle}} &
        \multirow{1}{*}{\begin{tabular}{c} 35k $\times$ (2\,t.s.) \\   3,584 km²\;-\; 3.6 GPix \end{tabular}} & \multirow{1}{*}{\begin{tabular}{c}
             $S=320$m\\
             $P=10$m 
        \end{tabular}} & S2 & 10m & 13-148 
         & 10 \\
         &&& Landsat 8/9 & 10m$^\dagger$ & 8-34
         & 8
         \\\greyrule
         \multirow{2}{*}{TimeSen2Crop \cite{weikmann2021timesen2crop}} &
        \multirow{2}{*}{\begin{tabular}{c} 1.2M $\times$ (1\,t.s.) \\   120 km²\;-\; 35 MPix \end{tabular}} & \multirow{2}{*}{\begin{tabular}{c}
             $S=10$m\\
             $P=10$m 
        \end{tabular}} & \multirow{2}{*}{S2} & \multirow{2}{*}{10m} & \multirow{2}{*}{29} 
         & \multirow{2}{*}{10} \\
         \\\greyrule
         \multirow{2}{*}{Sen1floods11 \cite{bonafilia2020sen1floods11}} &
        \multirow{2}{*}{\begin{tabular}{c} 4.8k $\times$ (2\,img) \\   125,829 km²\;-\; 2.6 GPix \end{tabular}} & \multirow{2}{*}{\begin{tabular}{c}
             $S=5120$m\\
             $P=80$m 
        \end{tabular}} & S2 & 10m & 1 
         & 10 \\
         &&& S1 & 10m & 1 
         & 3 \\\greyrule
         \multirow{2}{*}{So2Sat \cite{zhu2019so2sat}} &
        \multirow{2}{*}{\begin{tabular}{c} 400k $\times$ (2 img) \\   41,029 km²\;-\; 82 GPix \end{tabular}} & \multirow{2}{*}{\begin{tabular}{c}
             $S=320$m\\
             $P=10$m 
        \end{tabular}} & S2 & 10m & 1 
         & 10 \\
         &&& S1 & 10m & 1 
         & 3 
         \\\greyrule
         \multirow{2}{*}{HLS Burn Scar \cite{HLS_Foundation_2023}} &
        \multirow{2}{*}{\begin{tabular}{c} 804 $\times$ (1\,t.s.) \\  188,208 km²\;-\; 211 MPix \end{tabular}} & \multirow{2}{*}{\begin{tabular}{c}
             $S=15300$m\\
             $P=240$m 
        \end{tabular}} & \multirow{2}{*}{HLS} & \multirow{2}{*}{30m} & \multirow{2}{*}{1} 
         & \multirow{2}{*}{6} \\
         \\
        \bottomrule
    \end{tabular}
}
\end{table*}

\paragraph{Network Architecture.} AnySat's architecture follows the Vision Transformer (ViT) template and has 125M learnable parameters, of which 73.6\% are modality-agnostic and resolution-adaptive. The components of the model are:
\begin{compactitem}
    \item {\bf Modality Projectors $\ProjEncod_m$ (33M parameters for $11$ projectors).}  These modules are MLPs responsible for projecting the input data of each modality into a common feature space. 
    
    \item {\bf Spatial Transformer $\TransEncod$ (45M parameters).}  Composed of three self-attention transformer blocks, this module captures the spatial relationships between subpatches for each modality and patch.
   
    \item {\bf Modality Combiner $\Combiner$  (49M parameters).} This module consists of three self-attention blocks followed by a cross-attention block, and merges the representations from different modalities into a unified feature vector for each patch.  
    \item {\bf Predictor $\Predictor$ (29M parameters).}  Exclusive to the student, this module is a single self-attention block and predicts the teacher's embeddings for the dropped patches.
\end{compactitem}

\paragraph{Handling MODIS data.} In the Planted dataset \cite{pazos2024planted}, MODIS observations are included, but their resolution (250~meters) is larger than the entire observed tile (120~meters). We treat these observations as \textit{context} tokens: we concatenate their $\PatchEncod$ embeddings to the $\vert \bM \vert \cdot (S / P)^2$ tokens from all other modalities.
We do not add positional encoding, and this token is not included in the contrastive loss.

\paragraph{Optimization Parameters.} 
 To better manage our memory usage, we adapt the batch size to the size of the samples of each dataset: TreesatAI-TD: 384, PASTIS-HD: 8, FLAIR: 96, PLANTED: 2048, S2NAIP: 16.
We use 8 NVIDIA H100 for experiments on GeoPlex, PLANTED and Pastis-HD , and a smaller cluster of 3 A600 for TreeSatAI-TS and FLAIR.
 
Beyond the changes above, all optimization parameters are shared across all datasets. We used the AdamW \cite{loshchilovdecoupled} optimizer with a learning rate of $5 \times 10^{-5}$ for all our experiments (pretraining and fine-tuning). 
We used a \texttt{LinearWarmupCosineAnnealingLR} \cite{linearwarmup} for classification and \texttt{ReduceLROnPlateau} \cite{reduce} scheduler for pretraining and segmentation.

We set he contrastive temperature $\gamma$ to $0.1$ to n Eq.~X. We used an EMA decay of $0.996$. All other hyperparameters are shared with \href{https://github.com/facebookresearch/ijepa}{original JEPA implementation}.

\paragraph{Position Encodings.} We describe here our scale-adaptive positional encoding which allows us to use the same encoders for different resolutions, scales, and patch size.
The input tokens to the modality combiner $\Combiner$ correspond to patches of size $P \times P$ meters, while those to the spatial transformer $\TransEncod$ represent subpatches of size $(R_m \delta_m) \times (R_m \delta_m)$ meters. Here, $R_m$ varies per sensor modality $m$, and $P$ is randomly chosen for each batch during training. To train a single scale-aware model capable of handling varying resolutions, we employ a scale-adaptive positional encoding inspired by Scale-MAE \cite{reed2023scale}.

We use the same positional encodings in $\Combiner$ and $\TransEncod$.
We first describe the positional encoding of a token by $\Combiner$. 
We denote by $\pos_x$ the index of the token's patch within its tile along the $x$-axis; similarly, $\pos_y$ along the $y$-axis.
If the embeddings of the token have a dimension $D$, the positional encodings $\mu_x(\pos_x, i)$ and equivalently $\mu_y(\pos_y, i)$ are of size $D/2$. For $i \in [0,D/2[$ we have:
\begin{align}
\mu_x(\pos_x, i) = \sin
\left( 
\frac{g}{G}
 \frac{\pos_x}{10000^{\frac{i}{E}}} 
+ \frac\pi2\text{mod}(i,2)
\right)~,
\end{align}
where $g=P$ is the size in meter of the patch  considered unit: patch of size  for $\Combiner$, and $G$ is a reference length that we set to one meter.
We compute $\mu_y(\pos_y, i)$ similarly, and the positional encoding is the channelwise concatenation of both vectors. The positional encoding is directly added to the embeddings.

For $\TransEncod$, we define the positional encoding of each subpatch within its patch with the same formula, but set $g$ to $g=R_m \delta_m$, the size of the subpatch in meter.

\ifx
\begin{align}
\small
v_{x}(\pos, 2i)&=\sin \frac{g}{G}\frac{\pos}{10000^{\frac{2i}{E}}} \\
v_{y}(\pos, 2i+1)&=\cos \frac{g}{G}\frac{\pos}{10000^{\frac{2i}{E}}}
\label{eq:pos_enc}
\end{align}

where $G$ is set to $1$ meter. For the modality combiner, the value $g$ is $P$ and for the spatial transformer, it is $R_m \delta_m$.
\fi


\section{Datasets and Tasks}
\label{sec:datasets}

Here, we provide more details about the datasets used to train and evaluate AnySat and their associated tasks. See \cref{tab:GeoPlex} for an overview of the datasets used in GeoPlex.

 \paragraph{ TreeSatAI-TS \cite{ahlswede2022treesatai,astruc2024omnisat}:} This multimodal dataset is designed for tree species identification and consists of 50,381 tiles, each covering an area of 60$\times$60 meters, with multi-label annotations across 20 classes. All data were collected in Germany. The dataset includes Very High Resolution (VHR) images at 0.2 m with a NIR band, Sentinel-2 time series, and Sentinel-1 time series.
    
    \paragraph{ PASTIS-HD \cite{garnot2022multi, astruc2024omnisat}:} This crop mapping dataset supports classification, semantic segmentation, and panoptic segmentation. Each agricultural parcel is delineated at a resolution of 10 m and annotated across 18 crop types. The dataset contains 2,433 tiles with an extent of 1,280$\times$1,280 m, including Sentinel-2 time series, Sentinel-1 time series (we use only the ascending orbit), and SPOT6 VHR imagery at 1.5~m resolution.
    
    \paragraph{ FLAIR \cite{garioud2023flair}:} This dataset combines VHR aerial imagery at a 0.2 m resolution with Sentinel-2 time series data and comprises 77,762 tiles acquired across metropolitan France. The VHR images include five channels: RGB, near-infrared, and a normalized digital surface model derived by photogrammetry. Each VHR pixel is annotated with one of 13 land cover classes.
 
    \paragraph{ PLANTED \cite{pazos2024planted}:}  The PLANTED dataset is specifically designed for tree species identification and features 1,346,662 tiles of planted forest across the world. Each tile is associated with one of 40 distinct classes. This dataset integrates imagery from five different satellites with various resolutions: Sentinel-2 (10 m), Landsat-7 (30 m), MODIS (250 m), as well as radar time series from Sentinel-1 (10 m) and ALOS-2 (30 m). The time series are temporally aggregated at various intervals---seasonally, monthly, or yearly.
    
    \paragraph{ S2Naip-Urban \cite{s2naip,satlassuperres}:} This dataset includes images captured at the same locations as the S2NAIP-Urban super-resolution dataset \cite{satlassuperres}, which is a subset of the extensive S2NAIP \cite{s2naip} dataset focused on urban areas. This split comprises 515,270 tiles, featuring imagery from NAIP at a 1.25~m resolution, Sentinel-2 and Sentinel-1 time series, and Landsat-8/9 data rescaled to a 10~m resolution. We use this dataset for pretraining only because there are no official labels and evaluations.
    
    \paragraph{ BraDD-S1TS \cite{karaman2023deforestation}:}
    BraDD-S1TS (Brazilian Deforestation Detection) is a change detection dataset comprising Sentinel-1 time series of the Amazon rainforest, aiming to segment deforested areas. It includes 13,234 tiles covering regions with varying deforestation rates, providing pixel-wise binary annotations for deforestation events occurring between the time series' first and last radar image.
    
    \paragraph{ Sickle \cite{sani2024sickle}:} SICKLE is a multimodal crop mapping dataset from India containing 34,848 tiles with Sentinel-1, Sentinel-2, and Landsat-8 time series.
    We use the paddy / non-paddy culture binary semantic segmentation task. As the test set has not been released by the authors, we perform our experiments on the validation set.
    
    \paragraph{  TimeSen2Crop \cite{weikmann2021timesen2crop}:}
    TimeSen2Crop is a crop mapping dataset consisting of 1,212,224 single-pixel Sentinel-2 time series, a configuration not present in GeoPlex. It includes data from Slovenia with annotations for 16 different crop types.

     \paragraph{  Sen1floods11 \cite{bonafilia2020sen1floods11}:}
    Sen1Floods11 is a flood segmentation dataset featuring 4,831 pairs of Sentinel-1 and Sentinel-2 images, each annotated with dense flooded/not-flooded labels. The dataset spans diverse global regions, with each tile covering a 5120 $\times$ 5120 m area (~2600 hectares) and containing a single acquisition date per sensor.

    \paragraph{So2Sat \cite{zhu2019so2sat}:}
    {So2Sat is a local climate zone classification dataset containing co-registered single-date Sentinel-1 and Sentinel-2 imagery across multiple cities worldwide. It comprises 400,673 image patches, each annotated with one of 17 local climate zone classes according to the LCZ scheme. An image represents a zone of size 320 $\times$ 320 m. So2Sat specifically targets urban morphology classification tasks for sustainable urban planning and climate studies.}
    
    \paragraph{ HLS Burn Scar \cite{HLS_Foundation_2023}:}
    {HLS Burn Scar is designed for post-fire burn scar detection using Harmonized Landsat-Sentinel (HLS) imagery. It contains 804 tiles covering a 15.3 $\times$ 15.3 km area 23400 hectares) at $30$m resolution and covering multiple wildfire events across diverse ecosystems in the United States.}
    
\ifx
\begin{table}[]
   \caption{{\bf Experiments (1/2).} We report the detailed results for all experiments with a large array of methods. For a time series, we denote by \mono~ the single-date version where a single date has been selected, and \stacked~the stacked version, where the seasonal medians have been concatenated in the channel dimension.\\
   VHR: very high resolution image, S1/2: Sentinel 1/2} 
    \label{tab:experiments1}
    \centering
\small{
\begin{tabular}{p{2.5cm}@{}p{1.8cm}@{}
>{\centering\arraybackslash}p{0.9cm}@{} 
>{\centering\arraybackslash}p{0.9cm}@{} 
>{\centering\arraybackslash}p{0.9cm}@{} >{\raggedleft\arraybackslash}p{0.75cm}}
\toprule
Model & Pre-training & \multicolumn{3}{c}{Modalities} \\
\midrule
\multicolumn{2}{l}{TSAI-TS - multilabel classif.} &  VHR & S1 & S2 & wF1 \\\midrule
\bf AnySat (ours) & None &  \yes & \yes & \yes & 72.7 \\ 
\bf AnySat (ours)& GeoPlex &  \yes & \yes & \yes & \bf 75.1\\\greyrule
OmniSat \cite{astruc2024omnisat} & TSAI-TS &  \yes & \yes & \yes & 74.2 \\
DOFA \cite{xiong2024dofa} & DOFA &  \yes & \mono & \mono  & 71.6\\
PSE+LTAE \cite{garnot2020lightweight} & None &   & \yes & \yes & 71.2 \\
ResNet \cite{he2016deep} & ImageNet & \yes &   &   & 71.0 \\
PSE + ResNet \cite{astruc2024omnisat} & None & \yes & \mono & \mono & 68.1\\
ScaleMAE \cite{reed2023scale} & TSAI & \yes &   & \mono & 62.5 \\
SatMAE \cite{cong2022satmae} & TSAI & \yes &  & \mono &  61.5 \\
CROMA \cite{fuller2023croma} & TSAI & \yes &   & \mono &  61.0 \\
UT\&T \cite{garioud2023flair} & ImageNet &  \yes & \yes & \yes & 56.7 \\
MOSAIKS\cite{rolf2021generalizable} & TSAI & & & \mono  & 56.0 \\
PRESTO \cite{tseng2023lightweight} & PRESTO & & & \mono  & 46.3 
\end{tabular}
\begin{tabular}{p{2.5cm}@{}p{1.8cm}@{}
>{\centering\arraybackslash}p{0.9cm}@{} 
>{\centering\arraybackslash}p{0.9cm}@{} 
>{\centering\arraybackslash}p{0.9cm}@{} >{\raggedleft\arraybackslash}p{0.75cm}}
\midrule
\multicolumn{2}{l}{PASTIS-HD - multilabel classif.} &  VHR & S1 & S2 & mF1
\\\midrule
\bf AnySat (ours)& None &  \yes & \yes & \yes & 65.5 \\ 
\bf AnySat (ours) & GeoPlex & \yes & \yes & \yes & \bf 72.8 \\\greyrule
OmniSat \cite{astruc2024omnisat} & PASTIS-HD & \yes & \yes & \yes   & 69.9 \\
CROMA \cite{fuller2023croma} &  PASTIS-HD & & \stacked & \stacked & 60.1 \\
ResNet \cite{he2016deep} & ImageNet & \yes &  &  & 59.3 \\
DOFA \cite{xiong2024dofa} & DOFA & \yes & \stacked & \stacked & 55.7 \\
UT\&T \cite{garioud2023flair} & ImageNet & \yes & \yes & \yes & 53.5 \\
UTAE \cite{garnot2021panoptic} & None & & \yes & \yes & 46.9 \\
ScaleMAE \cite{reed2023scale} & PASTIS-HD & \yes &  & \stacked & 42.2
\end{tabular}
\ifx
\begin{tabular}{p{2.5cm}@{}p{1.8cm}@{}
>{\centering\arraybackslash}p{1.35cm}@{}
>{\centering\arraybackslash}p{1.35cm}@{}
>{\raggedleft\arraybackslash}p{0.75cm}}
\midrule
\multicolumn{2}{l}{FLAIR - multilabel classif.} & VHR & S2 & mF1
\\\midrule
\bf AnySat (ours) & None &  \yes & \yes & 71.8 \\ 
\bf AnySat (ours) & GeoPlex & \yes & \yes & * \\\greyrule
DOFA \cite{xiong2024dofa} & DOFA & \yes & \stacked & \bf 74.9\\
OmniSat \cite{astruc2024omnisat} & FLAIR & \yes & \yes & 73.4 \\
ScaleMAE \cite{reed2023scale} & FLAIR & \yes & \stacked & 70.0 \\
UT\&T \cite{garioud2023flair} & ImageNet & \yes & \yes & 48.8
\end{tabular}
\fi
}
\end{table}

\begin{table}[]
   \caption{{\bf Experiments (2/2).} We report the detailed results for all experiments with a large array of methods. $\star$ Performance on PASTIS-MM, a comparable dataset to PASTIS-HD.\\
  \noindent VHR: very high resolution image, S1/2: Sentinel 1/2, LS: Landsat, AL: ALOS-2, MO: MODIS} 
    \label{tab:experiments2}
    \centering
\small{
\begin{tabular}{p{2.5cm}@{}p{1.8cm}@{}
>{\centering\arraybackslash}p{0.54cm}@{} 
>{\centering\arraybackslash}p{0.54cm}@{}
>{\centering\arraybackslash}p{0.54cm}@{} 
>{\centering\arraybackslash}p{0.54cm}@{} 
>{\centering\arraybackslash}p{0.54cm}@{} 
>{\raggedleft\arraybackslash}p{0.75cm}}
\toprule
Model & Pre-training & \multicolumn{5}{c}{Modalities} \\
\midrule
\multicolumn{2}{l}{PLANTED - classif.} & S1 & S2 & LS & AL & MO & mF1
\\\midrule
\bf AnySat (ours) & None &  \yes & \yes & \yes & \yes & \yes & 61.2 \\
\bf AnySat (ours) & GeoPlex & \yes & \yes & \yes & \yes &\yes  & 61.5 \\\greyrule
ViViT \cite{pazos2024planted,arnab2021vivit} & None & \yes & \yes &  &   &  & \bf 62.2 \\
ViViT \cite{pazos2024planted,arnab2021vivit} & None & \yes & \yes & \yes & \yes &\yes & 59.3\\
\end{tabular}
\begin{tabular}{p{2.5cm}@{}p{1.8cm}@{}
>{\centering\arraybackslash}p{0.65cm}@{} 
>{\centering\arraybackslash}p{0.65cm}@{} 
>{\centering\arraybackslash}p{0.65cm}@{} 
>{\centering\arraybackslash}p{0.75cm}@{}
>{\raggedleft\arraybackslash}p{0.75cm}}
\midrule
\multicolumn{2}{l}{PASTIS-HD - semantic seg} &  VHR & S1 & S2 & OA & mIoU
\\\midrule
\bf AnySat (ours) & None &  \yes & \yes & \yes & 84.8 & 66.3 \\ 
 \bf 66.9 \\ 
\bf AnySat (ours) & GeoPlex & \yes & \yes & \yes & 85.0 & \bf 66.4 \\\greyrule
SkySense $^\star$ \cite{guo2024skysense} & SkySense & \yes & \yes & \yes & \bf 85.9 & -\;\;\, \\ 
UTAE-MM \cite{garnot2022multi} 
& None & & \yes & \yes & 84.2 & 66.3 \\
TSViT \cite{tarasiou2023vits} & None & & & \yes &  83.4 & 65.4 \\
UTAE \cite{garnot2021panoptic} & None & & &\yes  & -  & 63.1 \\
SITSMamba \cite{xiaolei2024sitsmamba} & None & & & \yes & - & 50.1 \\
\end{tabular}
\begin{tabular}{p{2.5cm}@{}p{1.8cm}@{}
>{\centering\arraybackslash}p{1.35cm}@{}
>{\centering\arraybackslash}p{1.35cm}@{}
>{\raggedleft\arraybackslash}p{0.75cm}}
\midrule
\multicolumn{2}{l}{FLAIR - semantic seg} & VHR & S2 & mIoU
\\\midrule
\bf AnySat (ours) & None &  \yes & \yes & 55.6 \\
\bf AnySat (ours) & FLAIR & \yes & \yes& * \\
\bf AnySat (ours) & GeoPlex & \yes & \yes & * \\\greyrule
UT\&T MIT \cite{straka2024modernized} & ImageNet & \yes & \yes & \bf 62.6  \\
UT\&T \cite{garioud2023flair} & ImageNet & \yes & \yes &  56.9 \\
UNet \cite{he2016deep} & ImageNet & \yes &  & 54.7 \\
UTAE \cite{garnot2021panoptic} & None &  & \yes & 36.1
\end{tabular}
\ifx
\begin{tabular}{p{2.5cm}@{}p{1.8cm}@{}
>{\centering\arraybackslash}p{0.675cm}@{} 
>{\centering\arraybackslash}p{0.675cm}@{}
>{\centering\arraybackslash}p{0.675cm}@{} 
>{\centering\arraybackslash}p{0.675cm}@{} 
>{\raggedleft\arraybackslash}p{0.75cm}}
\midrule
\multicolumn{2}{l}{S2NAIP - semantic seg.} & VHR & S1 & S2 & LS & mIoU
\\\midrule
\bf AnySat (ours) & None &  \yes & \yes & \yes & \yes & * \\
\bf AnySat (ours) & S2NAIP & \yes  & \yes & \yes  & \yes & * \\
\bf AnySat (ours) & GeoPlex & \yes & \yes & \yes &\yes  & * \\\greyrule
UT\&T \cite{garioud2023flair} & ImageNet & \yes & \yes & \yes & \yes & * \\
UNet \cite{he2016deep} & ImageNet & \yes & & & & 37.9\\
\end{tabular}
\fi
\begin{tabular}{p{2.5cm}@{}p{1.8cm}@{}
>{\centering\arraybackslash}p{0.9cm}@{} 
>{\centering\arraybackslash}p{0.9cm}@{} 
>{\centering\arraybackslash}p{0.9cm}@{} >{\raggedleft\arraybackslash}p{0.75cm}}
\midrule
\multicolumn{2}{l}{PASTIS-HD - panoptic seg} &  VHR & S1 & S2 & mF1
\\\midrule
\bf AnySat (ours) & None &  \yes & \yes & \yes & * \\
\bf AnySat (ours) & PASTIS-HD & \yes & \yes & \yes& * \\
\bf AnySat (ours) & GeoPlex & \yes & \yes & \yes & * \\\greyrule
UTAE \cite{garnot2021panoptic} & None & & & \yes & \bf 43.8 \\
UTAE-MM \cite{garnot2021panoptic} & None & & \yes & \yes & 42.0 \\
\end{tabular}
}
\end{table}

\fi
\end{document}